\documentclass[preprint,12pt]{elsarticle}

\usepackage{amssymb}
\usepackage{amsmath}
\usepackage[colorlinks=false]{hyperref}
\usepackage{cleveref}
\usepackage{algorithm}
\usepackage{algpseudocode}
\usepackage{tikz}
\usepackage{xspace}
\usepackage{siunitx}
\newcommand{\Po}[0]{VIB\xspace}
\newcommand{\EMBL}[0]{HeLa-EMBL\xspace}
\newcommand{\EPFL}[0]{EPFL\xspace}
\newcommand{\revision}[1]{#1}
\newcommand{\revisiontwo}[1]{#1}
\usepackage{multirow}
\usepackage{subfig}
\usepackage{tikz-3dplot}
\usepackage{xcolor}

\usepackage{setspace}

\journal{Computers in Biology and Medicine}

\begin{document}

\begin{frontmatter}

\title{Patch-based Querying Identifies Structures of Interest in Electron Microscopy} %

\author[telin]{Niels Vyncke\corref{cor}}
\ead{niels.vyncke@ugent.be}
\author[telin,kikirpa]{Nicolas Nadisic}
\author[sterre,vib]{Yvan Saeys}
\author[telin]{Aleksandra Pi\v{z}urica}

\affiliation[telin]{organization={Department of Telecommunications and Information Processing},
            addressline={Sint-Pietersnieuwstraat 41},
            city={Ghent},
            postcode={9000},
            country={Belgium}}
\affiliation[kikirpa]{organization={Royal Institute for Cultural Heritage (KIK-IRPA)},
            addressline={Jubelpark 1},
            city={Brussels},
            postcode={1000},
            country={Belgium}}

\affiliation[sterre]{organization={Department of Mathematics, Computer Science
		and Statistics},
            addressline={Krijgslaan 281},
            city={Ghent},
            postcode={9000},
            country={Belgium}}

\affiliation[vib]{organization={VIB-UGent Center for Inflammation Research},
            addressline={Technologiepark-Zwijnaarde 75},
            city={Ghent},
            postcode={9052},
            country={Belgium}}

\cortext[cor]{Corresponding author.}

\begin{abstract}
Volume electron microscopy (vEM) has emerged as an essential sensing technique in biomedical research, allowing the three-dimensional imaging of biological cells and tissues at nanometer\revision{-scale} resolution. The ability to generate extensive datasets has reached the limitations of downstream analysis processes, which depend significantly on the intervention of human experts for preprocessing and annotation. We propose an efficient and reliable patch-based retrieval framework based on self-supervised learning of local image descriptors to locate self-similar structures in vEM datasets. Given a few manual annotations of a given cellular structure, our method can retrieve similar structures across the EM volume. Our framework is interactive, allowing the human expert to refine the search queries and retrieve relevant image patches quickly and using little labeled data. \revisiontwo{Experiments on real-world vEM images of biological tissues demonstrate that our framework can reliably identify relevant cellular structures, generalize across different organelles and acquisition modalities, and substantially reduce the search space for downstream analysis.}
\end{abstract}


\end{frontmatter}

\section{Introduction}
\label{sec:introduction}

\begin{figure}[t]
	\centering
	\includegraphics[width=0.7\linewidth]{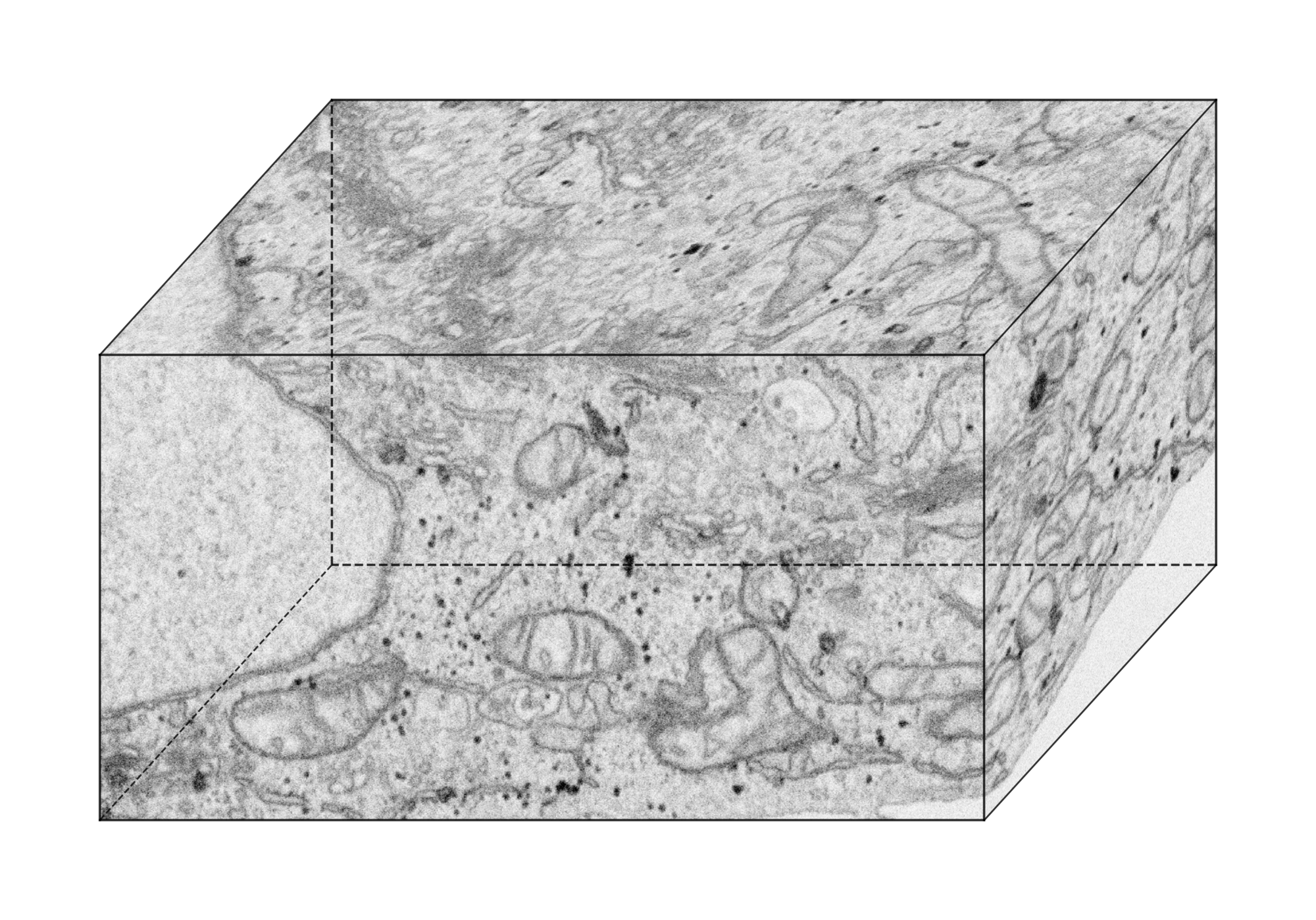}
	\caption{\revision{An illustration of a volume electron microscopy dataset. The dataset contains 377 slices, each of size $508 \times 966$.}}
	\label{fig:dataset}
\end{figure}

High-content microscopy imaging techniques such as volume Electron Microscopy (vEM) have drastically increased our capabilities to image cells in 3D, allowing biologists to study cellular structures at nanometer\revision{-scale} resolution~\citep{czymmek_accelerating_2024,peddie_volume_2022}.
Unlike traditional electron microscopy, which often limits observation to two-dimensional sections, vEM allows reconstructing entire cell volumes, thereby enabling a more comprehensive understanding of subcellular architecture~\citep{peddie_volume_2022,collinson_volume_2023}. \revisiontwo{The} \revision{vEM datasets represent} a sequence of two-dimensional slices representing the whole volume, \revision{which are acquired} by iteratively removing thin layers of the sample -- using methods such as mechanical sectioning, laser ablation, or other physical or optical techniques -- as illustrated in \Cref{fig:dataset}.

\revision{Nowadays, volume electron microscopy techniques generate data faster than downstream analysis can be performed, since exploration and interpretation remain complex and are limited by available human and computational resources}\revisiontwo{~\citep{czymmek_accelerating_2024}.}
Existing deep learning techniques have been used successfully, but they often require substantial amounts of expert-labeled data~\citep{moen2019deep}.
\revisiontwo{Moreover, many supervised vEM pipelines are tailored to a specific structure, remain fixed after training, and provide limited opportunities for user interaction.}
\revisiontwo{Adapting them to a new target structure may therefore require additional annotation and retraining.}

\revisiontwo{Our work instead addresses an upstream stage of the analysis pipeline: given only a few example patches, it returns a ranked set of candidate regions rather than pixel-accurate labels.}
\revisiontwo{We stress that this is not intended to \emph{compete} with structure-specific segmentation methods. Rather, it addresses the complementary problem of locating \emph{any} structure of interest before committing expert time or a dedicated model to the full volume. The framework provides (i) \emph{target flexibility}, by querying new structures through examples; (ii) \emph{data triage}, by discarding irrelevant regions before downstream analysis; and (iii) \emph{human-in-the-loop exploration}, by allowing queries to be refined interactively.}

In this work, our goal is to enable efficient search in large image stacks by developing an interactive patch-based querying framework. 
This method is designed to assist researchers in conducting studies of specific regions of interest by enabling them to retrieve relevant structures within vEM datasets.
The application of learned local image descriptors represents a novel approach that has the potential to considerably accelerate the search process in comparison to exhaustive manual search.

Local image descriptors are dimensionality reduction techniques that provide a compact set of features describing a local image region, such as a two-dimensional image patch. 
By efficiently encoding and compressing the essential visual content, the local image descriptors facilitate comparisons between image regions and search for patterns of interest. This is essential for a wide range of image processing tasks, such as object detection or pattern recognition.
	
Traditional handcrafted local image descriptors such as SIFT~\citep{lowe_object_1999}, SURF~\citep{bay_surf_2006}, and HOG~\citep{dalal_histograms_2005} have been widely used in computer vision. However, in recent years, the focus has shifted to learned descriptors~\citep{4269996,zizakic_efficient_2022,zizakic_nina_and_ito_izumi_and_pizurica_aleksandra_learning_2019}, that are based on features learned by a deep learning model. 
In contrast to handcrafted descriptors, learned descriptors can be fine-tuned on a particular target dataset without manual parameter tuning. 
Previous research on learned descriptors has mainly focused on supervised learning, which relies on abundant labeled data. 
In contrast, self-supervised methods such as autoencoders have been much less explored in this context~\citep{zizakic_nina_and_ito_izumi_and_pizurica_aleksandra_learning_2019}, although they offer huge advantages in the sense that they learn from unlabeled data.
	
A key objective of the present work is to explore the potential of self-supervised learning for encoding efficiently patterns of interest in vEM imaging and to develop an effective search framework based on these principles. In contrast to other representation learning models~\citep{zinchenko_morphofeatures,dorkenwald_multi-layered_2023,han_self-supervised_2022}, 
\revisiontwo{our method can be applied
after a lightweight, unsupervised fine-tuning step on each target dataset. We evaluate this on three \revisiontwo{focused ion beam scanning electron microscopy (FIB-SEM)} datasets (\Cref{sec:datasets}), and one \revisiontwo{serial block-face scanning electron microscopy (SBF-SEM)} dataset, that are each fine-tuned and evaluated separately.}

Our main contributions are as follows:
\begin{itemize}

    \item We conduct an elaborate study into the performance of learned descriptors to locate self-similar structures of interest in vEM datasets. To the best of our knowledge, no work has reported the use of local image descriptors to search for self-similar structures in EM datasets.
    \revisiontwo{We further demonstrate that the proposed descriptors can retrieve morphologically diverse cellular structures, including mitochondria, endoplasmic reticulum, nuclear envelope, vesicles, alpha granules, and the canalicular system, and generalize to strongly anisotropic SBF-SEM data.}

    \item We propose an original and efficient framework for identifying the structures of interest in large datasets based on local image descriptors. Its important innovation lies in improving the search accuracy by allowing multiple positive and negative query patches, elegantly encoded into positive and negative search trees. Another important novelty is an effective similarity metric defined with respect to the encodings of the positive and negative search trees, which \revisiontwo{enables reliable identification of} candidate patches that contain structures of interest for a given task or query.
    \revisiontwo{The resulting framework provides a screening step that reduces the amount of data requiring manual inspection, without training a dedicated structure-specific detector.}

    \item Our study \revisiontwo{also} reveals the effect of fine-tuning the pretrained descriptors, which were initially trained on general image benchmarks, by adapting them to specific vEM datasets. This process helps to tailor the learned features more effectively to the particular dataset. We compare the performance of both the pretrained and fine-tuned descriptors through visual and quantitative evaluations.
    \revisiontwo{We additionally compare the learned descriptors with classical handcrafted alternatives, including SIFT, HOG, and LBP, and show improved retrieval performance across multiple structures and datasets.}

    \revisiontwo{\item We quantify the practical value of the proposed retrieval framework as a screening step for downstream analysis. %
    Our experiments show that a substantial fraction of patches can be discarded based on their similarity scores while retaining most occurrences of the target structure, thereby reducing the search space for subsequent analysis.}

\end{itemize}

The remainder of this paper is organized as follows.
In \Cref{sec:methods}, we describe the proposed method for retrieving structures of interest in vEM data.
In \Cref{sec:implementation}, we provide additional implementation details beyond the core methodology, with a particular focus on the descriptor models and how they were trained.
In \Cref{sec:results}, we present the experimental evaluation conducted on three \revisiontwo{FIB-SEM} vEM datasets\revisiontwo{, and one SBF-SEM dataset}. 
We first describe the datasets, evaluation metrics, and the automatic query selection strategy. We then analyze the discriminative capability of self-supervised local image descriptors for different biological structures. Finally, we perform a comprehensive evaluation of the proposed search framework in a real-world detection scenario\revisiontwo{, including comparisons with handcrafted descriptors, an analysis of its potential to reduce the search space for downstream processing, and an evaluation on strongly anisotropic SBF-SEM data}.

\section{The proposed patch-based search framework}
\label{sec:methods}

This section describes the proposed method for retrieving structures of interest using learned local image descriptors. We start by outlining the general methodology. 
Then, we detail the computation of local image descriptors, and we introduce a novel similarity metric to compare them.
Finally, we describe the overall search framework.

\subsection{Self-supervised learning of local image descriptors}
\label{sec:general-methodology}

In vEM data, each 2D slice is a grayscale image. We consider an image patch to be a small (typically rectangular) region of a 2D slice, i.e. a patch is a small 2D image. 
\revisiontwo{Note that extending the current framework to 3D patches would require a substantially different approach, and it is out of the scope of the present work.}

We aim to learn local image descriptors that map each patch to an encoding, which is a representation in a lower-dimensional latent space, so that patch similarities can be assessed efficiently.

\textit{Autoencoders} (AEs)~\cite{hinton_reducing_2006, baldi_autoencoders_2012, bank_autoencoders_2023} provide a natural framework for learning such local image descriptors in a self-supervised manner. An autoencoder is a neural network trained to compress input data into a low-dimensional latent representation while preserving the information necessary to reconstruct the original input. When applied to image patches, convolutional autoencoders learn compact encodings that capture salient local structures, making the latent representations suitable as patch descriptors.

Concretely, the encoder $\mathcal{E}$ maps an input patch $\mathbf{p}$ to a latent code $\mathbf{z}=\mathcal{E}(\mathbf{p})$, which serves as the descriptor of the patch, while the decoder $\mathcal{D}$ reconstructs the patch as $\hat{\mathbf{p}}=\mathcal{D}(\mathbf{z})$. The dimensionality of $\mathbf{z}$ is chosen such that $d \ll MN$, enforcing a bottleneck that encourages the encoder to retain only the most informative features of the patch.

Autoencoders are trained by minimizing a \textit{reconstruction loss} over a set of patches $\mathcal{P}$:
\begin{equation}
\mathcal{J}_{\text{AE}}(\mathcal{P}, \mathcal{E}, \mathcal{D}) = \sum_{\mathbf{p}\in \mathcal{P}}\mathcal{L}(\mathbf{p}, \mathcal{D}(\mathcal{E}(\mathbf{p}))),
\label{eq:ae-loss}
\end{equation}
where $\mathcal{L}$ measures the discrepancy between the input patch and its reconstruction.

By optimizing this objective, the encoder $\mathcal{E}$ learns to map patches with similar local appearance and structure to nearby points in the latent space. As a result, similarities between patches can be efficiently assessed by computing distances between their latent codes, without requiring explicit reconstruction.

Let $\mathcal{E}(\mathbf{p})\in \mathbb{R}^d$ denote the $d$-dimensional encoding of the image patch $\mathbf{p}\in \mathbb{R}^{M\times N}$. In this work, we assume a pretrained encoder $\mathcal{E}$~\citep{zizakic_efficient_2022}, trained as part of a self-supervised autoencoder
\begin{equation}
\mathcal{A}: \mathbb{R}^{M\times N} \rightarrow \mathbb{R}^{M\times N},\mathbf{p}\mapsto\hat{\mathbf{p}}=\mathcal{D}\left(\mathcal{E}(\mathbf{p})\right),
\end{equation}
which is optimized such that $\hat{\mathbf{p}}\approx\mathbf{p}$; where $\mathcal{D}$ represents the decoder part of $\mathcal{A}$. The encoder $\mathcal{E}$ is subsequently used as a fixed feature extractor to compute local descriptors for image patches.

\subsection{Discriminative analysis of the latent feature space}

\begin{figure}[t]
	\centering
	\includegraphics[width=0.7\linewidth]{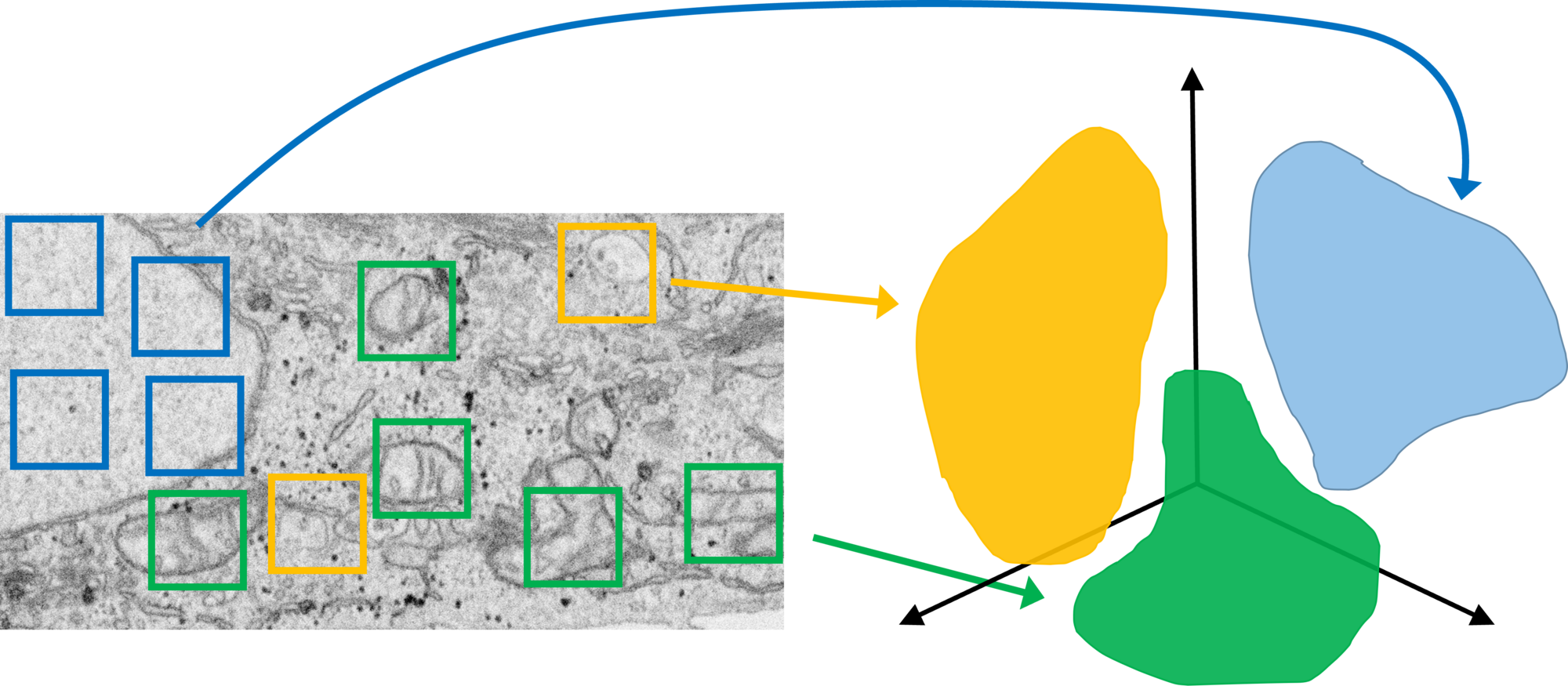}
	\caption{Visualization of the proposed methodology. Ideally, the {local image descriptor} maps patches corresponding to different structures onto distinct manifolds in the latent space, which can be learned by extracting and encoding query patches.}
	\label{fig:methodology}
\end{figure}

We first investigate the potential of learned local descriptors to map the different image structures to distinct clusters in the latent space. This idea is visualized in \Cref{fig:methodology}, where we select patches of three structures whose encodings are mapped onto lower-dimensional manifolds in the latent space.

If the latent space of the learned image descriptors shows meaningful clustering of the structures of interest, this is a clear indication that the learned descriptor allows the construction of an effective patch search framework.
We will develop a concrete patch search framework based on the nearest neighbors of a set of query patches.

\begin{figure}[t]
    \resizebox{\linewidth}{!}{\begin{tikzpicture}
        \node[anchor=west] at (0,0) {\includegraphics[width=0.45\linewidth]{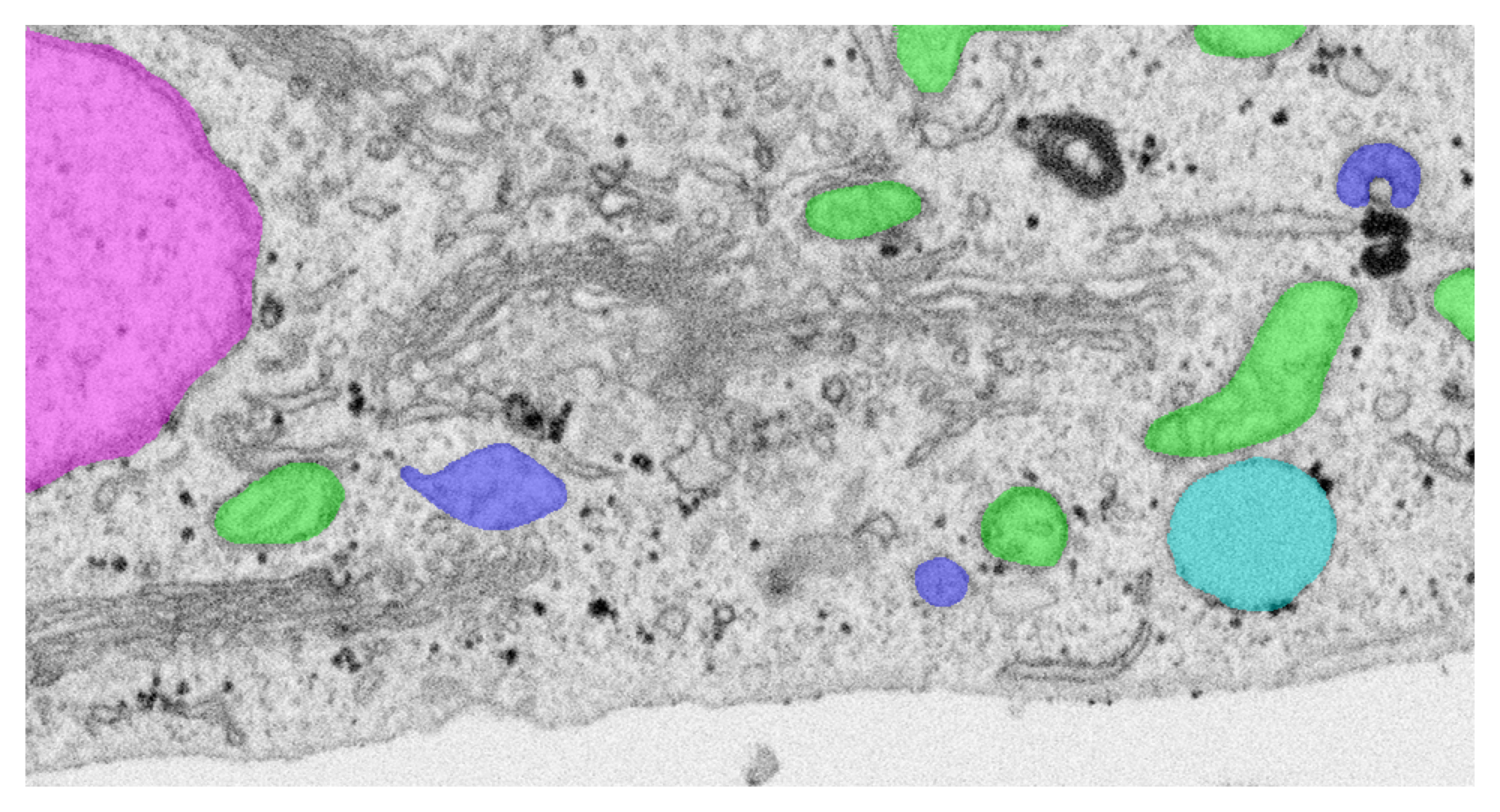}};
        \node[anchor=west] at (0.5\linewidth,0) {\includegraphics[width=0.45\linewidth,trim={60 0 60 0},clip]{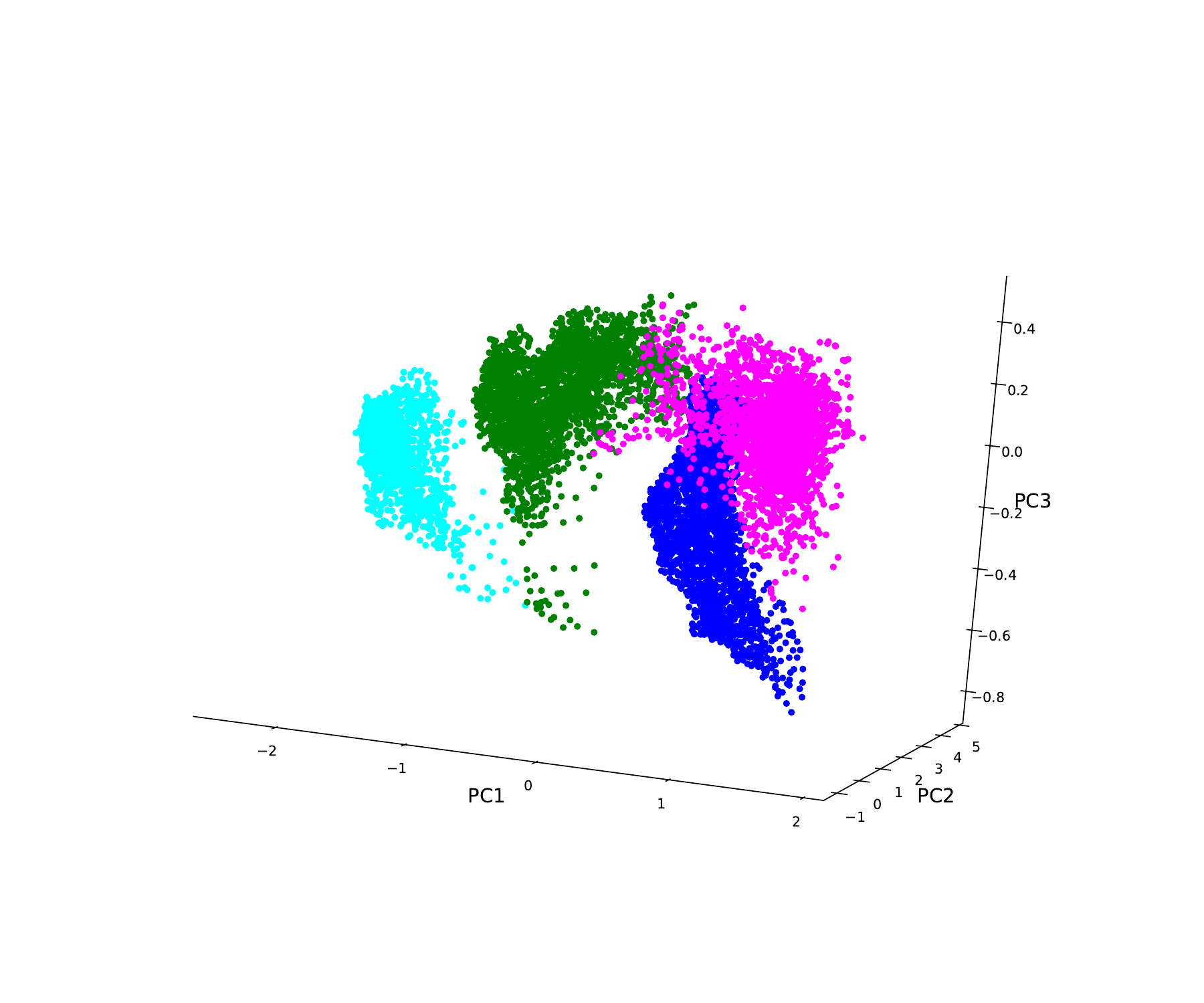}};
        \draw[->,thick]  (0.47\linewidth, 0) -- (0.6\linewidth,0) node[midway,below,font=\small] {PCA}; 
    \end{tikzpicture}}
    \caption{Proof-of-concept visualization of the proposed descriptor. (Left) \revision{Color-coded overlay of the patches selected as references for each of the four structures (one color per structure).} \revisiontwo{The pixel-level colors are manual annotations used only to identify which patches contain each structure; the proposed method neither predicts nor reconstructs these pixel-level masks, and no segmentation is performed.} (Right) \revision{For every patch in the slice, a four-dimensional descriptor is computed as the average \revisiontwo{mean squared error (MSE)} distance to each structure's reference set; the resulting points are projected to three dimensions using principal component analysis (PCA).}}
    \label{fig:proof-of-concept}
\end{figure}

As illustrated in \Cref{fig:proof-of-concept}, the learned descriptor organizes the latent space in a meaningful and structured manner, leading to clear clustering of patches associated with distinct anatomical structures. This behavior provides strong empirical support for the validity of the proposed methodology. To obtain this result, patches were extracted from four different structures, with ten patches per structure selected to form reference sets. For each remaining patch, the average mean squared error (MSE) distance to each reference set was computed, yielding a four-dimensional representation. This 4D space was subsequently projected into three dimensions using principal component analysis for visualization. The resulting separation between clusters demonstrates the discriminative capability of the learned descriptor and its effectiveness in capturing structural differences.

\subsection{Search with positive and negative query trees}

Here, we present the proposed patch-based search algorithm to locate cellular structures of interest in volume electron microscopy images. 

The core idea of the proposed approach is to efficiently locate candidate patches that contain biomedical features of interest. To achieve this, we define a set of query patches that serve as reference examples for the structures we aim to detect.

Given the large scale of vEM datasets, exhaustive comparison of each candidate patch against all query patches would be computationally prohibitive. To enable efficient similarity search, we employ approximate nearest neighbor (ANN) search trees, which allow sublinear-time retrieval of the most similar query patches for any given candidate.

To enable discrimination between relevant and irrelevant regions, we create sets of positive and negative query patches: $\mathcal{Q}^+$ and $\mathcal{Q}^-$ respectively, where positive patches contain the structure of interest and negative patches do not. The encodings of these query patches are stored in separate ANN search trees, allowing fast retrieval of the closest query patches for any given candidate patch. We use the Spotify Annoy library for creating the ANN search trees~\citep{annoy}.

Let $\mathcal{T}^+$ and $\mathcal{T}^-$ denote the search trees containing respectively the encodings from $\mathcal{Q}^+$ and $\mathcal{Q}^-$ using $\mathcal{E}$. The search trees are constructed to store the feature vectors or encodings of the query patches. Given two patch encodings $\mathbf{x}$ and $\mathbf{y}$, we measure their similarity using Euclidean distance, defined as:

\begin{equation}
    \left\|\mathbf{x}-\mathbf{y}\right\|_2=\sqrt{\sum\limits_{i=1}^n\left(x_i-y_i\right)^2}.
    \label{eq:eucl}
\end{equation}

Let $\mathbf{p}_i,\, i=1,...,m$ denote all patches from the given vEM dataset that we want to search. For each patch $\mathbf{p}_i$, we compute the embedding $\mathcal{E}(\mathbf{p}_i)$ and search both trees $\mathcal{T}^+$ and $\mathcal{T}^-$ for the $k$ nearest neighbors, with respect to the Euclidean distance.

Then, let  $\mathcal{D}^+$ and $\mathcal{D}^-$ denote the sets of Euclidean distances to the nearest $k$ positive queries and $k$ negative queries, respectively\revision{, and let $\left\langle\mathcal{D}^+\right\rangle$ and $\left\langle\mathcal{D}^-\right\rangle$ denote the arithmetic means of these two sets (still in the distance domain, $\in[0,\infty)$).}
\revision{We then convert each mean distance to a similarity in the range $(0,1]$ by a single application of the negative exponential function $f(x)=\exp(-x)$, i.e.\ $\exp(-\langle\mathcal{D}^+\rangle)$ and $\exp(-\langle\mathcal{D}^-\rangle)$.}
We then define the similarity score, being the overall similarity of the patch with respect to the query patches $\mathcal{Q}^+$ and $\mathcal{Q}^-$ as
\begin{equation}
	\mathcal{S}\left(\mathcal{D}^+,\mathcal{D}^-\right) = \exp{\left(-\left\langle\mathcal{D}^+\right\rangle\right)}-\exp{\left(-\left\langle\mathcal{D}^-\right\rangle\right)}.
	\label{eq:similarity}
\end{equation}
The proposed similarity metric \eqref{eq:similarity} balances the importance of positive and negative query patches by projecting them on the same interval.
By subtracting the similarity value with respect to $\mathcal{D}^-$ from that with respect to $\mathcal{D}^+$, as we do in \eqref{eq:similarity}, we obtain an overall similarity value between $-1$ and $1$, where higher values indicate greater similarity and lower values indicate greater dissimilarity.

The primary objective of the proposed methodology is to \revisiontwo{reliably identify} locations in large datasets where structures of interest are likely to be present. This should reduce the amount of data to be processed by a human expert looking for \revisiontwo{a} specific structure, and/or facilitate subsequent automatic processing steps by discarding irrelevant regions while retaining potential regions of interest. Therefore, false positives will be less penalized than false negatives.

Having computed the similarity score $S_i$ for all the patches $\mathbf{p}_i$, we retain $N$ patches with the highest similarity values. Practically, we do this by maintaining a list $L$ of tuples $(\mathbf{p}_i, S_i)$ as it is detailed in \Cref{alg:method}.

\begin{figure*}
	\centering
    \resizebox{\linewidth}{!}{\input{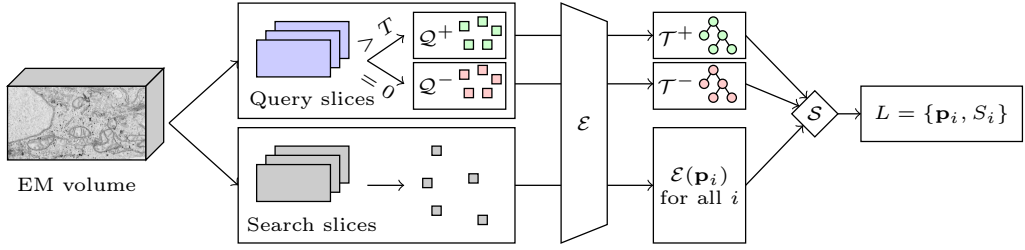}}
	\caption{Illustration of our patch-based querying method, with $\mathcal{E}$ representing the encoder model, $\mathcal{Q}^+$ and $\mathcal{Q}^-$ representing the sets of positive and negative queries, with $\mathcal{T}^+$ and $\mathcal{T}^-$ representing the positive and negative search trees, and $\mathcal{S}$ the similarity function as defined by \Cref{eq:similarity}.}
	\label{fig:method}
\end{figure*}

The proposed framework is visually represented in \Cref{fig:method}.
\begin{singlespace}
\renewcommand{\algorithmicrequire}{\textbf{Input:}}
\renewcommand{\algorithmicensure}{\textbf{Output:}}
\begin{algorithm}
	\caption{The proposed patch-based search framework.}
	\label{alg:method}
		\begin{algorithmic}[1]
			\Require \parbox[t]{\linewidth}{2D patches $\mathbf{p}_i$, $i=1,...,m$ \\
				A pretrained encoder model $\mathcal{E}$\\
				A set $\mathcal{Q}^+$ of $n^+$ positive query patches  \\
				A set $\mathcal{Q}^-$ of $n^-$ negative query patches \\
				A number $N$ of patches to return\\}
			\Ensure \parbox[t]{\linewidth}{A list of $N$ {selected} patches \\}
			
			\Function{CreateTree}{$\mathcal{E}, \mathcal{Q}$}
			\State Initialize search tree $\mathcal{T}$
			\For{\textbf{each} query patch $\mathbf{q}_i \in \mathcal{Q}$}
			\State Compute encoding $\mathcal{E}(\mathbf{q}_i)$
			\State $\mathcal{T} \gets \mathcal{T} \cup \mathcal{E}(\mathbf{q}_i)$ \Comment{Add encoding to search tree}
			\EndFor
			\State \Return $\mathcal{T}$
			\EndFunction
			
			\State
			\State $\mathcal{T}^+ \gets$ \Call{CreateTree}{$\mathcal{E},\mathcal{Q}^+$} \Comment{Create pos search tree}
			\State $\mathcal{T}^- \gets$ \Call{CreateTree}{$\mathcal{E},\mathcal{Q}^-$} \Comment{Create neg search tree}
			\State
			\State Initialize new list $L$
			\For{\textbf{each} image patch $\mathbf{p}_i$ for $i \in \{1, \ldots, m\}$}
			\State Compute encoding $\mathcal{E}(\mathbf{p}_i)$
			\State
			\State $\triangleright$ \textbf{Get k-NNs from pos and neg search tree}
			\State $\mathcal{D}^+ \gets$ \Call{Query}{$\mathcal{T}^+,\mathcal{E}(\mathbf{p}_i),k$} \Comment{Query pos tree}
			\State $\mathcal{D}^- \gets$ \Call{Query}{$\mathcal{T}^-,\mathcal{E}(\mathbf{p}_i),k$} \Comment{Query neg tree}
			\State
			\State $S_i \gets \mathcal{S}(\mathcal{D}^+,\mathcal{D}^-)$\Comment{Compute similarity $S_i$ (\cref{eq:similarity})}
			\State $L \gets L \cup (\mathbf{p}_i,S_i)$
			\EndFor
			\State
			\State\Return The $N$ tuples $(\mathbf{p}_i, S_i)$ corresponding to the top $N$ similarities $S_i$.
	\end{algorithmic}
\end{algorithm}
\end{singlespace}

\section{Implementation details}
\label{sec:implementation}

This section describes the more technical implementation details, including model architectures, pretraining, and fine-tuning procedures.

\subsection{Model architectures}
\label{sec:models}

For the search framework, we considered both a classical and a variational autoencoder.

The encoder part of the classical autoencoder consists of 3 convolutional layers with kernel size 3 $\times$ 3, each of them followed by a max-pooling layer and one fully connected layer at the end. The decoder part is the inverse of the encoder, with transposed convolutional layers with kernel size 2 $\times$ 2 and stride 2, and without max-pooling layers. The classical autoencoder is trained using the multiscale structural similarity (MS-SSIM)~\citep{1292216}, a perceptual loss function.

The variational autoencoder is based on the classical autoencoder, and the fully connected layer on the encoder side is replaced by two parallel ones, one for the mean and one for the variance of the Gaussian distribution. The variational autoencoder has an additional KL-divergence term, weighted by the $\beta$ parameter. The models were implemented using the PyTorch library. For the precise details on the implementation, we refer to \cite{zizakic_efficient_2022} and the supplementary code repository.

\subsection{Pretraining and fine-tuning procedures}
\label{sec:training-fine-tuning}

In this section, we discuss why and how we use pretraining and fine-tuning of the autoencoder models described previously.

Pretrained models have learned general image features that can be used for specific image processing applications. Fine-tuning these models provides the additional benefit of slightly adapting these general features to the target dataset. We use the classical and variational autoencoders proposed by
\v{Z}i\v{z}aki\'{c} \& Pi\v{z}urica~\citep{zizakic_efficient_2022} to solve these types of problems by extending the classical autoencoder~\citep{hinton_reducing_2006}
and variational autoencoder~\citep{kingma_auto-encoding_2022}.

The decision to start from generally trained models and fine-tune them can be motivated as follows:
\begin{itemize}
	\item Fine-tuning existing models helps mitigate overfitting to the target dataset, which is particularly important in vEM, where the statistical properties of different datasets can vary significantly.
	\item The reconstruction loss achieved during the training of descriptor models does not accurately reflect the model's performance when applied to image processing tasks. Consequently, these models are typically evaluated on specific tasks using benchmarks like HPatches~\citep{HPatches}, which helps determine the optimal hyperparameters. However, in the case of vEM, no equivalent benchmark exists for such evaluations.
	\item Our previous experiments have shown inferior results when training from scratch on vEM data compared to the pretrained models. Therefore, we apply a transfer learning approach~\citep{technologies11020040}, adapting general image features to model biomedical image details, which was achieved through a fine-tuning process as explained in \Cref{sec:fine-tuning}.
\end{itemize}

\subsubsection{Pretraining}

The models are trained using 125,000 patches of size 65 $\times$ 65 from different image datasets: ImageNet~\citep{5206848}, KonIQ~\citep{koniq10k} and Visual Genome~\citep{visualgenome}. The ratio between the training, validation, and test sets is 8:1:1. The models are trained for 400 epochs at a learning rate of $10^{-4}$, with a batch size of 32 and a latent space size of 32. The ReLU activation function was used. The optimal $\beta$ parameter of the variational autoencoder is $10^{-5}$. No data augmentation is performed.

The HPatches~\citep{HPatches} benchmark is used for evaluating and tuning hyperparameters for various image processing tasks, including matching, retrieval, and verification. For more details on the pretraining process, we refer to \cite{zizakic_efficient_2022}.

\subsubsection{Fine-tuning}
\label{sec:fine-tuning}

The models are optimized through continuous learning at a diminished learning rate of $10^{-5}$ for a limited number of epochs, specifically 20. The remaining hyperparameters are consistent with those of the pretrained models. We extract 108,000 patches from the target datasets utilizing the patch selection strategy as outlined in \cite{zizakic_efficient_2022}.

\section{Results and discussion}
\label{sec:results}

In this section, we first test the capability of the learned descriptors to discriminate between different structures of interest. Then, we show some examples of returned patches given a particular positive and negative query. Finally, we conduct an empirical evaluation of our search framework on \revisiontwo{three primary FIB-SEM datasets, followed by an additional SBF-SEM evaluation}.

\subsection{Validation datasets}
\label{sec:datasets}

We used the following \revisiontwo{FIB-SEM} datasets for the fine-tuning and validation of the proposed framework:

\begin{itemize}
	\item \EMBL~\revision{\citep{hennies2020empiar10311}}:
    The \EMBL dataset is a $13 \times 30 \times 12 \, \si{\micro\meter\cubed}$ FIB-SEM dataset, acquired by Dr.~Anna Steyer and Dr.~Yannick Schwab of the Volume Correlative Light and Electron Microscopy facility at EMBL, which covers a complete HeLa cell. The full volume has $2592 \times 6004 \times 1465$ voxels and was acquired at a near-isotropic resolution of $5 \times 5 \times 8 \, \si{\nano\meter\cubed}$. A $512 \times 512 \times 64$ subvolume was labeled by an expert for mitochondria and ER (endoplasmic reticulum).
	\item \EPFL~\citep{EPFL_EM_dataset}: The \EPFL dataset 
    represents a $5\times 5\times 5~\si{\micro\meter\cubed}$ section taken from the CA1 hippocampus region of a mouse brain, corresponding to a $2048 \times 1536 \times 1065$ volume. Two $1048 \times 786 \times 165$ subvolumes were manually labeled by experts for mitochondria. The data was acquired by a focused ion-beam scanning EM, and the resolution of each voxel is approximately $5 \times 5 \times 5~\si{\nano\meter\cubed}$.
    \item \Po: The \Po dataset is an in-house FIB-SEM dataset generated at the Flemish Institute for Biotechnology Bioimaging Core. 
    \revisiontwo{It represents a $5\times 5\times 5~\si{\nano\meter\cubed}$ isotropic voxel resolution FIB-SEM dataset from a HeLa cell line (a cancer cell line).}
\end{itemize}

These datasets contain both the raw data and segmentation maps indicating mitochondria and other structures. The dimensions of the datasets are specified in \Cref{tab:dataset_properties}.

\begin{table}
	\caption{Dimensions of the vEM datasets studied.}
	\label{tab:dataset_properties}
	\centering
	\begin{tabular}{ccc}
		\hline
		\multirow{2}{*}{\textbf{Dataset}} & \multirow{2}{*}{\textbf{Number of slices}} & \textbf{Slice dimensions}\\
		&&\textbf{(h $\times$ w)}\\
		\hline
		{\EMBL} & 64 & 512$\times$512\\
		{\Po} & 377 & 508$\times$966\\
		{\EPFL} & 330 & 768$\times$1024\\
        \hline
        Platelet-EM & 50 & 800$\times$800\\
		\hline
	\end{tabular}
\end{table}

\revisiontwo{In \Cref{sec:sbf-sem}, we additionally evaluate the framework on a strongly anisotropic SBF-SEM dataset: the Platelet-EM dataset~\citep{guay_platelet_2021}. The \revisiontwo{Platelet-EM} dataset is a publicly available SBF-SEM volume of a human platelet, acquired with a Gatan 3View system at an approximate resolution of $10 \times 10 \times 40$--$50\,\si{\nano\meter\cubed}$. We use the $50$-slice, $800\times 800$ subvolume, which provides semantic labels for the cell body and five organelles (mitochondria, alpha granule, canalicular system, dense granule body, and dense granule core).}

\subsection{Evaluation metrics}

We evaluate our framework using the following metrics:

\begin{itemize}
	\item \textit{Precision,} that is the
	number of returned patches that overlap with the ground truth regions of the structure of interest, divided by the total number of returned patches. Or the number of true positives (TP) divided by the number of true and false positives (FP).
	\begin{equation}
	\text{Precision}=\frac{TP}{TP+FP}
	\label{eq:precision}
	\end{equation}
	An overlap of 1 pixel is already sufficient because our search framework aims to locate a structure of interest, rather than delineating the exact boundary.
	\item \textit{Recall,} that is the
	number of pixels of retrieved structures, divided by the total number of pixels of all occurrences of the structure. Or the number of true positives divided by the number of true positives and false negatives (FN).
	\begin{equation}
	\text{Recall}=\frac{TP}{TP+FN}
	\label{eq:recall}
	\end{equation}
	A structure is retrieved if at least one patch overlaps with it. We take into account the number of pixels to give more importance to larger structures.
\end{itemize}

\revision{Precision measures the quality of the retrieved patches, that is, what fraction of the retrieved patches truly belong to the structure of interest. Recall measures the completeness of the retrieval, using the pixel-weighted fraction of retrieved structures so that larger structures contribute proportionally more.} Ideally, \revision{both} precision \revision{and recall} should be 1.

We use precision and recall to visualize the Precision-Recall curves (PR-curves), which are often used as an evaluation method for retrieval frameworks. PR-curves are computed using different thresholds, which objectively measure the retrieval performance irrespective of the chosen threshold.

Finally, we also compute the area under the PR curves, which is called the average precision (AP). 
The average precision can be used to compare the performance of different techniques.

\subsection{Query selection strategy}

\begin{figure*}
    \centering
    \includegraphics[width=\linewidth]{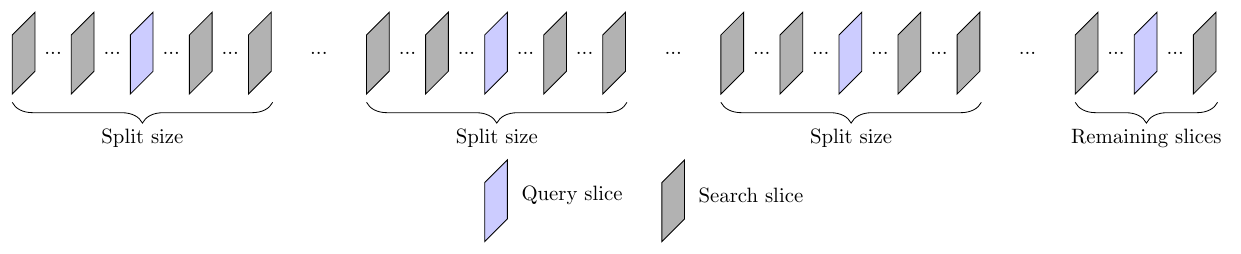}
    \caption{Illustration of the query selection strategy. The dataset is divided into equally sized groups (split size). From each group, the central slice is chosen as the query slice (blue), while the remaining slices together form a separate group.}
    \label{fig:query-selection}
\end{figure*}

To mimic the way a human user would input labeled data into the framework, and using the ground-truth segmentation masks provided with the datasets, we implement a query selection strategy to extract positive and negative query patches evenly distributed across the vEM volume.
We divide the dataset into groups with an equal number of slices, defined as the split size. The remaining slices are also considered as one group, even if it has fewer slices. The middle slice from each group is selected as a query slice, as is visually represented in \Cref{fig:query-selection}. From these query slices, we extract patches measuring $80 \times 80$ pixels \revision{using a regular non-overlapping sliding window (i.e.\ stride equal to the patch size, so that patches tile the slice without overlap). The same scheme is applied both at query time and at evaluation time on every non-query slice of the volume}. A patch is labeled as positive if it overlaps with a mitochondrion by at least 50~\% of its pixels and as negative if there is no overlap. \revision{Patches with intermediate overlap (between $0$\% and $50$\%) are intentionally discarded for query construction, since they would provide an ambiguous training signal to the descriptor-based nearest-neighbor lookup; at evaluation time, in contrast, all patches in the dataset are scored, and a retrieved patch is counted as a true positive whenever it overlaps a target structure.} The split size, which determines the number of slices per group, is set to 10, 20, 30, 50, or 100, corresponding to 10~\%, 5~\%, 3.33~\%, 2~\%, and 1~\% of labeled slices, respectively. However, for the \EMBL dataset, which contains only 64 slices, split size 100 is excluded. A $k$-value of 1 is used in the nearest neighbor search, which proved optimal for this query selection approach.

\subsection{Visual evaluation of the local image descriptors}
\label{sec:tsne}

To gain insights into the performance of AEs and VAEs for locating structures of interest, we analyze how different structures are mapped in the latent space using t-distributed stochastic neighbor embedding (t-SNE)~\citep{TSNE} via scikit-learn~\citep{scikit-learn}. \revisiontwo{We emphasize that t-SNE is used exclusively for post hoc visualization of the learned latent space in this subsection; it plays no role in descriptor computation, similarity scoring, or retrieval, which are described in \Cref{sec:methods} and rely solely on the (V)AE encoder $\mathcal{E}$ and the similarity metric of \Cref{eq:similarity}.} \revision{Whereas the proof-of-concept visualization of \Cref{fig:proof-of-concept} relied on PCA to inspect a hand-crafted four-dimensional descriptor (where preserving global geometric distances is the property of interest), we use t-SNE here because the relevant signal in the $32$-dimensional learned latent space is the local separability of clusters of similar patches, which non-linear neighbour-preserving methods such as t-SNE are explicitly designed to highlight.}

\subsubsection{Pretrained vs. fine-tuned descriptors}

\begin{figure*}[t]
	\centering
	\subfloat[Pretrained classical autoencoder.]{%
		\includegraphics[width=0.49\textwidth]{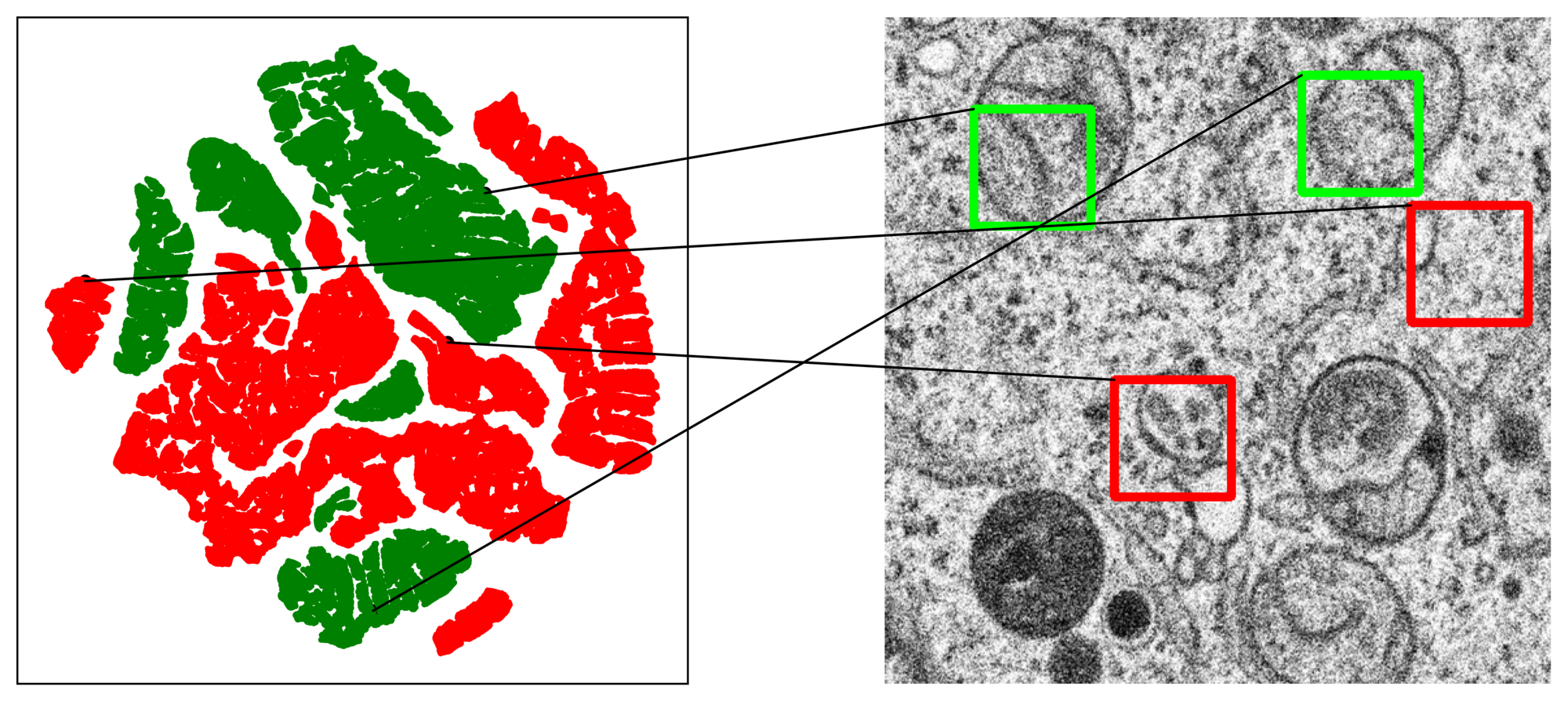}%
		\label{fig:clustering-ae}%
	}
	\hfill
	\subfloat[Fine-tuned classical autoencoder.]{%
		\includegraphics[width=0.49\textwidth]{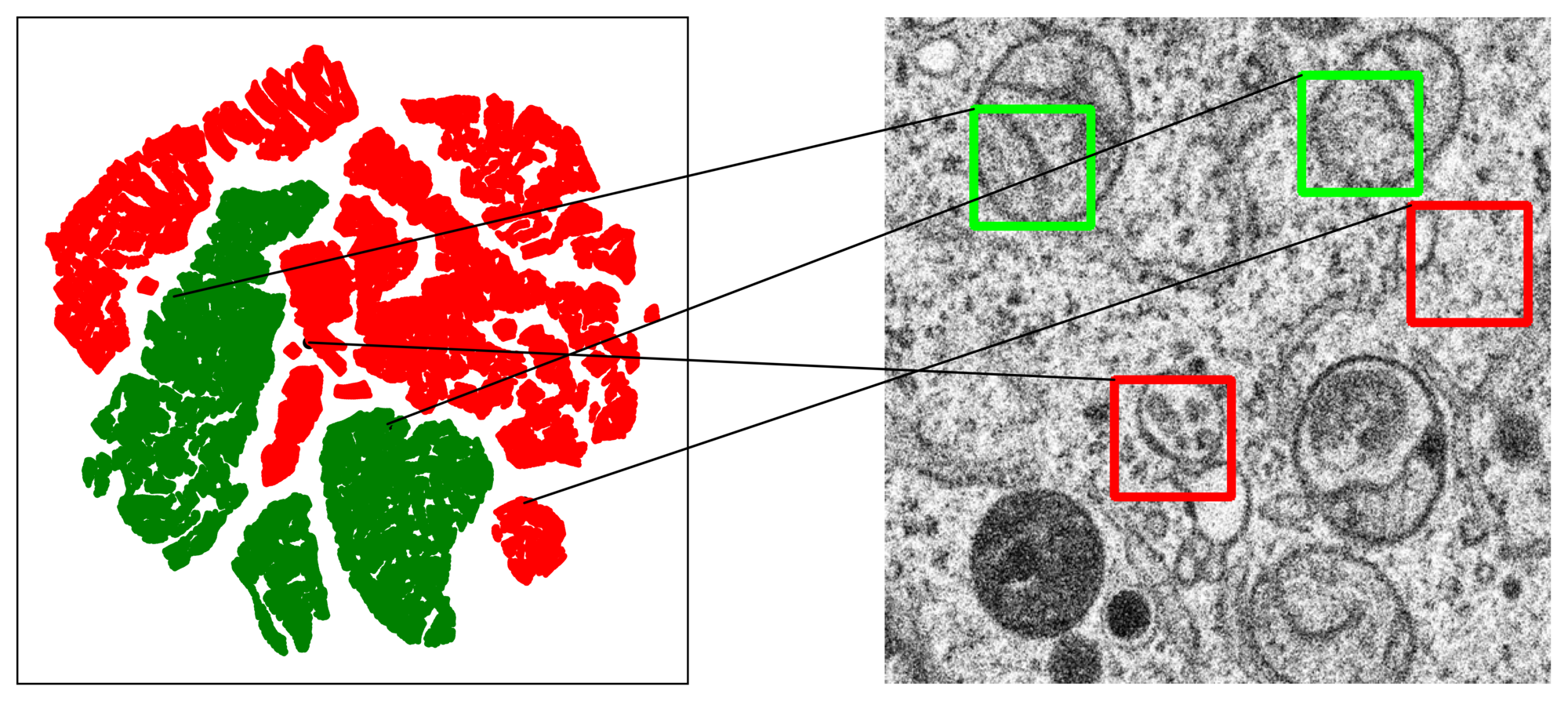}%
		\label{fig:clustering-fin-ae}%
	}
	\\
	\subfloat[Pretrained variational autoencoder.]{%
		\includegraphics[width=0.49\textwidth]{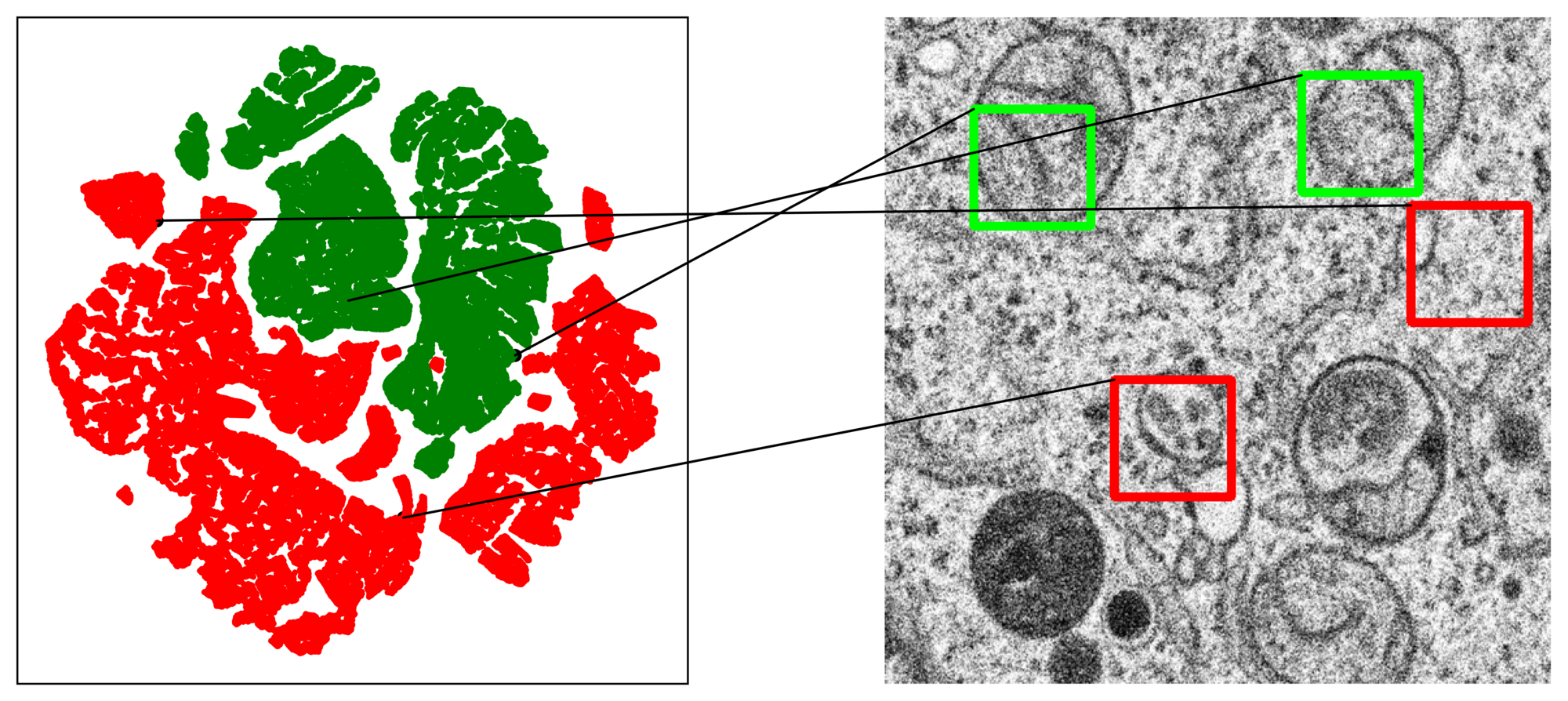}%
		\label{fig:clustering-vae}%
	}
	\hfill
	\subfloat[Fine-tuned variational autoencoder.]{%
		\includegraphics[width=0.49\textwidth]{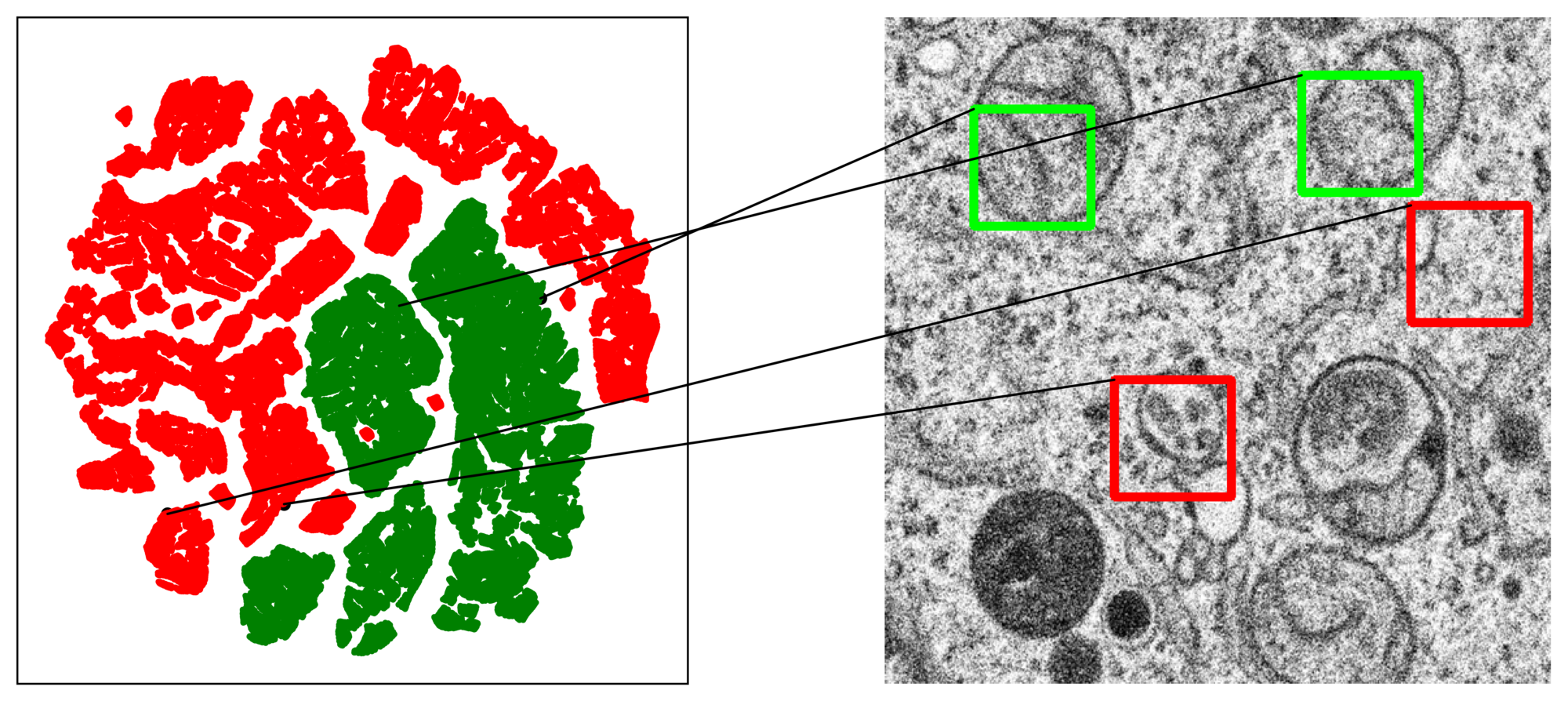}%
		\label{fig:clustering-fin-vae}%
	}
	\caption{t-SNE visualization of mitochondria and other patches from the \EMBL dataset for pretrained and fine-tuned classical autoencoder (\Cref{fig:clustering-ae,fig:clustering-fin-ae}) and variational autoencoder (\Cref{fig:clustering-vae,fig:clustering-fin-vae}, $\beta=10^{-5}$). Mitochondria patches ($> 50~\%$ overlap) in green, other patches ($0~\%$ overlap) in red.}
	\label{fig:clustering}
\end{figure*}

We compare the performance of the pretrained models, trained on generic images, and the fine-tuned models adapted to vEM datasets. The result is shown in \Cref{fig:clustering} for both the classical and variational autoencoder. We consider patches with at least 50~\% of their pixels corresponding to a mitochondrion as positive patches and patches with no overlap as negative patches, taken from 10 slices of the {\EMBL} dataset. The patches are of size $100 \times 100$, taken with a stride of $2$. We apply $1000$ iterations of the t-SNE procedure.

In \Cref{fig:clustering-ae}, the pretrained AE shows clear separation between positive and negative regions; however, the regions are intermingled with little apparent structure. For the fine-tuned AE (\Cref{fig:clustering-fin-ae}), we observe perfect separation with subregions appearing in closer proximity, revealing clearer underlying structure.

For the pretrained VAE (\Cref{fig:clustering-vae}), we observe near-perfect separation with a more structured latent space compared to the pretrained AE, though a small negative region is surrounded by positive regions. The fine-tuned VAE (\Cref{fig:clustering-fin-vae}) shows similar results, though a small negative region is surrounded by positive regions. From these results, we conclude that both autoencoders produce local image descriptors that discriminate between distinct structures, and fine-tuning improves performance.

\subsubsection{Local vs. global descriptor performance}

\begin{figure*}
	\centering
	\subfloat[All slices (0 -- 63).]{%
		\includegraphics[width=0.32\textwidth, clip]{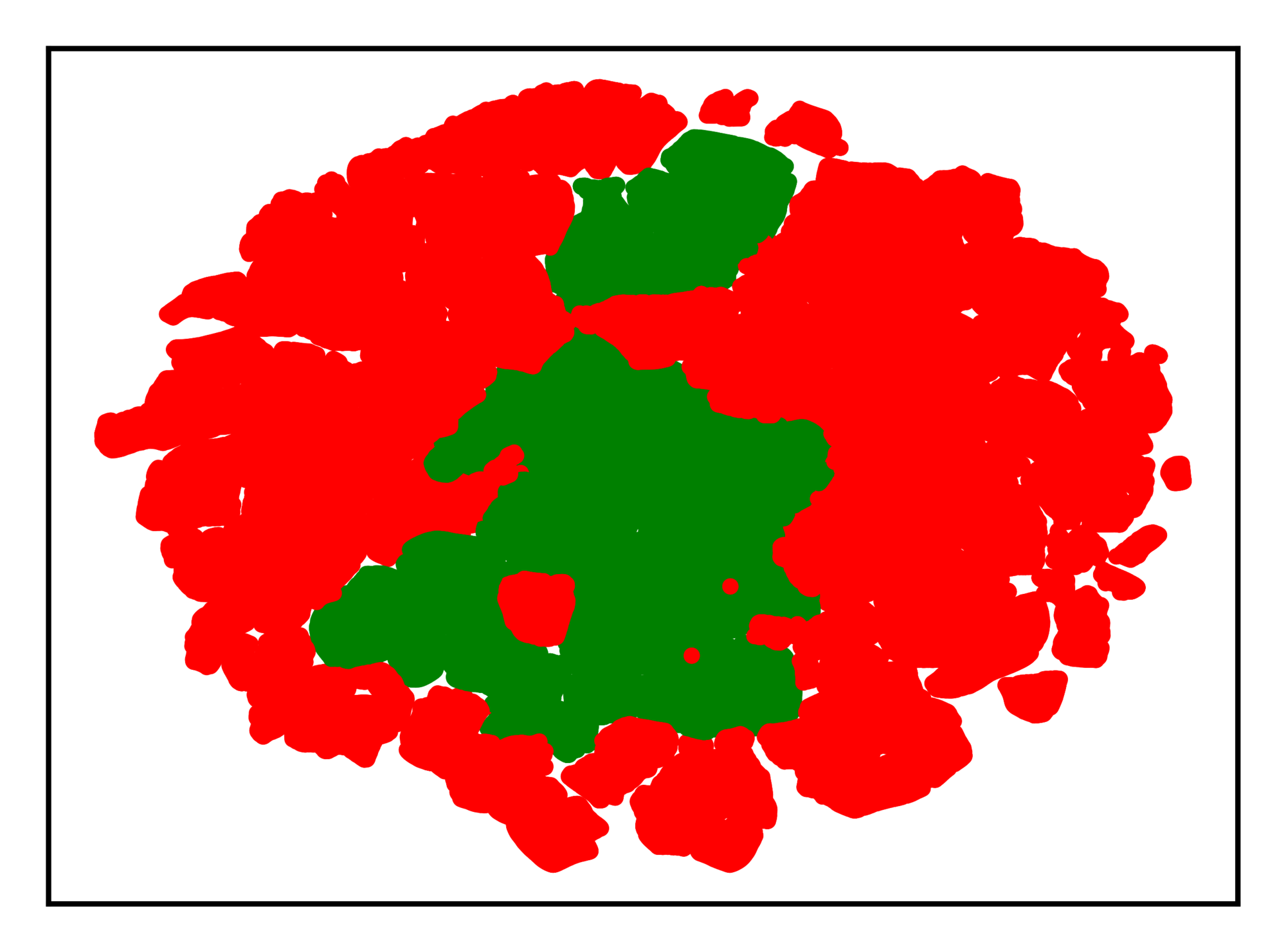}%
		\label{fig:ae-all}%
	}
	\\
	\subfloat[Slices 0 -- 9.]{%
		\includegraphics[width=0.23\textwidth, clip]{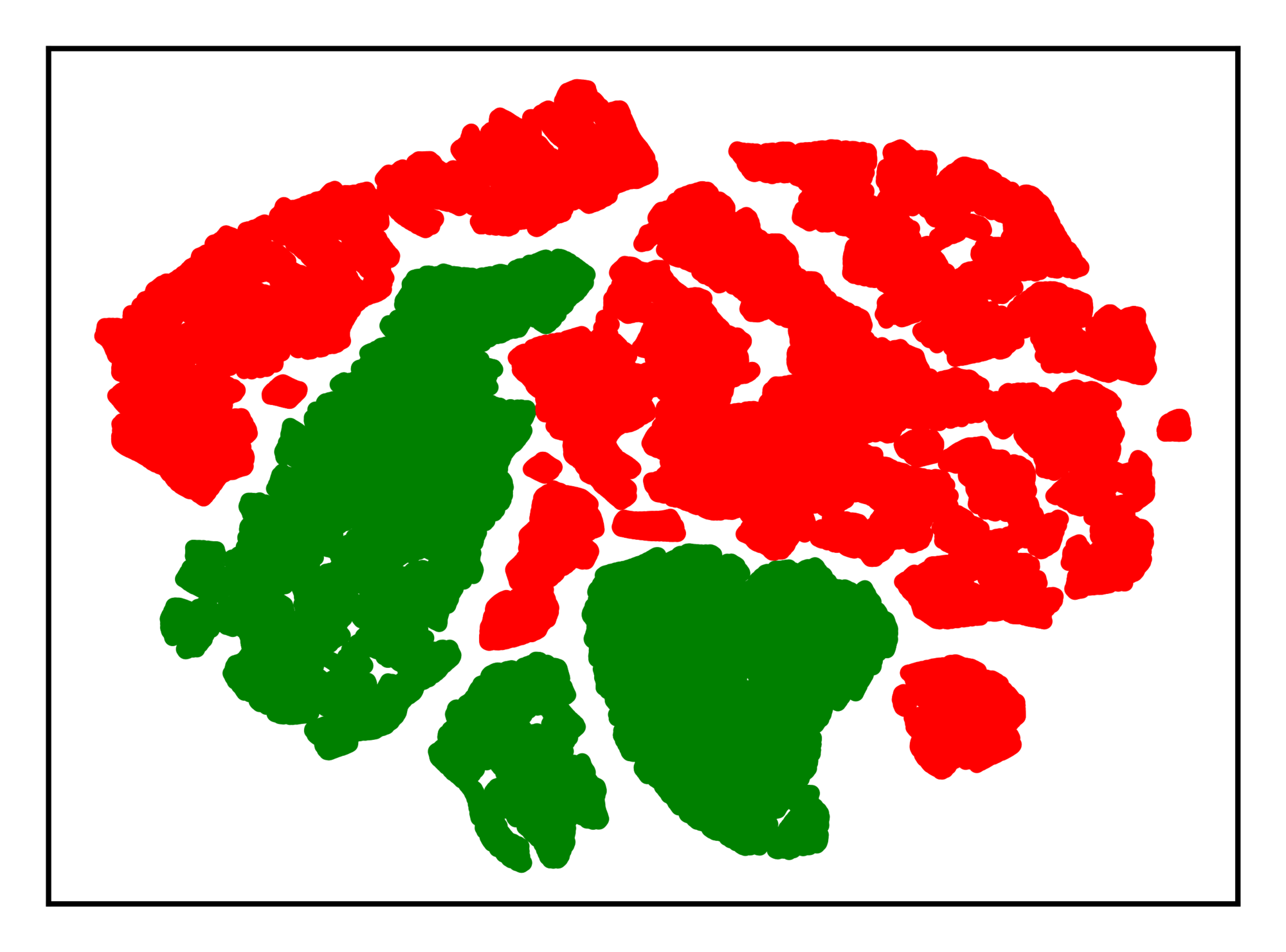}%
		\label{fig:ae-0-9}%
	}
	\subfloat[Slices 5 -- 14.]{%
		\includegraphics[width=0.23\textwidth, clip]{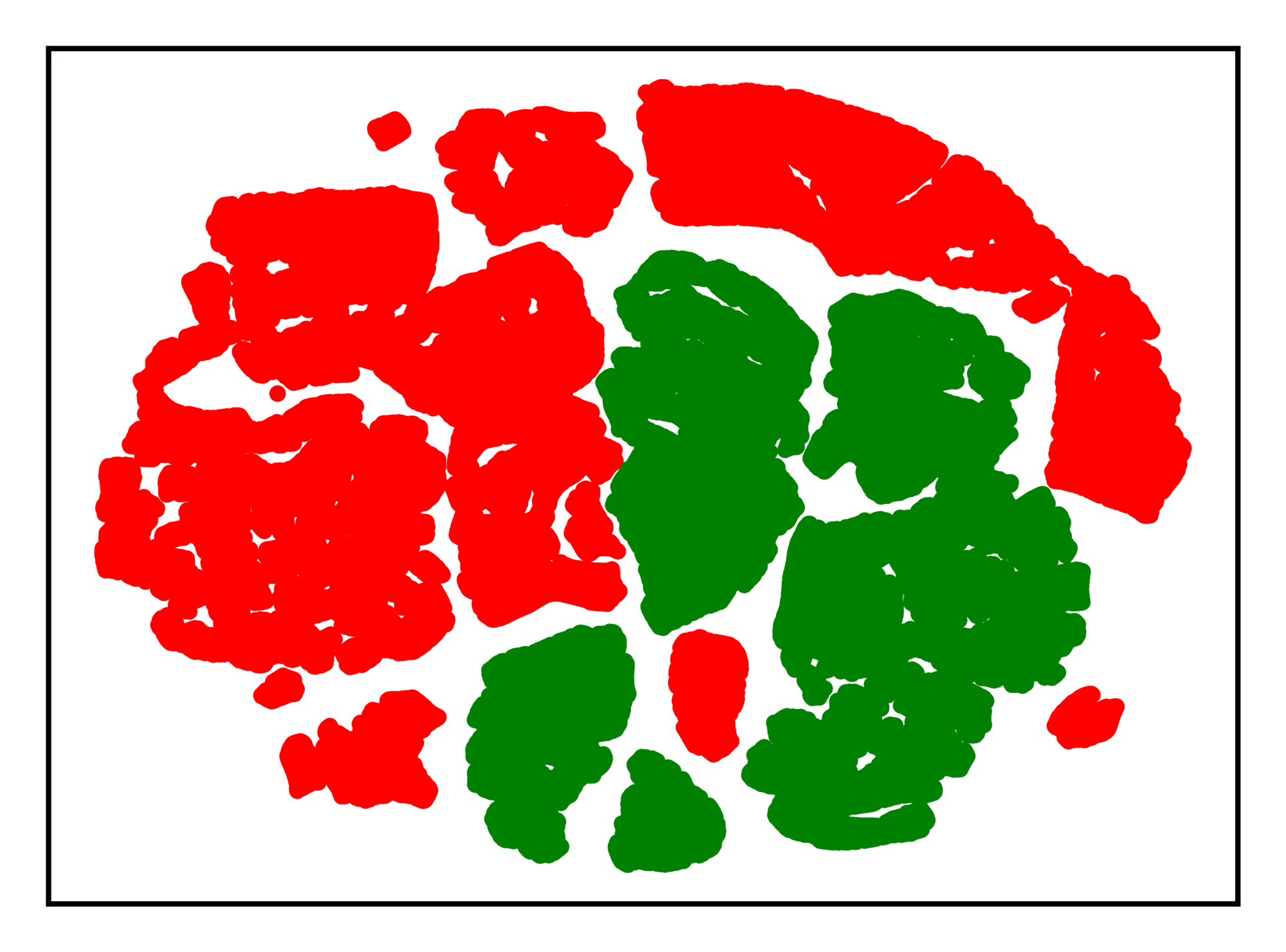}%
		\label{fig:ae-5-14}%
	}
	\subfloat[Slices 10 -- 19.]{%
		\includegraphics[width=0.23\textwidth, clip]{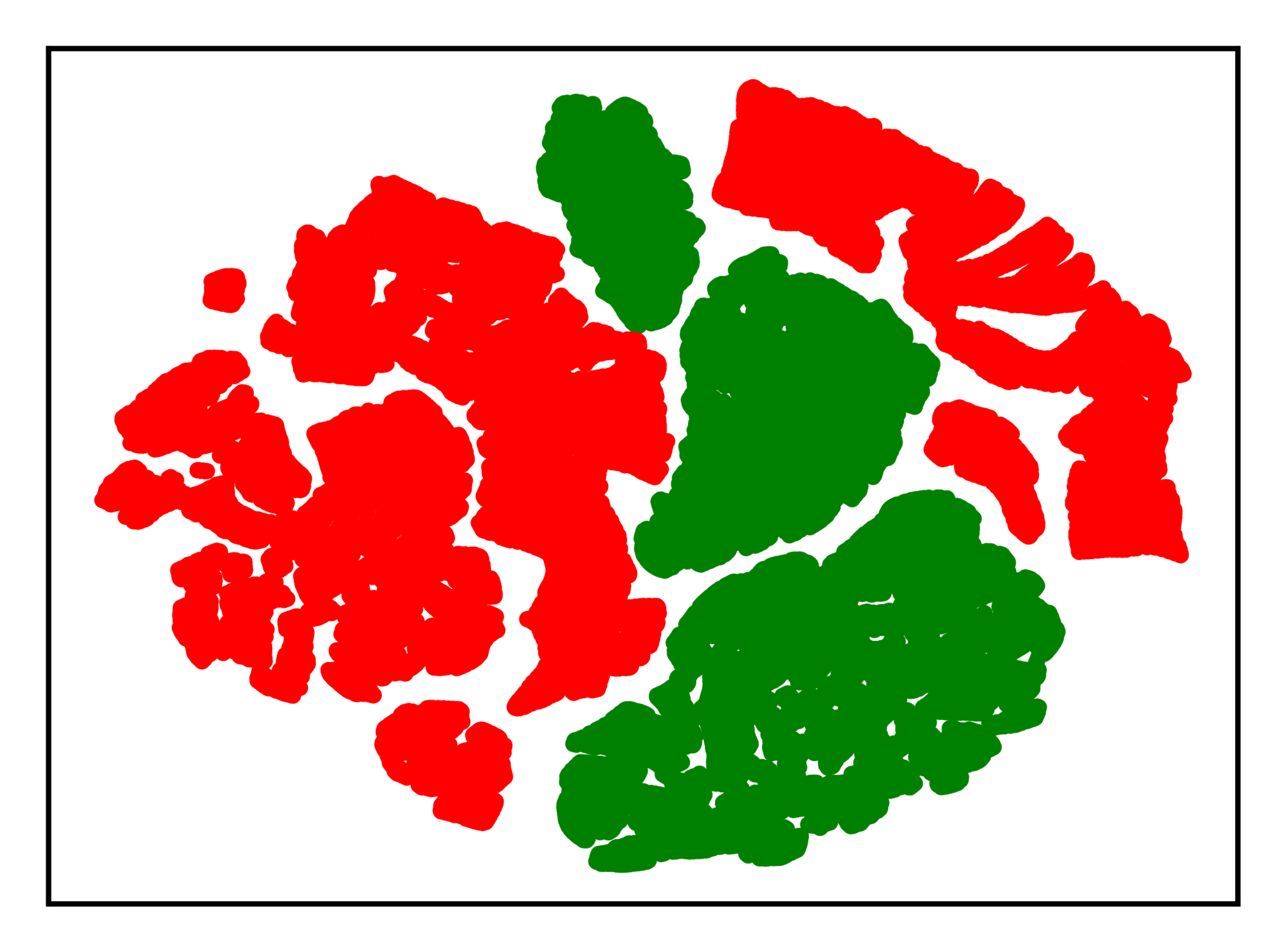}%
		\label{fig:ae-10-19}%
	}\\
	\subfloat[Slices 15 -- 24.]{%
		\includegraphics[width=0.23\textwidth, clip]{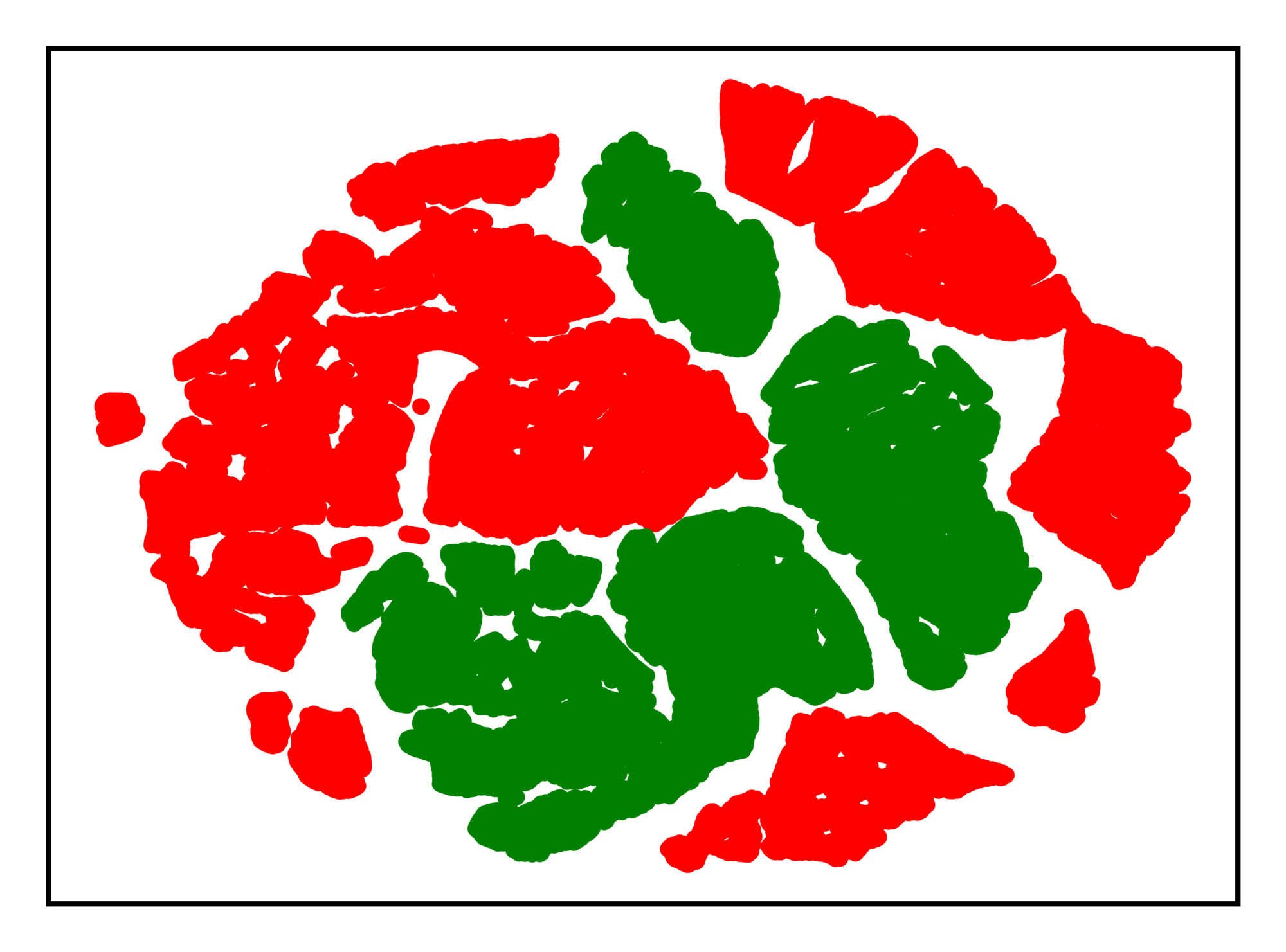}%
		\label{fig:ae-15-24}%
	}
	\subfloat[Slices 20 -- 29.]{%
		\includegraphics[width=0.23\textwidth, clip]{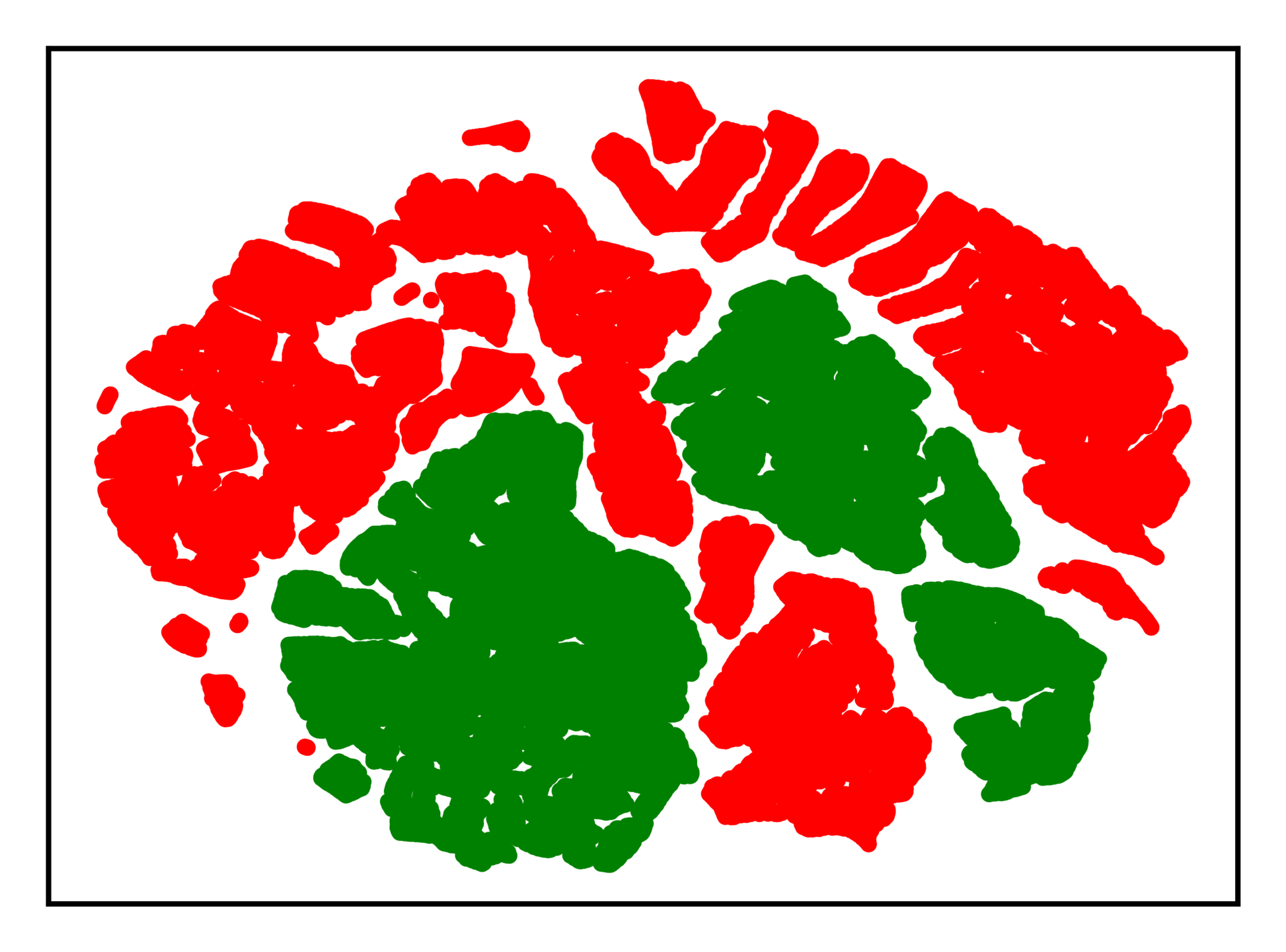}%
		\label{fig:ae-20-29}%
	}
	\subfloat[Slices 25 -- 34.]{%
		\includegraphics[width=0.23\textwidth, clip]{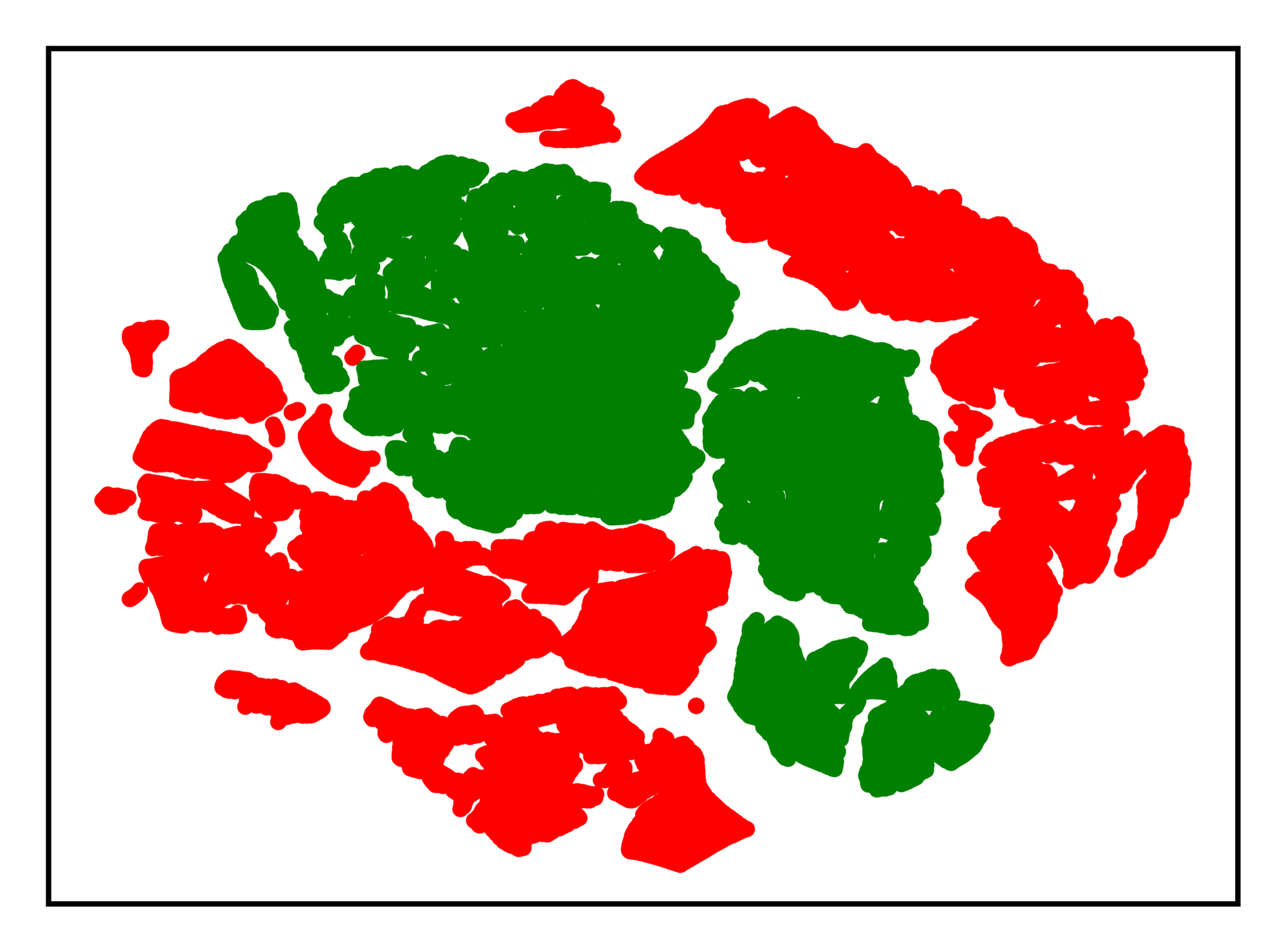}%
		\label{fig:ae-25-34}%
	}\\
	\subfloat[Slices 30 -- 39.]{%
		\includegraphics[width=0.23\textwidth, clip]{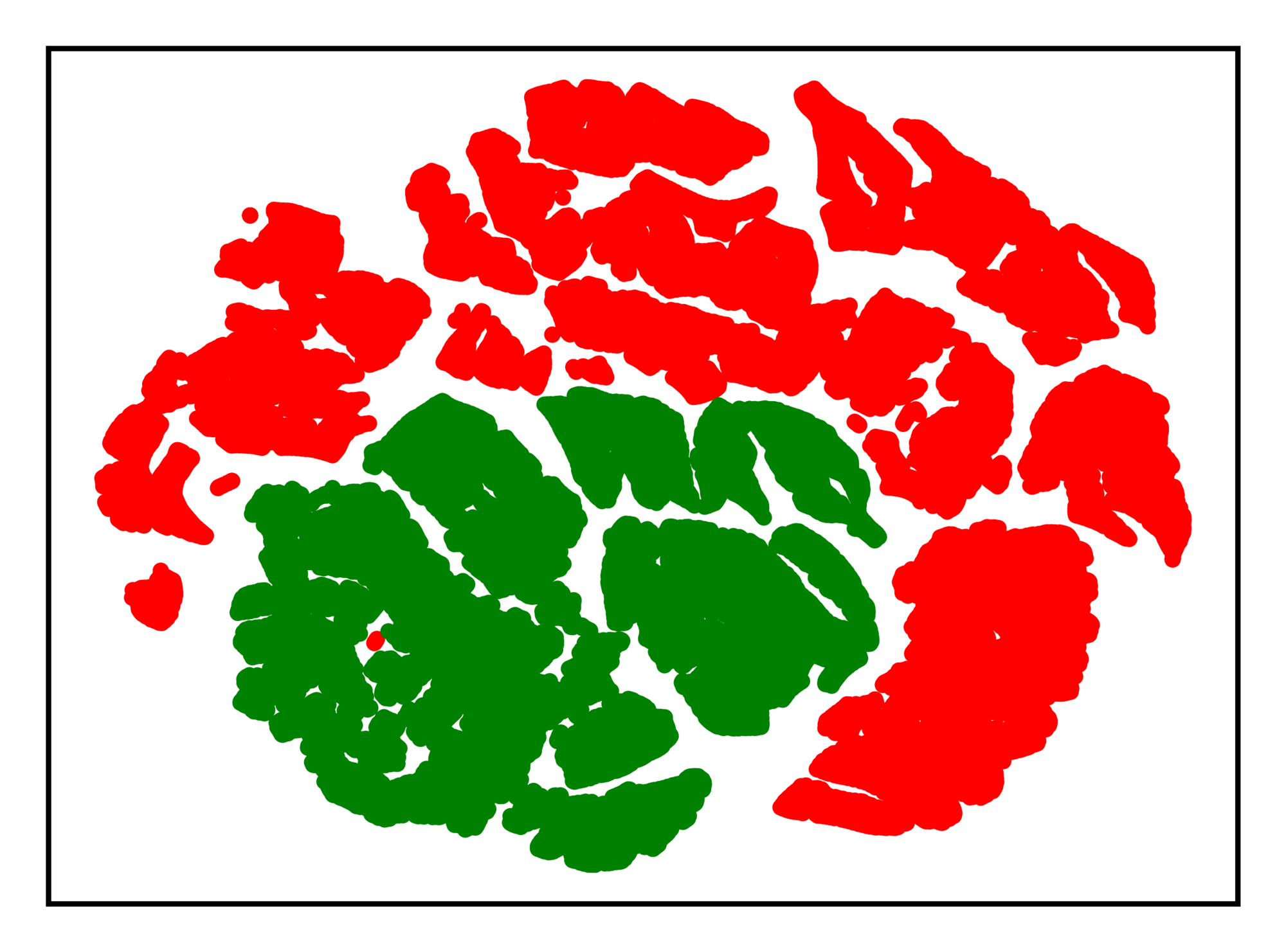}%
		\label{fig:ae-30-39}%
	}
	\subfloat[Slices 35 -- 44.]{%
		\includegraphics[width=0.23\textwidth, clip]{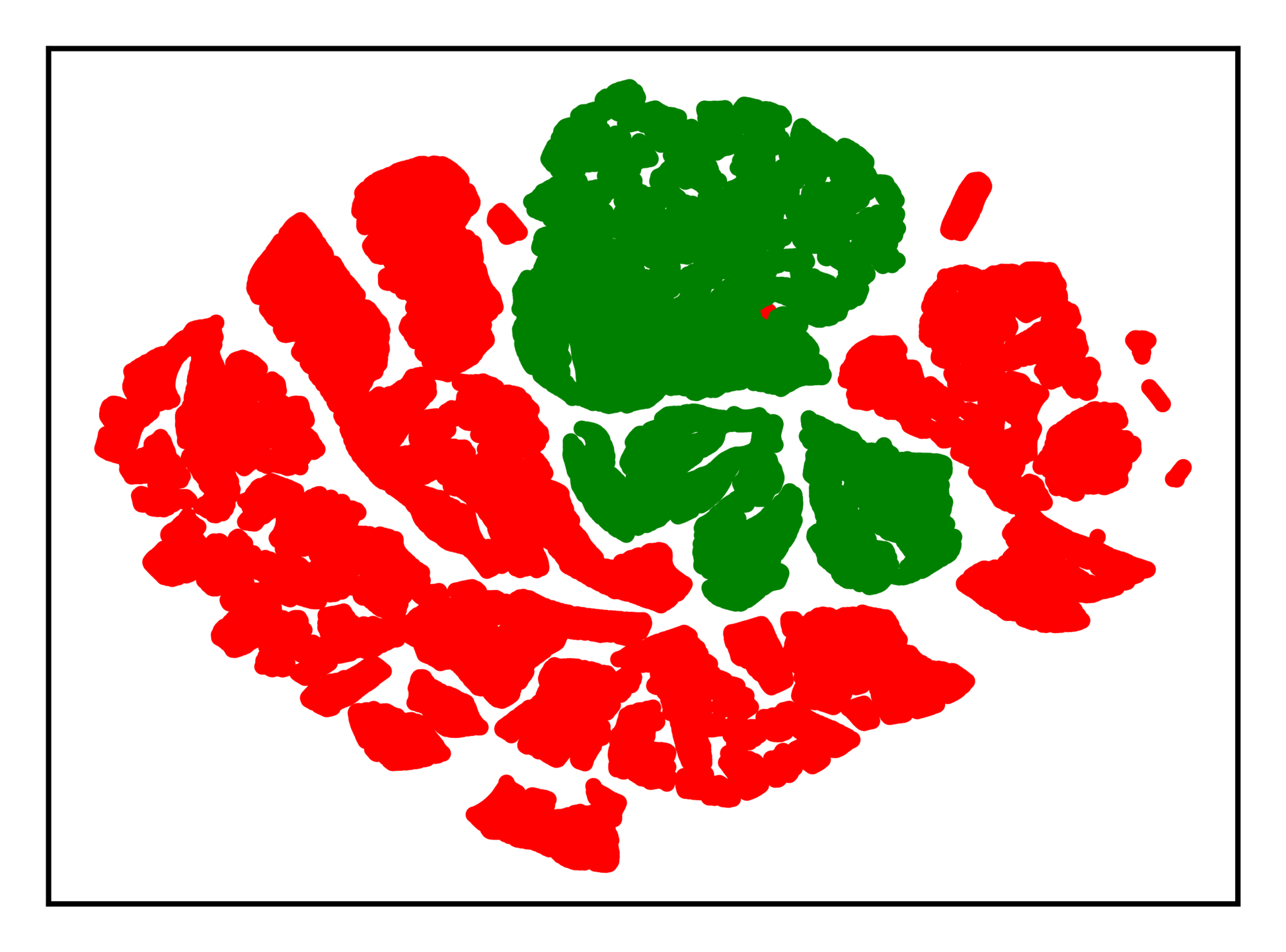}%
		\label{fig:ae-35-44}%
	}
	\subfloat[Slices 40 -- 49.]{%
		\includegraphics[width=0.23\textwidth, clip]{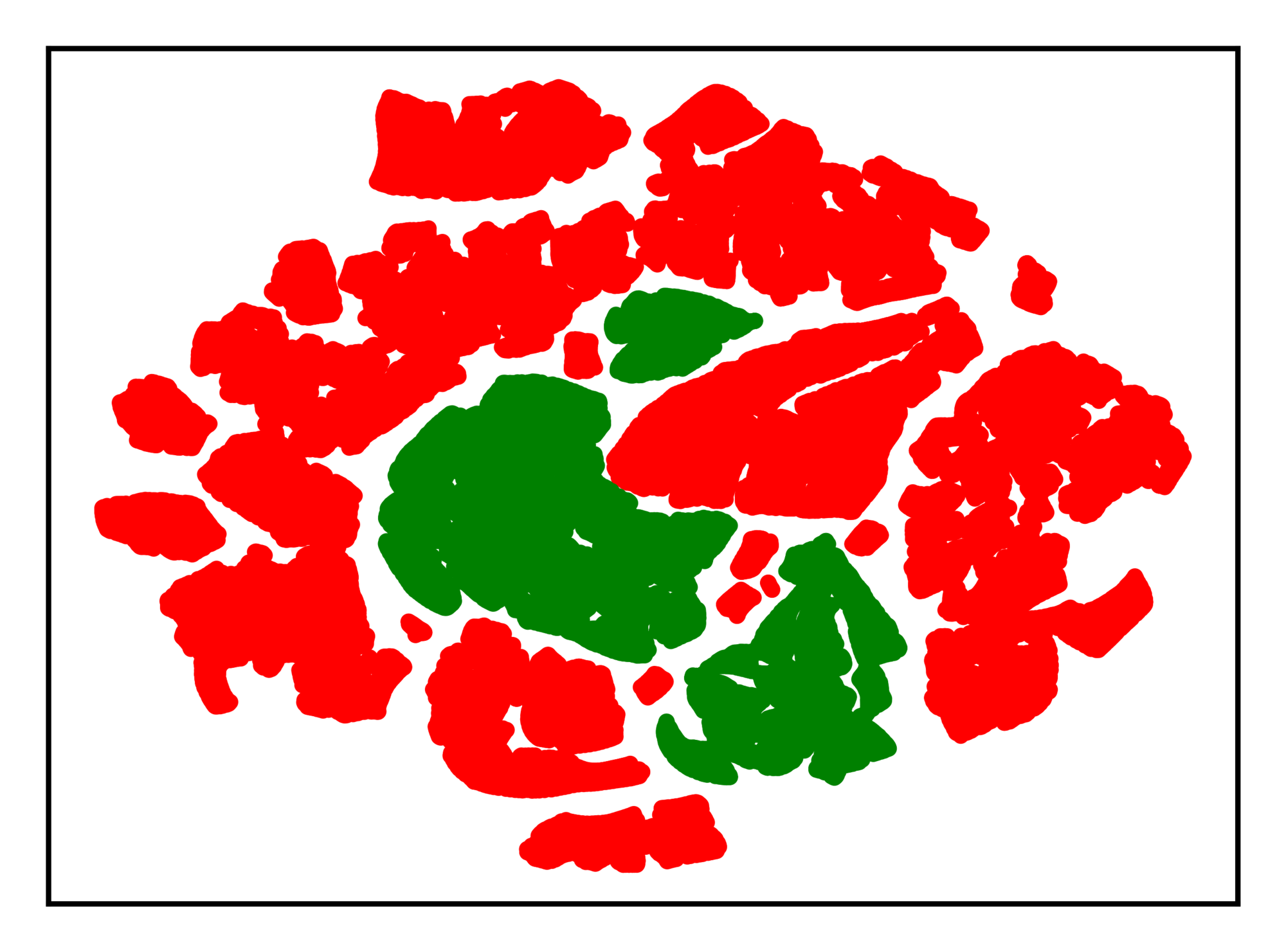}%
		\label{fig:ae-40-49}%
	}\\
	\subfloat[Slices 45 -- 54.]{%
		\includegraphics[width=0.23\textwidth, clip]{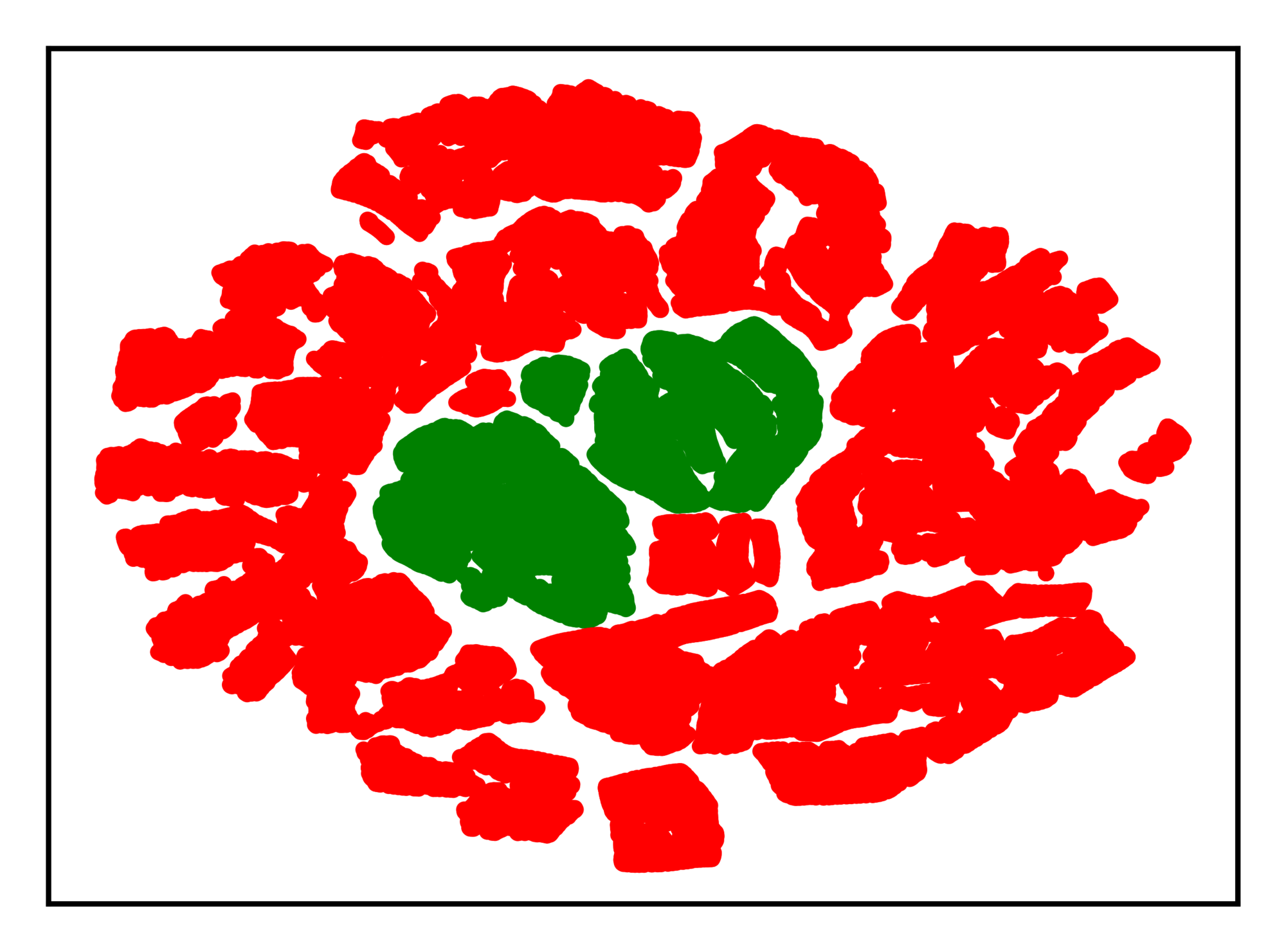}%
		\label{fig:ae-45-54}%
	}
	\subfloat[Slices 50 -- 59.]{%
		\includegraphics[width=0.23\textwidth, clip]{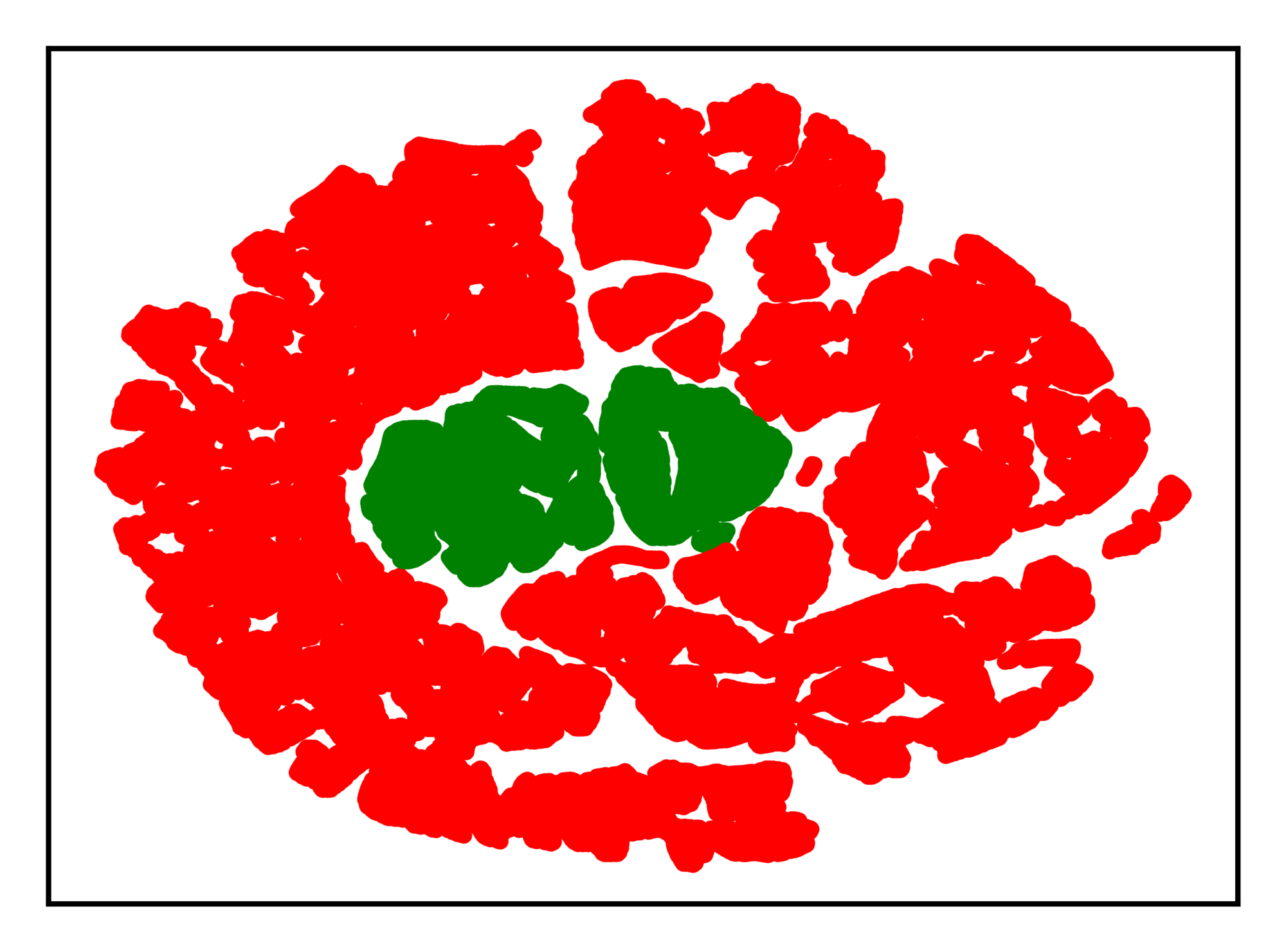}%
		\label{fig:ae-50-59}%
	}
	\subfloat[Slices 55 -- 63.]{%
		\includegraphics[width=0.23\textwidth, clip]{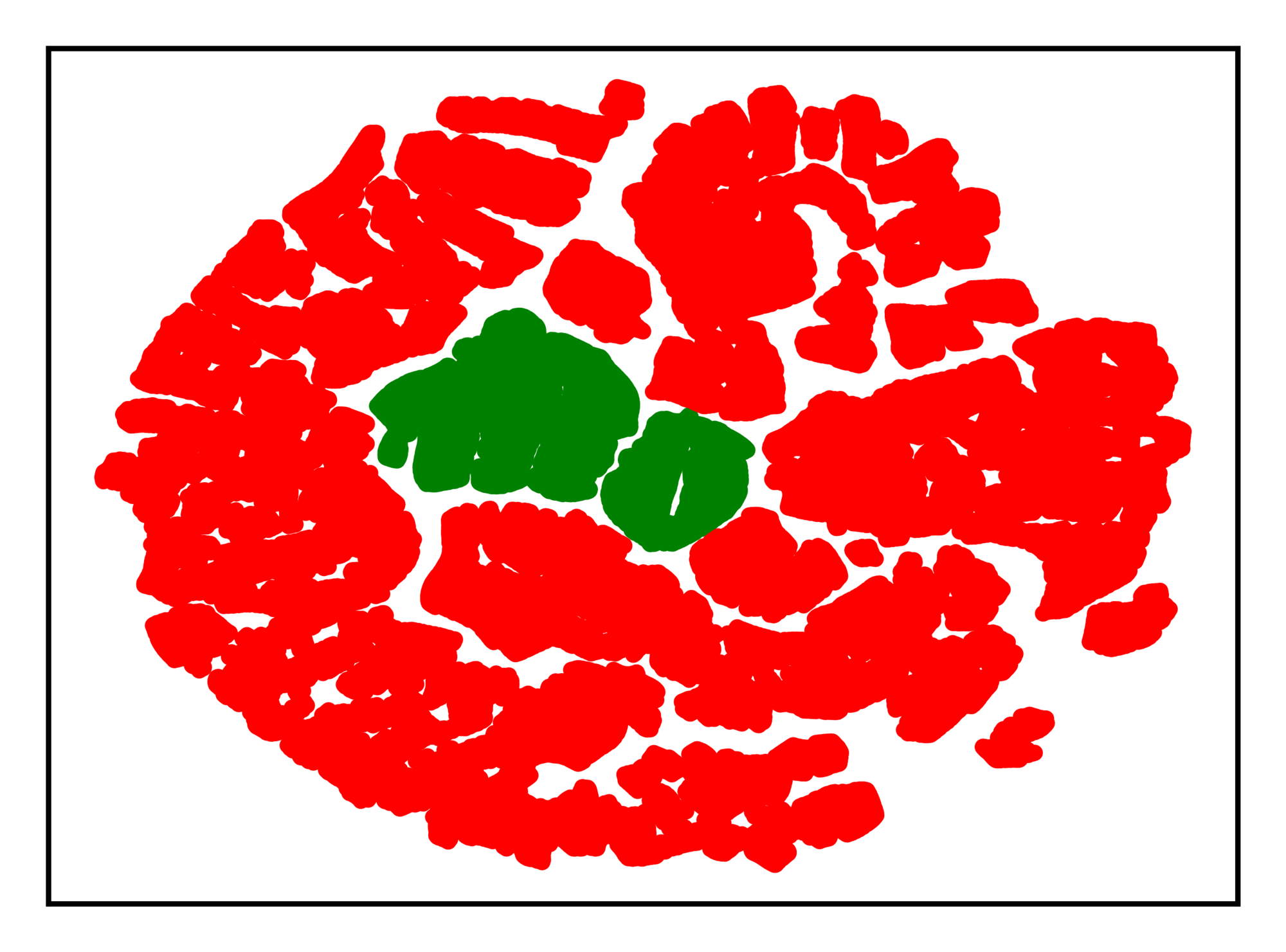}%
		\label{fig:ae-55-64}%
	}
	\caption{t-SNE visualization of mitochondria and other patches for the classical autoencoder across groups of 10 slices (\EMBL dataset). Mitochondria patches ($> 50~\%$ overlap) in green, other patches ($0~\%$ overlap) in red.}
	\label{fig:autoencoder-clusters}
\end{figure*}

\begin{figure*}
	\centering
	\subfloat[All slices (0 -- 63).]{%
		\includegraphics[width=0.32\textwidth, clip]{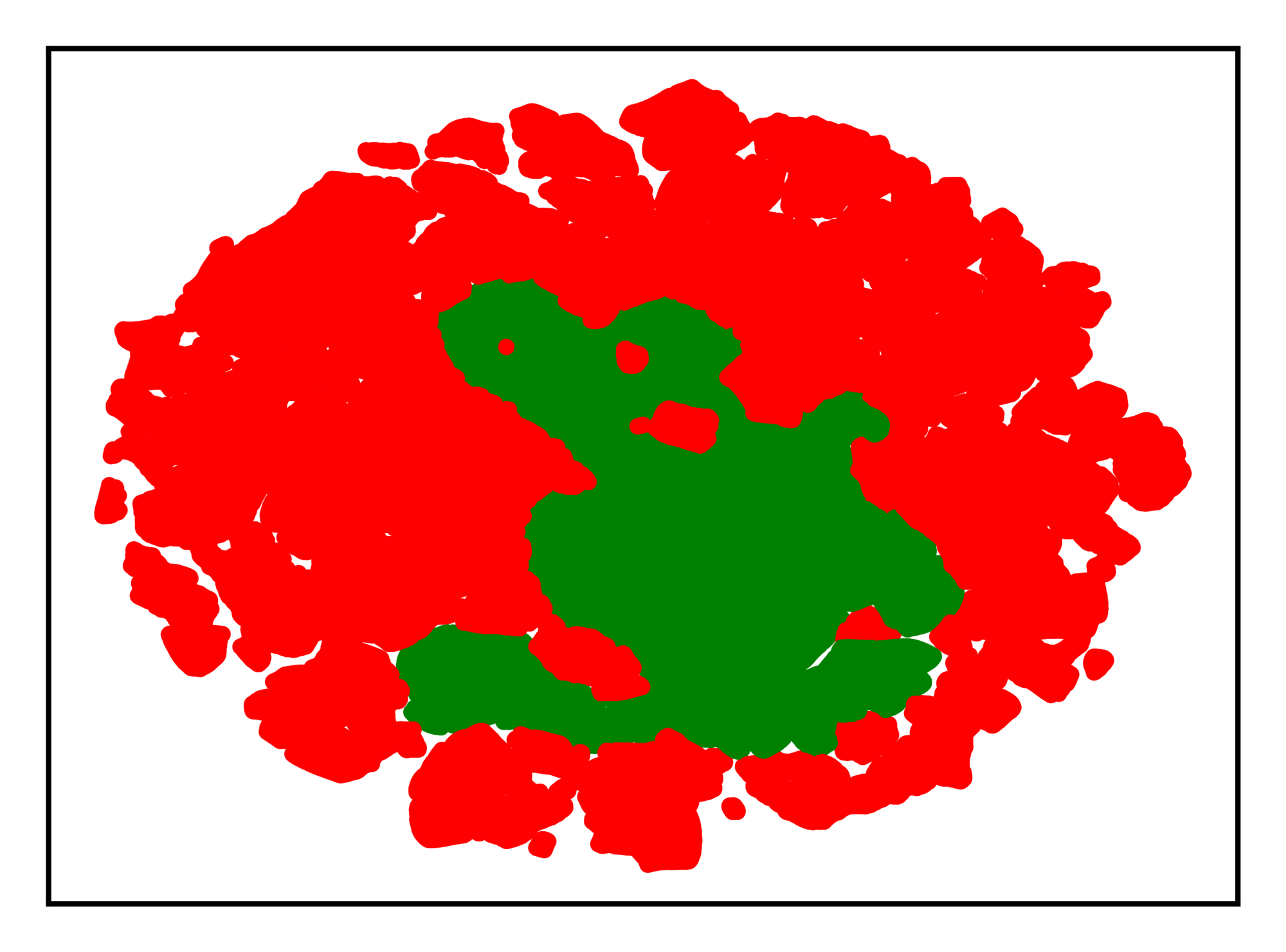}%
		\label{fig:vae-all}%
	}
	\\
	\subfloat[Slices 0 -- 9.]{%
		\includegraphics[width=0.23\textwidth, clip]{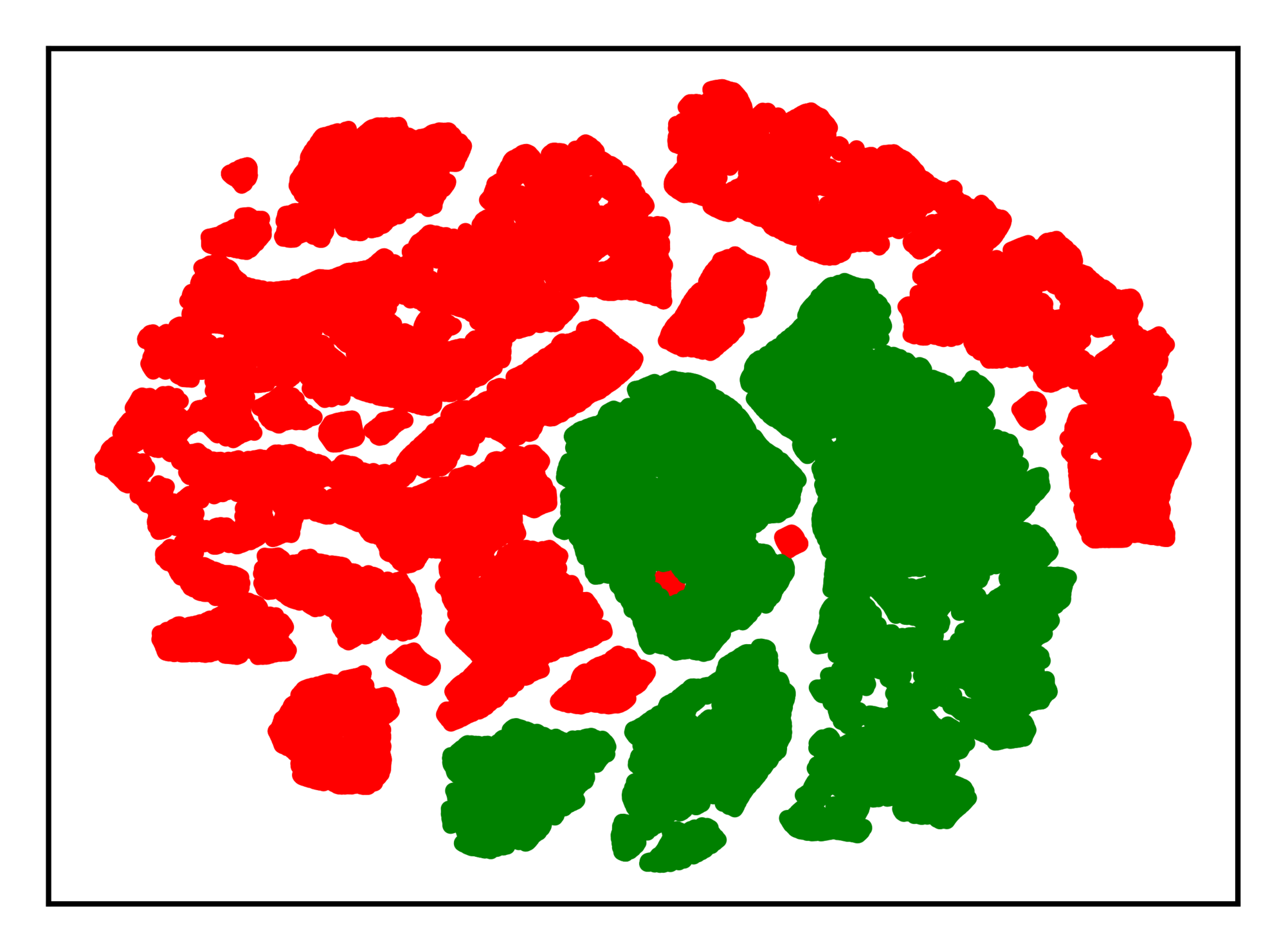}%
		\label{fig:vae-0-9}%
	}
	\subfloat[Slices 5 -- 14.]{%
		\includegraphics[width=0.23\textwidth, clip]{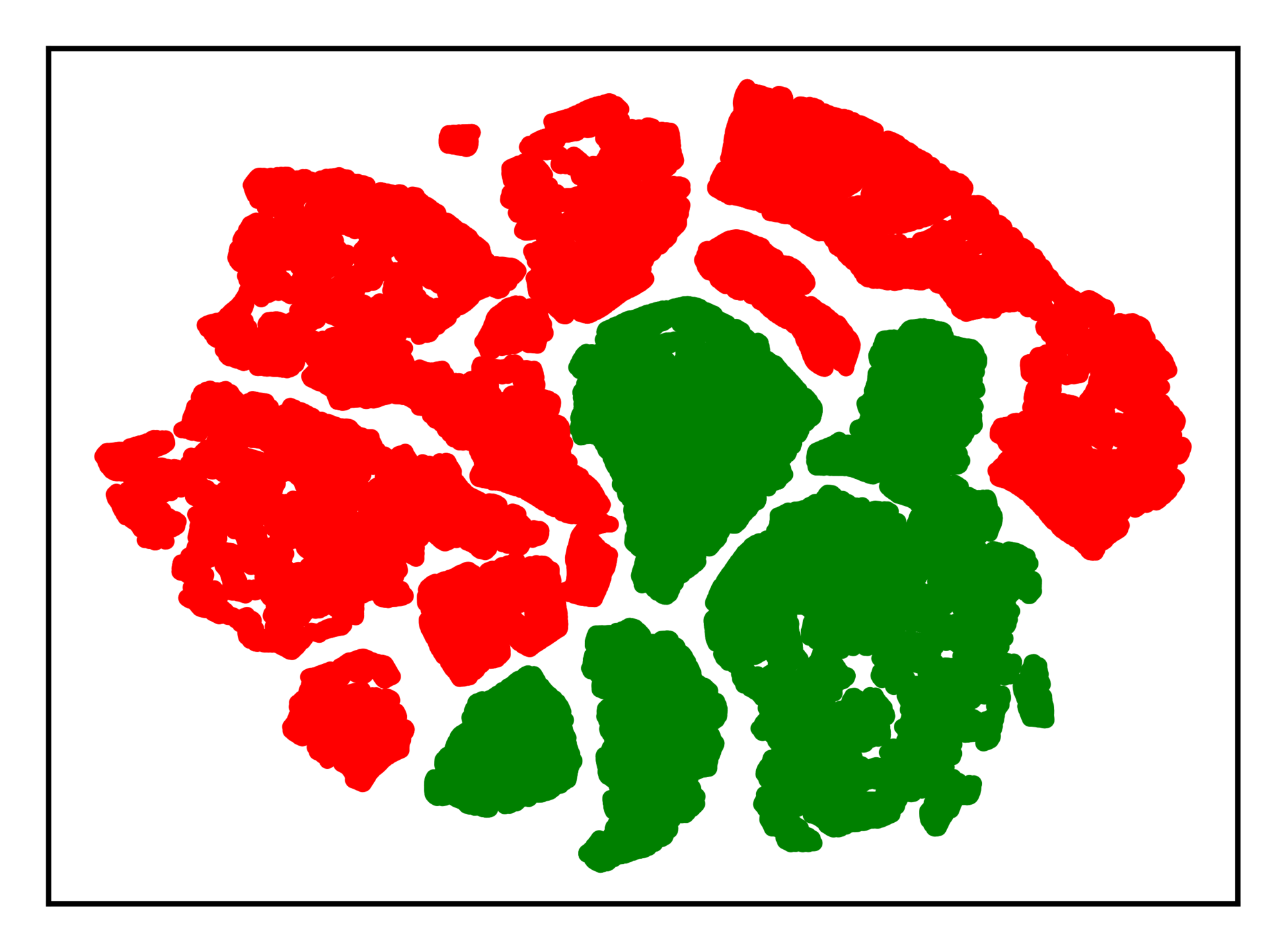}%
		\label{fig:vae-5-14}%
	}
	\subfloat[Slices 10 -- 19.]{%
		\includegraphics[width=0.23\textwidth, clip]{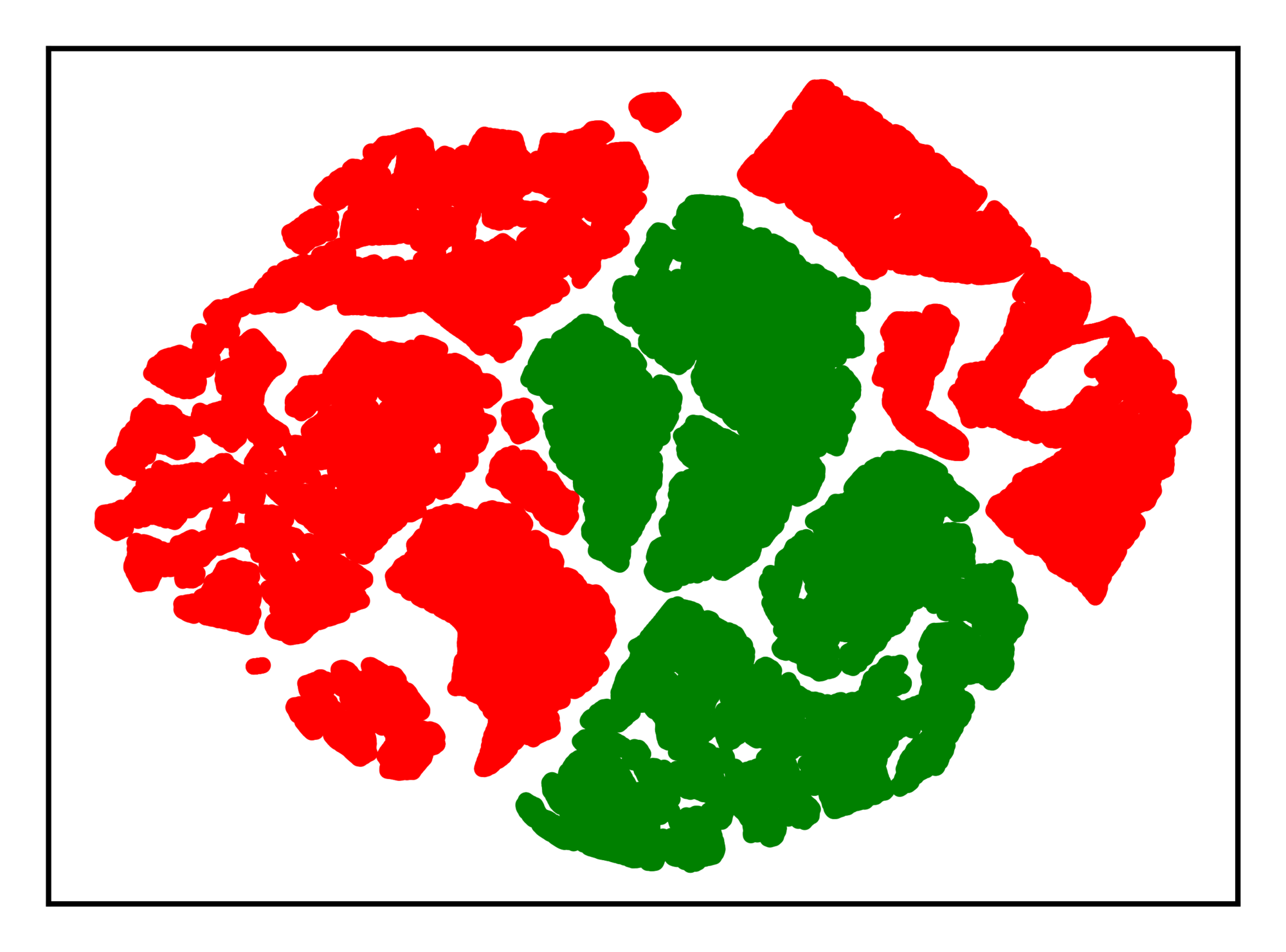}%
		\label{fig:vae-10-19}%
	}\\
	\subfloat[Slices 15 -- 24.]{%
		\includegraphics[width=0.23\textwidth, clip]{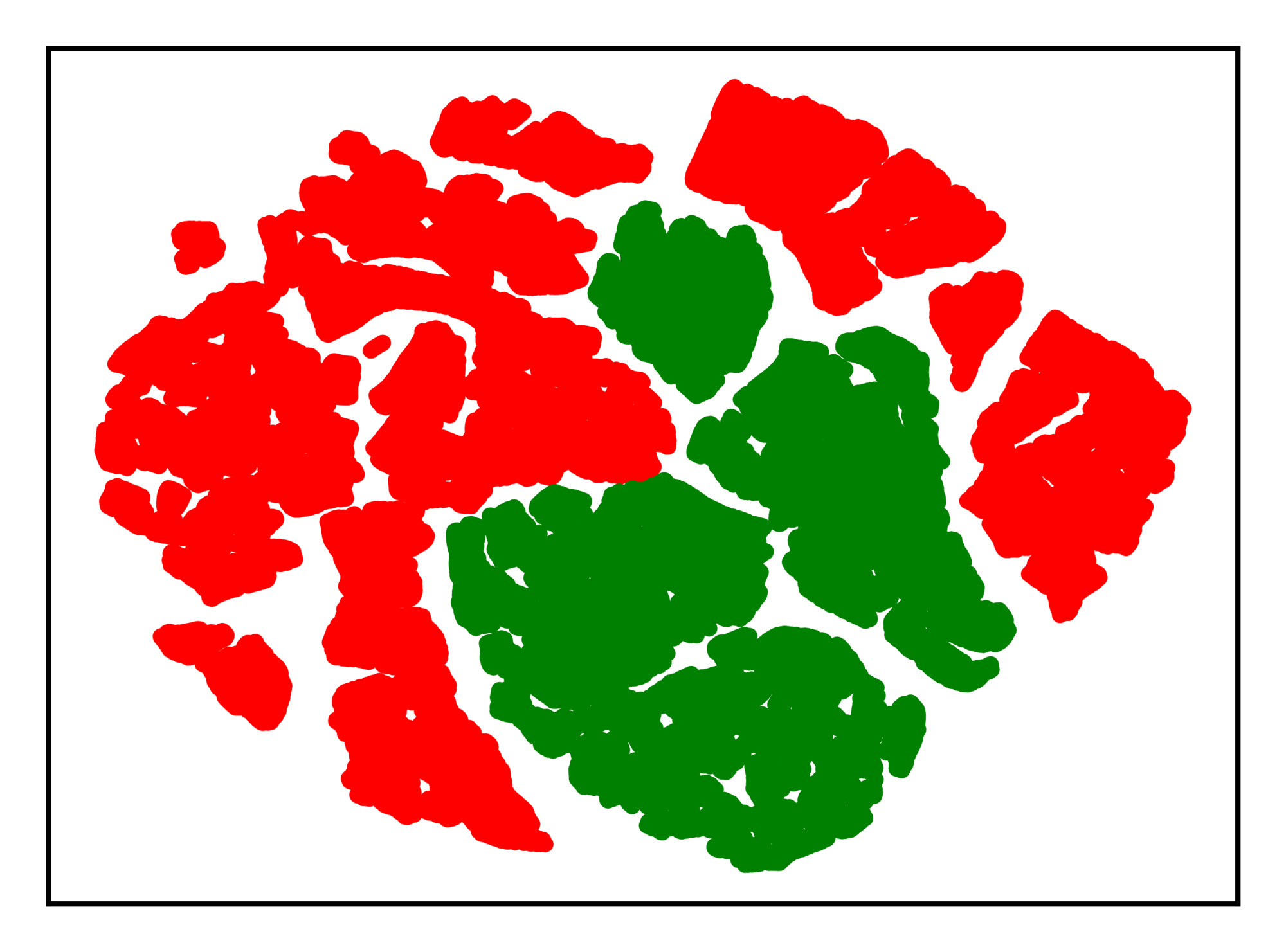}%
		\label{fig:vae-15-24}%
	}
	\subfloat[Slices 20 -- 29.]{%
		\includegraphics[width=0.23\textwidth, clip]{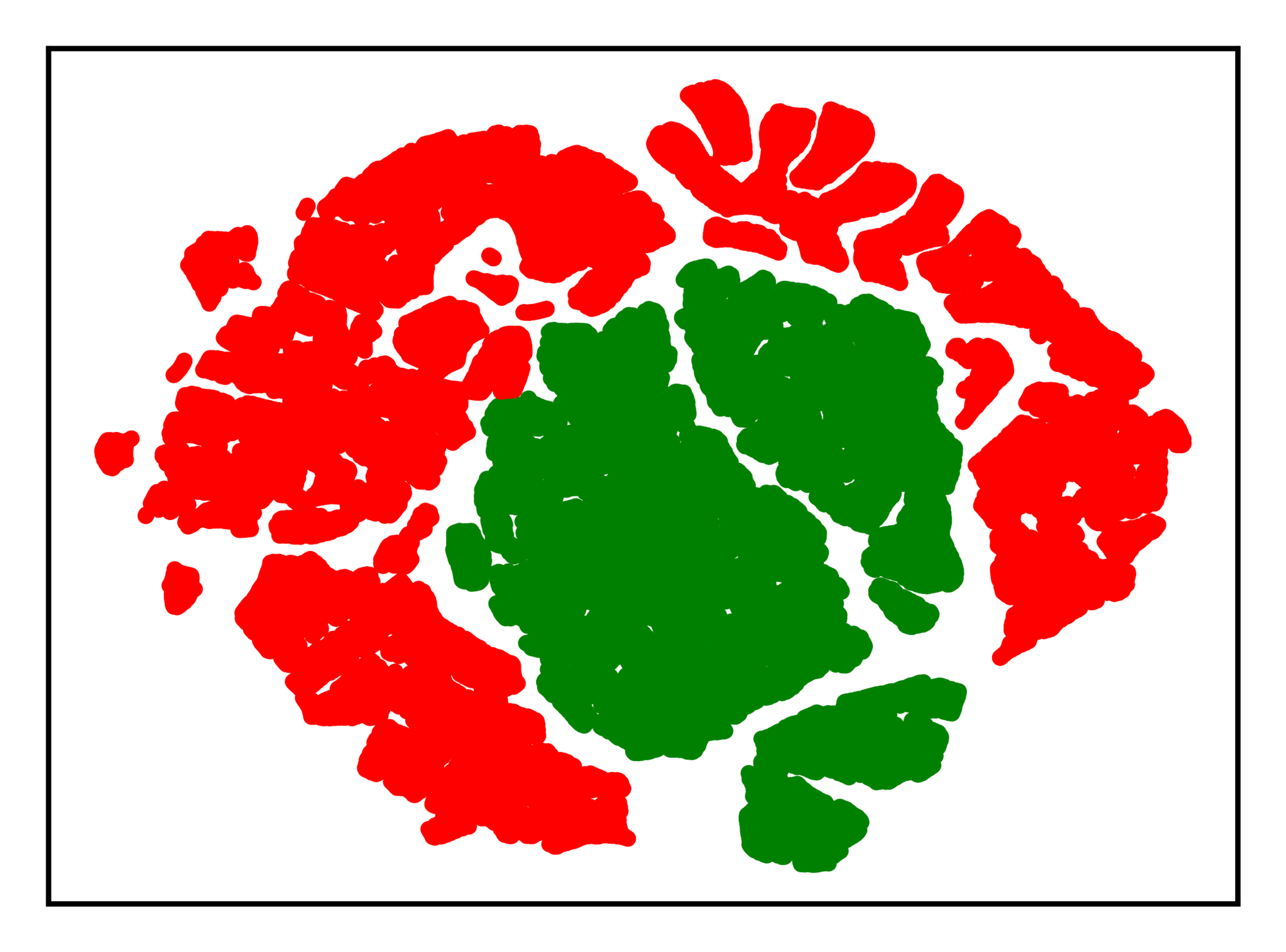}%
		\label{fig:vae-20-29}%
	}
	\subfloat[Slices 25 -- 34.]{%
		\includegraphics[width=0.23\textwidth, clip]{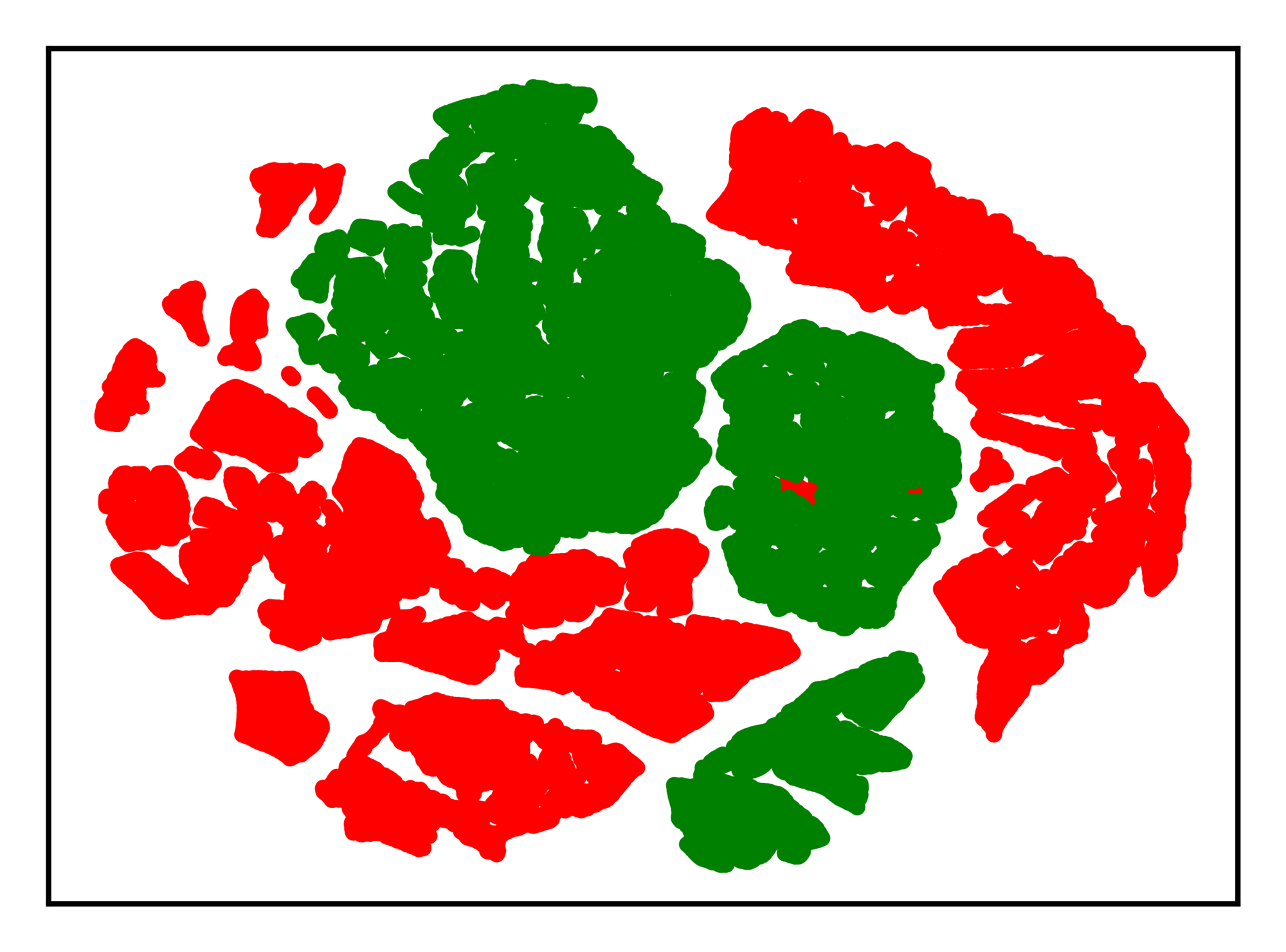}%
		\label{fig:vae-25-34}%
	}\\
	\subfloat[Slices 30 -- 39.]{%
		\includegraphics[width=0.23\textwidth, clip]{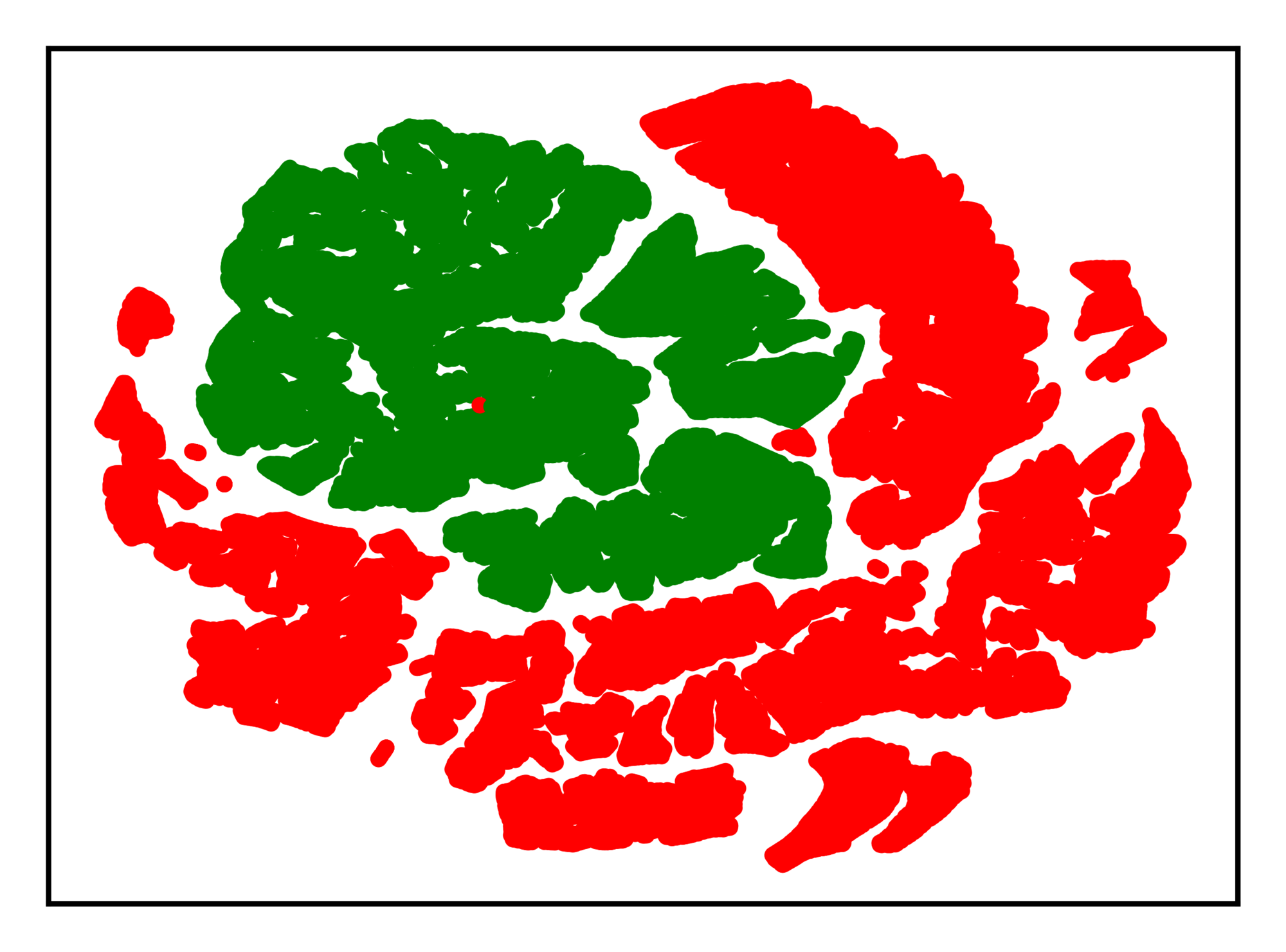}%
		\label{fig:vae-30-39}%
	}
	\subfloat[Slices 35 -- 44.]{%
		\includegraphics[width=0.23\textwidth, clip]{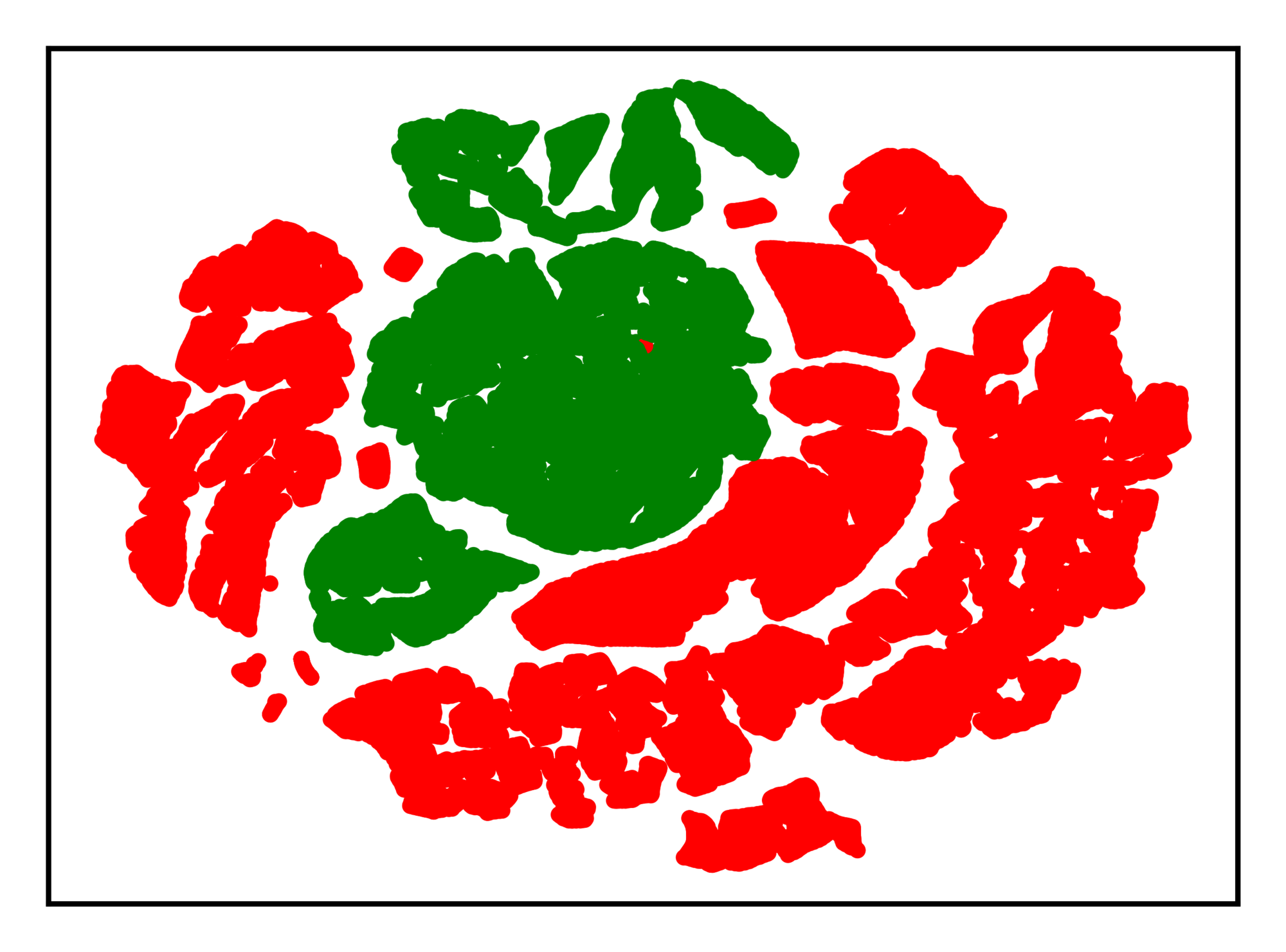}%
		\label{fig:vae-35-44}%
	}
	\subfloat[Slices 40 -- 49.]{%
		\includegraphics[width=0.23\textwidth, clip]{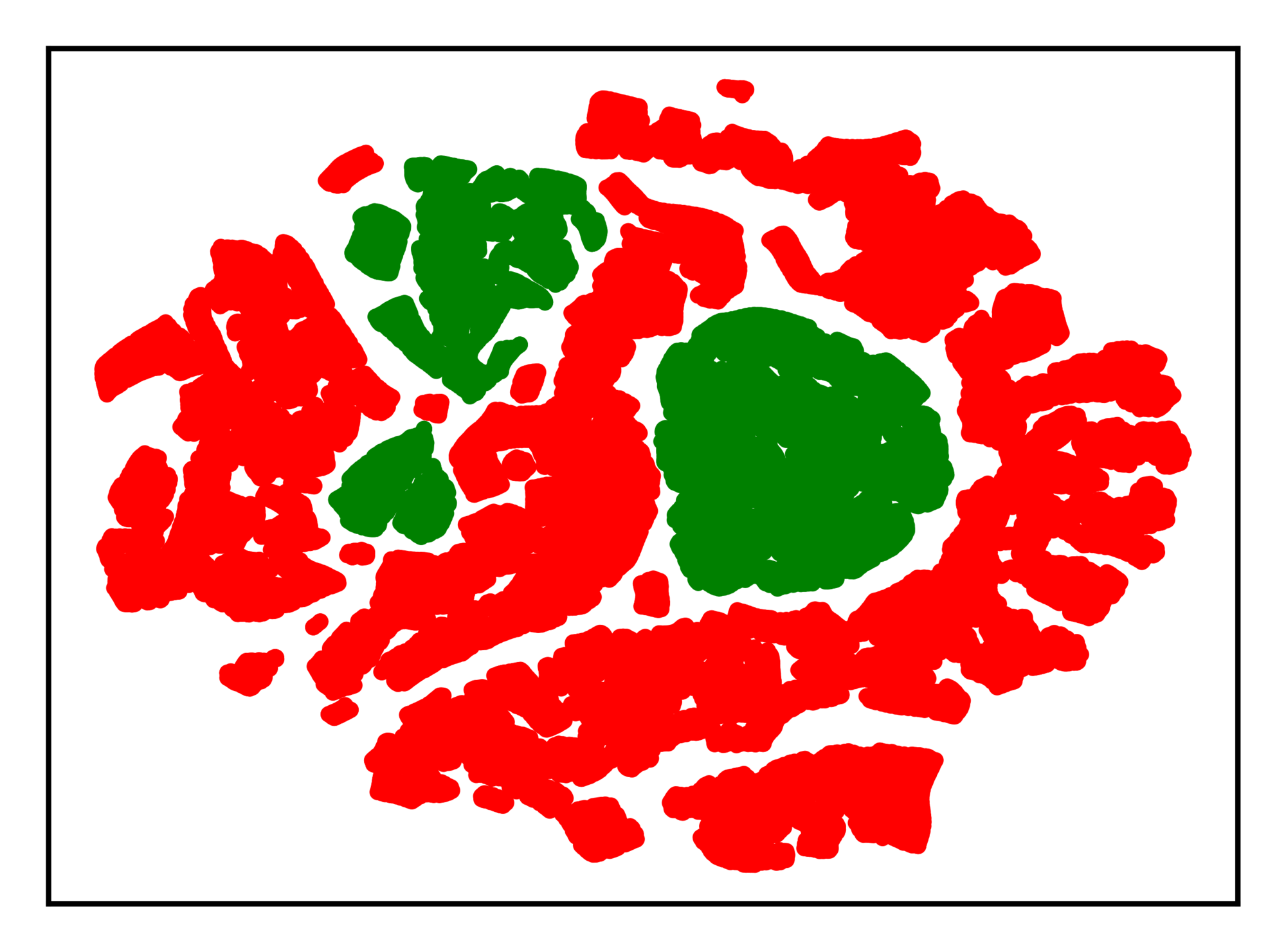}%
		\label{fig:vae-40-49}%
	}\\
	\subfloat[Slices 45 -- 54.]{%
		\includegraphics[width=0.23\textwidth, clip]{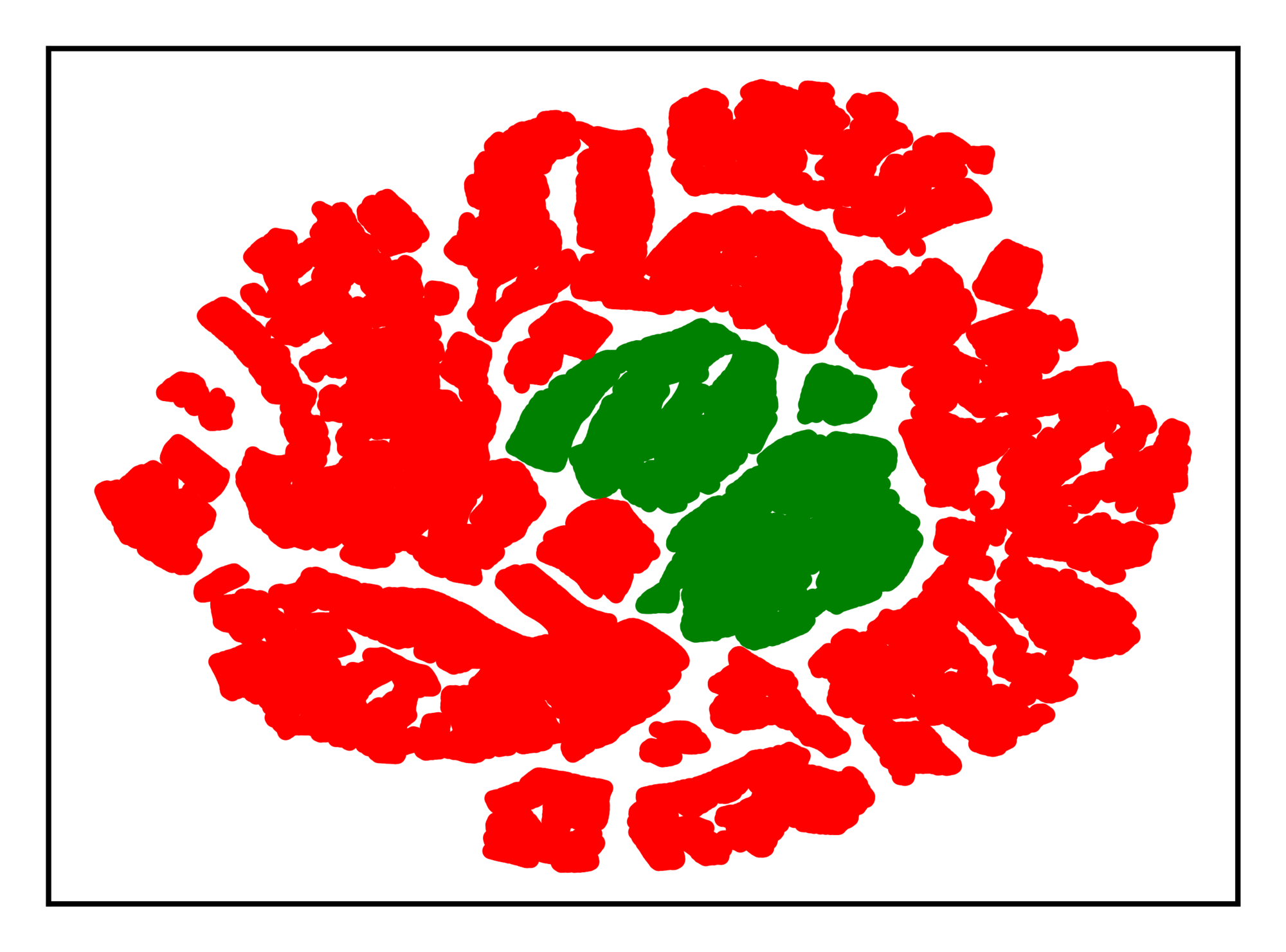}%
		\label{fig:vae-45-54}%
	}
	\subfloat[Slices 50 -- 59.]{%
		\includegraphics[width=0.23\textwidth, clip]{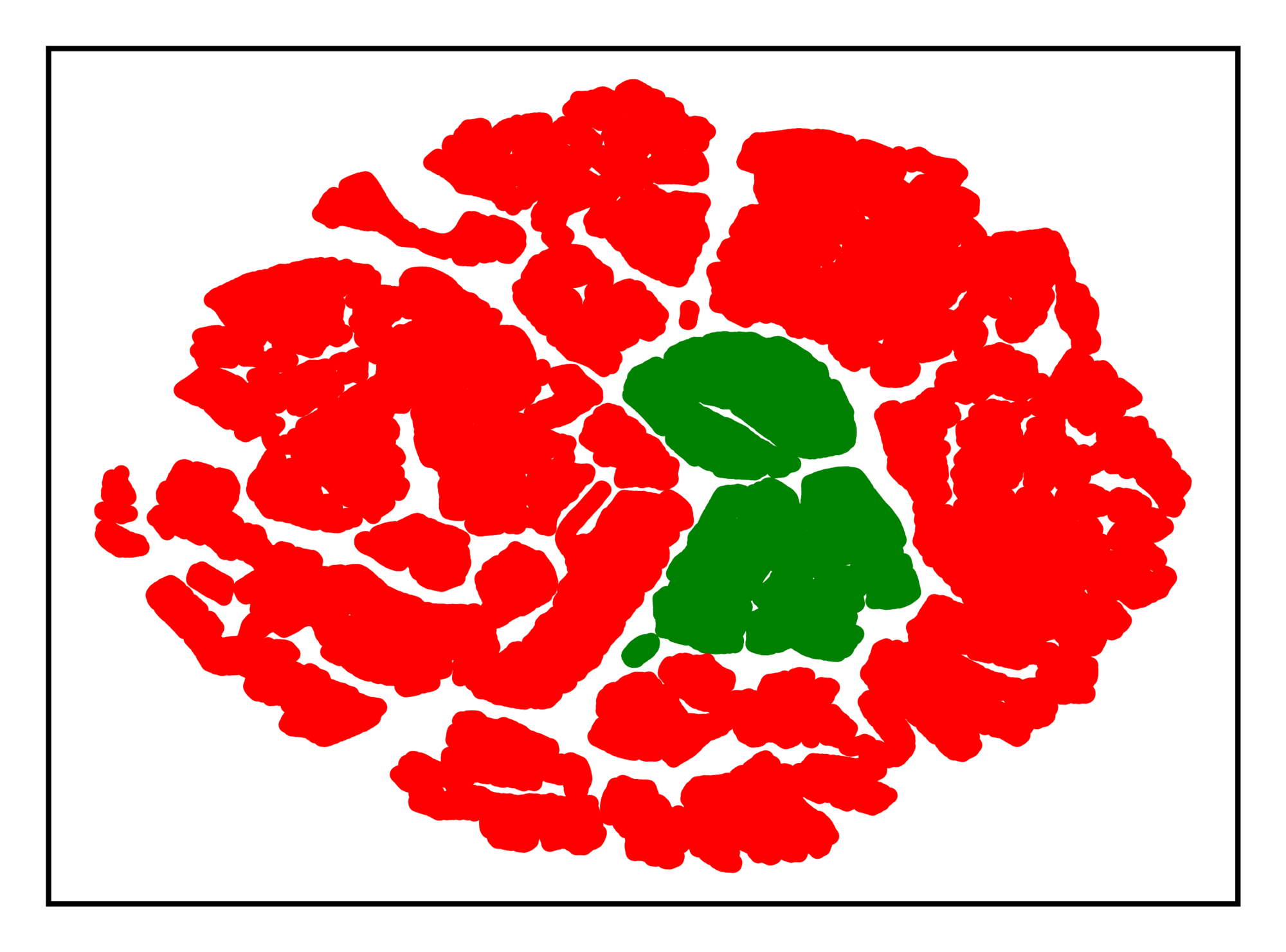}%
		\label{fig:vae-50-59}%
	}
	\subfloat[Slices 55 -- 63.]{%
		\includegraphics[width=0.23\textwidth, clip]{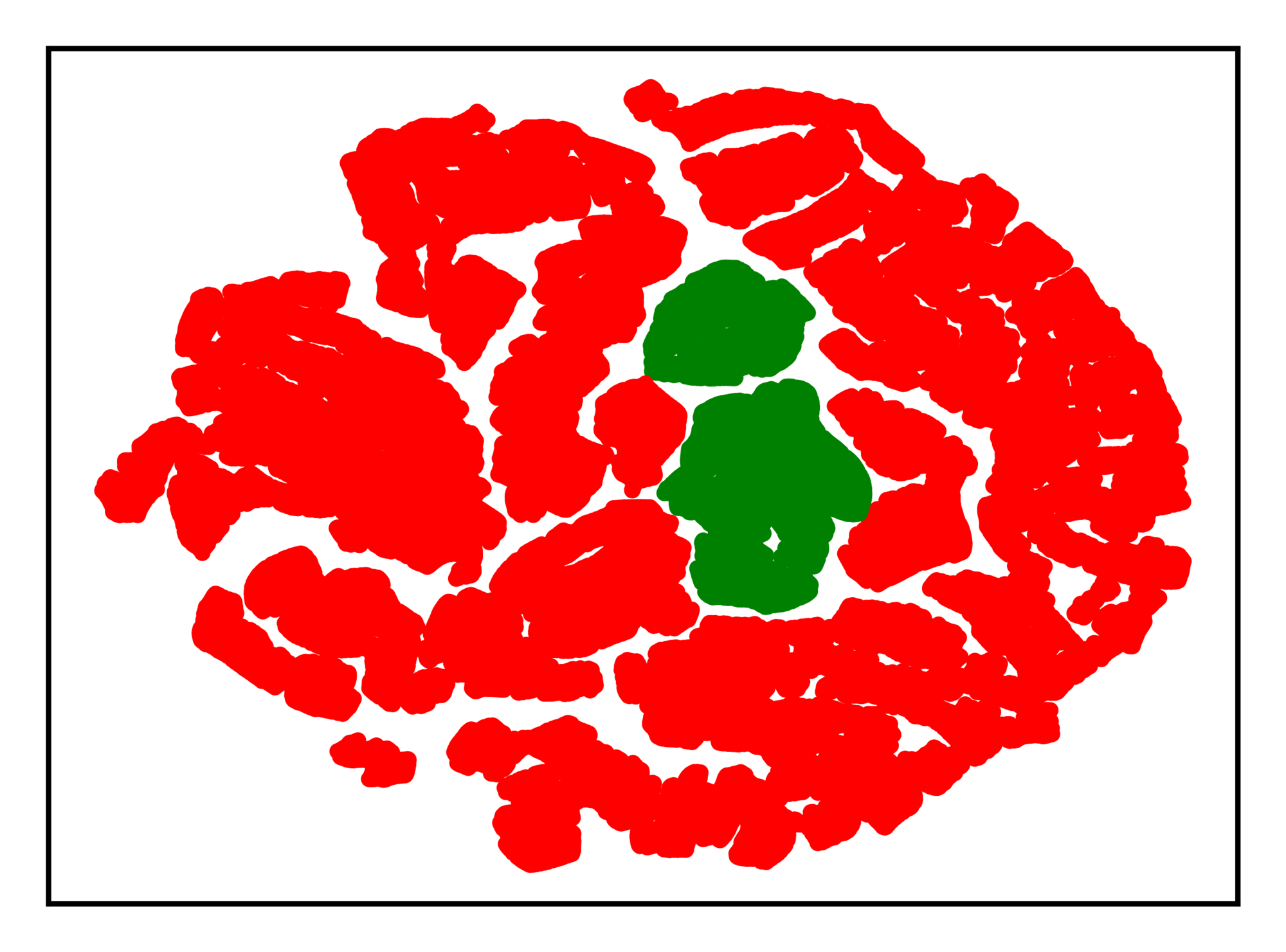}%
		\label{fig:vae-55-64}%
	}
	\caption{t-SNE visualization of mitochondria and other patches for the variational autoencoder ($\beta=10^{-5}$) across groups of 10 slices (\EMBL dataset). Mitochondria patches ($> 50~\%$ overlap) in green, other patches ($0~\%$ overlap) in red.}
	\label{fig:variational-autoencoder-clusters}
\end{figure*}

Next, we assess whether the discrimination results are representative of the entire dataset by evaluating performance across different groups of 10 slices. In \Cref{fig:autoencoder-clusters}, results for the classical AE show that most positive patches cluster together globally (\Cref{fig:ae-all}), though some patches end up in the wrong region. When focusing on smaller groups of 10 slices (\Cref{fig:ae-0-9,fig:ae-5-14,fig:ae-10-19,fig:ae-15-24,fig:ae-20-29,fig:ae-25-34,fig:ae-30-39,fig:ae-35-44,fig:ae-40-49,fig:ae-45-54,fig:ae-50-59,fig:ae-55-64}), we observe very good separation in all groups. Similar results are observed for the VAE in \Cref{fig:variational-autoencoder-clusters}. We conclude that autoencoders perform well in local regions but do not perfectly generalize to distant regions, suggesting that more queries should be provided when accuracy decreases.

\subsubsection{Multiple structures}
\label{sec:multiple-structures}

\begin{figure*}
	\centering
	\subfloat[Classical autoencoder.]{%
		\includegraphics[width=\textwidth]{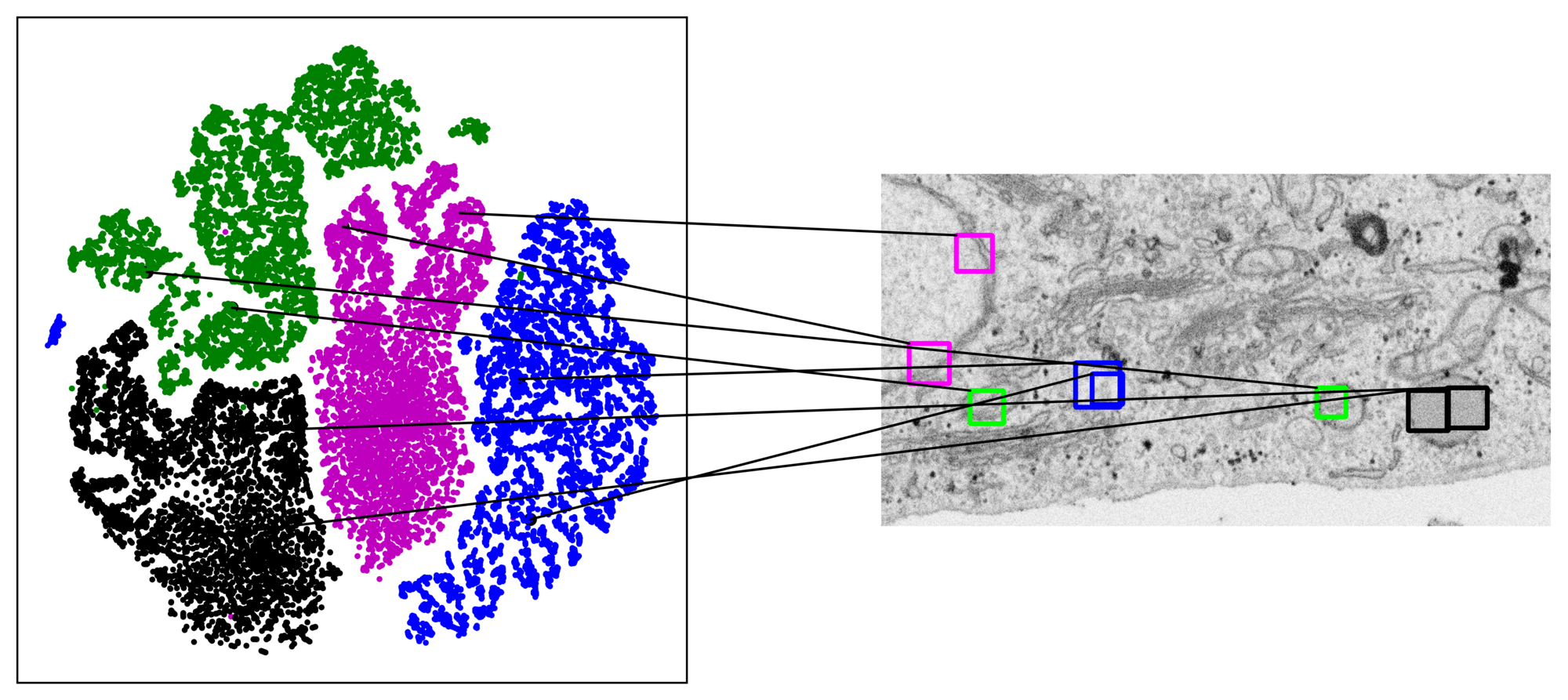}%
		\label{fig:multiple-structures-ae}%
	}
	\vspace{0.5cm}
	\subfloat[Variational autoencoder.]{%
		\includegraphics[width=\textwidth]{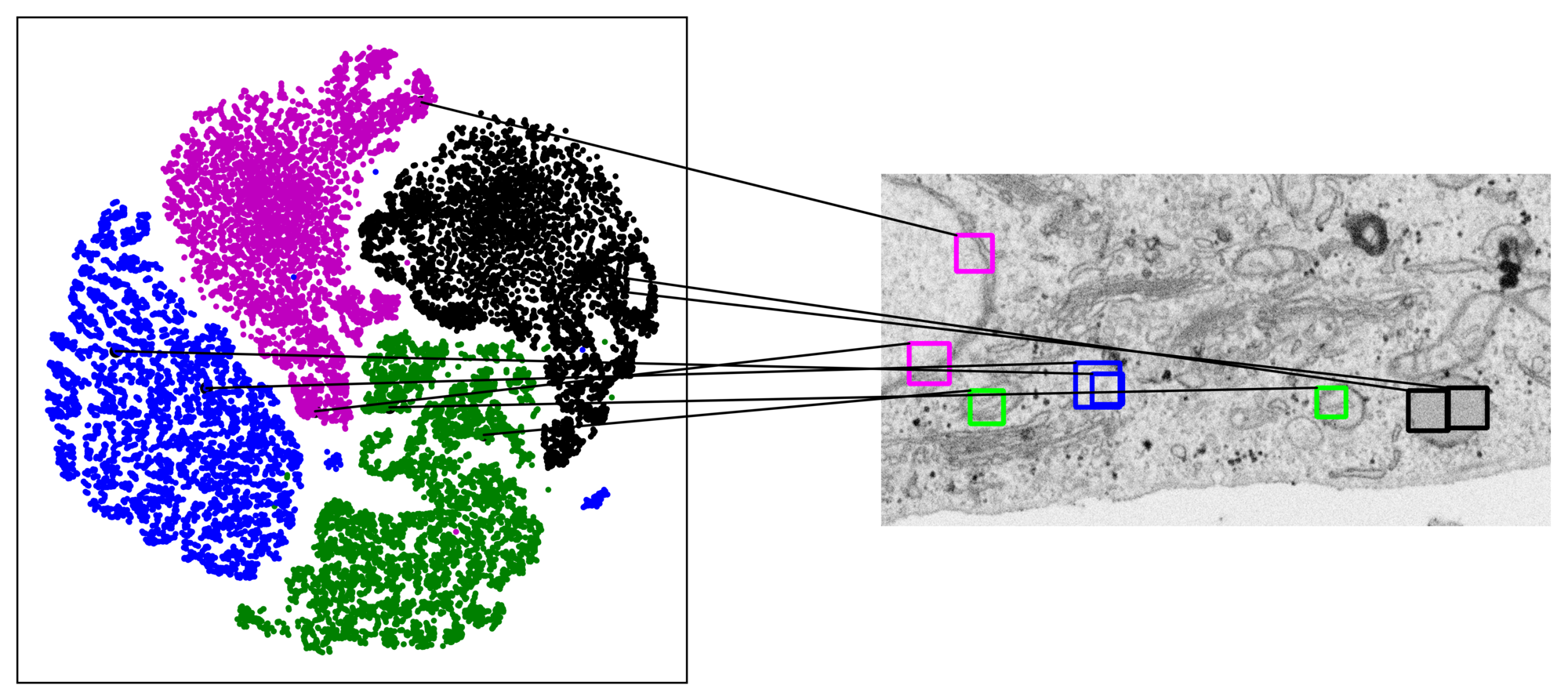}%
		\label{fig:multiple-structures-vae}%
	}
	\caption{t-SNE visualization of four different structures in the latent space for classical autoencoder (\Cref{fig:multiple-structures-ae}) and variational autoencoder (\Cref{fig:multiple-structures-vae}, $\beta=10^{-5}$), for the \Po dataset. \revisiontwo{Colors denote mitochondria (green), nuclear envelope (magenta), endoplasmic reticulum (blue), and vesicles (black).}}
	\label{fig:multiple-structures}
\end{figure*}

We investigate the capacity of fine-tuned models to differentiate between multiple structures. Using patches from 5 slices (210--214) of the \Po dataset with sizes 40--70 pixels, we consider four different structures with 75~\% overlap threshold: \revisiontwo{mitochondria (green), nuclear envelope (magenta), endoplasmic reticulum (blue), and vesicles (black)}. The results in \Cref{fig:multiple-structures} show nearly perfect separation between all structure clusters for both the classical autoencoder (\Cref{fig:multiple-structures-ae}) and variational autoencoder (\Cref{fig:multiple-structures-vae}).

\subsection{Visual evaluation of retrieved patches}

\begin{figure}[t]
	\centering
        \hfill
	\begin{minipage}[c]{0.2\linewidth}
        \centering
        $\underbrace{\includegraphics[width=0.4\linewidth]{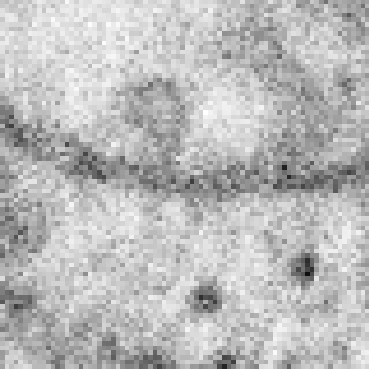}
        \hspace{1cm}
        \includegraphics[width=0.4\linewidth]{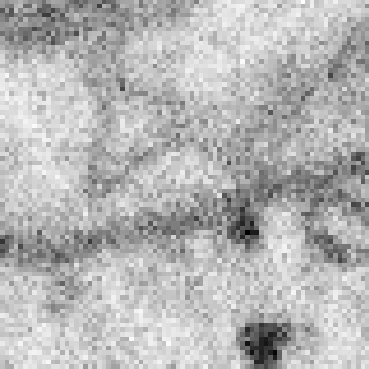}}_{\mathcal{Q}^+}$\\
        \vspace{1cm}
        $\underbrace{\includegraphics[width=0.4\linewidth]{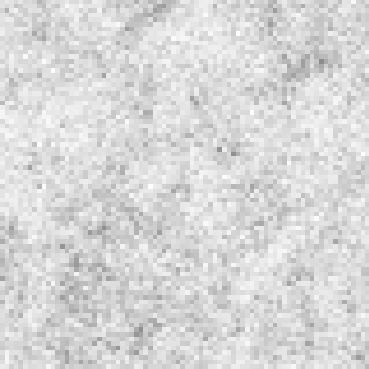}
        \hspace{1cm}
        \includegraphics[width=0.4\linewidth]{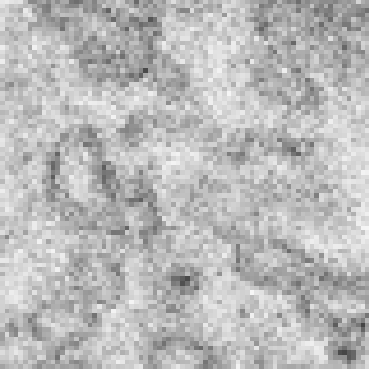}}_{\mathcal{Q}^-}$

    \end{minipage}%
    \hfill
    \begin{minipage}[c]{0.7\linewidth}
        \centering
        \includegraphics[width=0.7\linewidth]{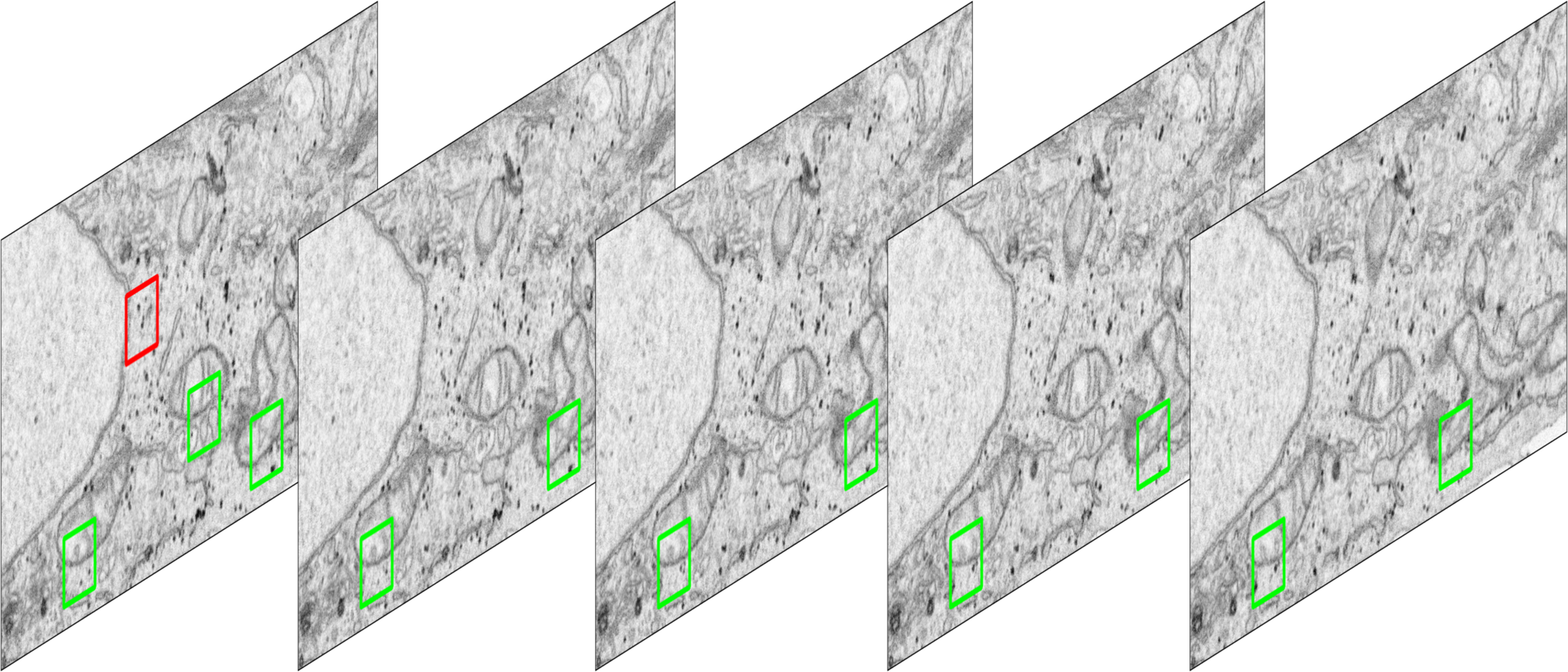}

        \vspace{3mm}
        
        \includegraphics[width=0.7\linewidth]{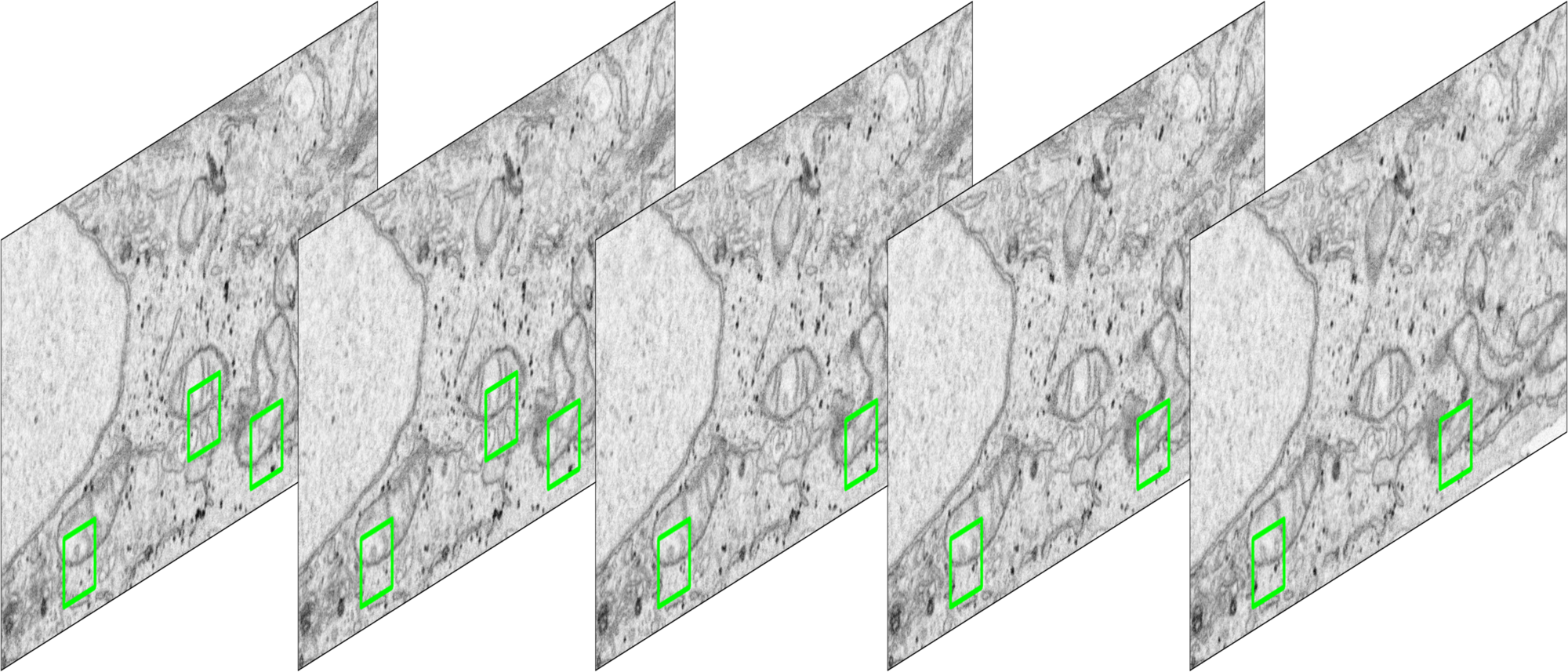}
    \end{minipage}

    \vspace{1cm}
    \includegraphics[width=\linewidth]{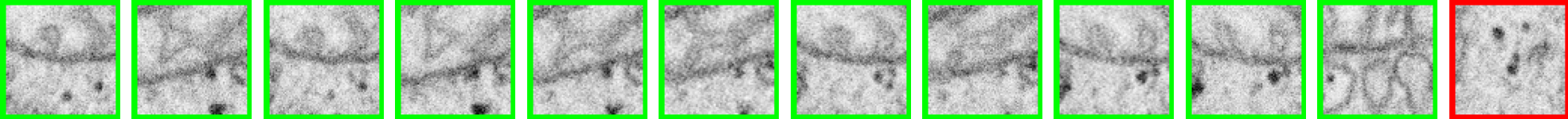}\\
    \vspace{0.5cm}
    \includegraphics[width=\linewidth]{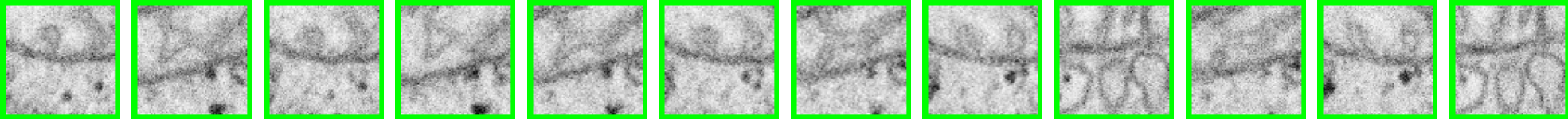}\\
	\hfill
	\caption{Examples of retrieved patches based on specified positive/negative queries (left) from the \Po{} dataset. Right: AE/VAE results; bottom rows show the retrieved patches sorted by decreasing similarity score. Green indicates true positives; red indicates false positives.}
	\label{fig:comparison_2017}
\end{figure}

We visually evaluate the proposed framework by examining returned patches in response to positive and negative queries.
In \Cref{fig:comparison_2017}, we present the top 12 retrieved patches for the fine-tuned autoencoder and variational autoencoder on the \Po dataset, based on the two positive and negative queries shown on the left. On the right, we highlight the retrieved patches on their corresponding slices, while the bottom rows show patches sorted by descending similarity score. Although the framework retrieves some false-positive patches, this is acceptable because the search space has already been substantially reduced.

\begin{figure}[t]
	\centering
        \hfill
	\begin{minipage}[c]{0.2\linewidth}
        \centering
        $\underbrace{\includegraphics[width=0.4\linewidth]{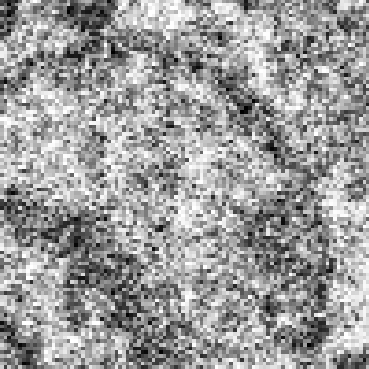}
        \hspace{1cm}
        \includegraphics[width=0.4\linewidth]{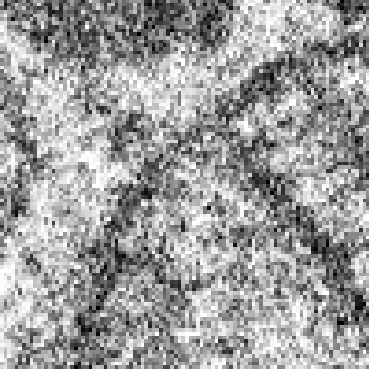}}_{\mathcal{Q}^+}$\\
        \vspace{1cm}
        $\underbrace{\includegraphics[width=0.4\linewidth]{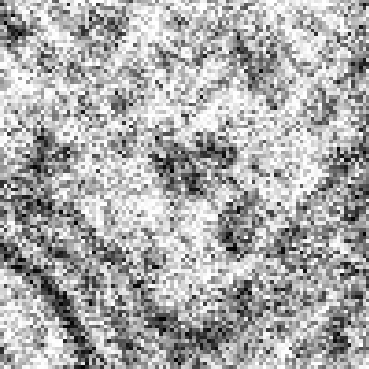}
        \hspace{1cm}
        \includegraphics[width=0.4\linewidth]{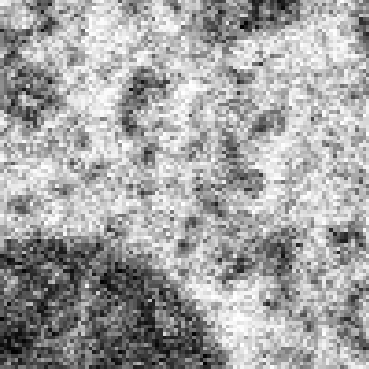}}_{\mathcal{Q}^-}$

    \end{minipage}%
    \hfill
    \begin{minipage}[c]{0.7\linewidth}
        \centering
        \includegraphics[width=0.7\linewidth]{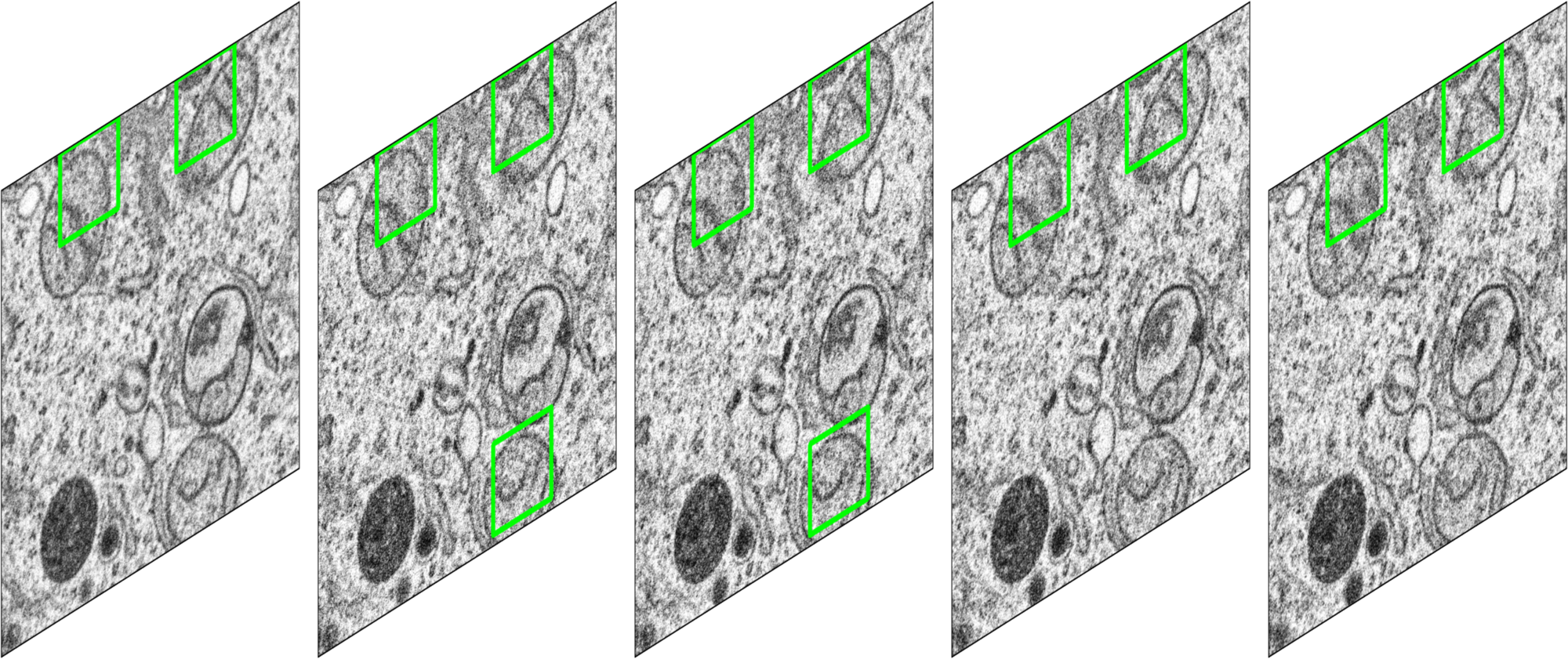}

        \vspace{3mm}
        
        \includegraphics[width=0.7\linewidth]{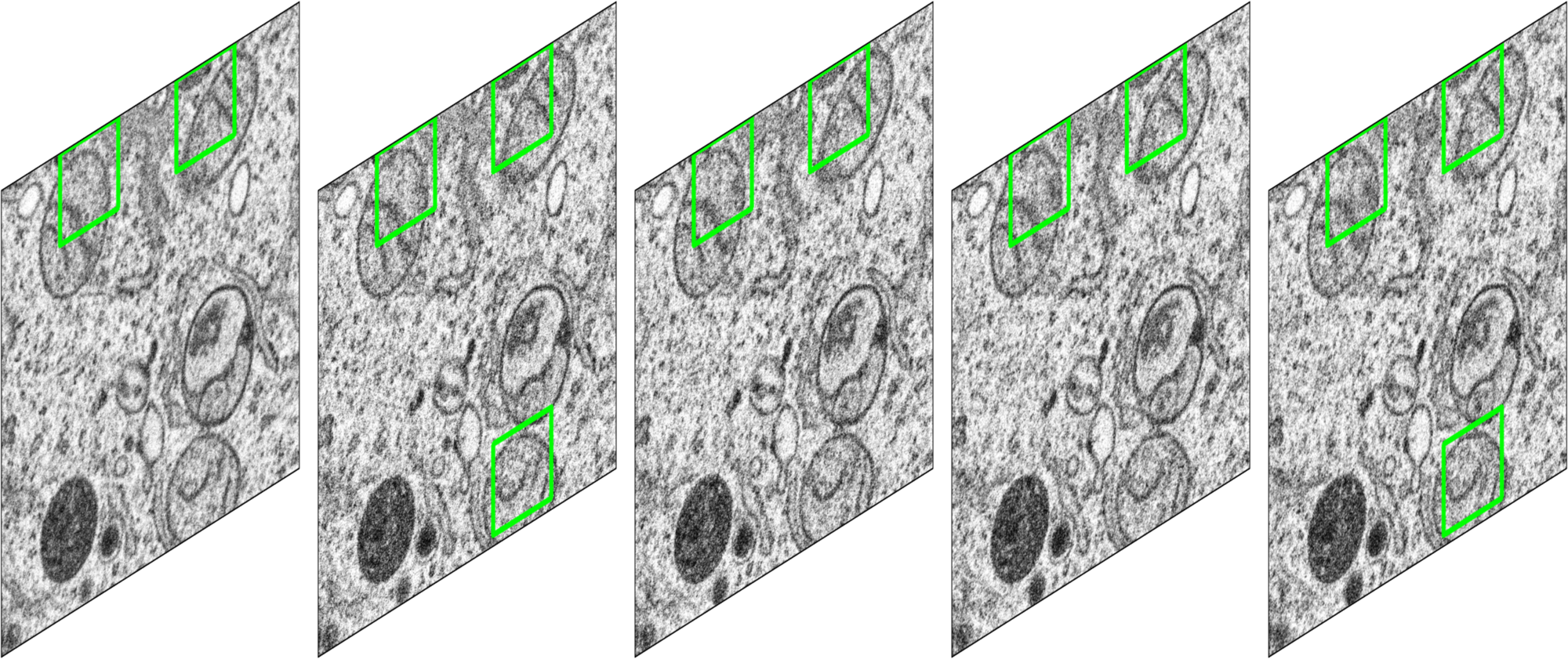}
    \end{minipage}

    \vspace{1cm}
    \includegraphics[width=\linewidth]{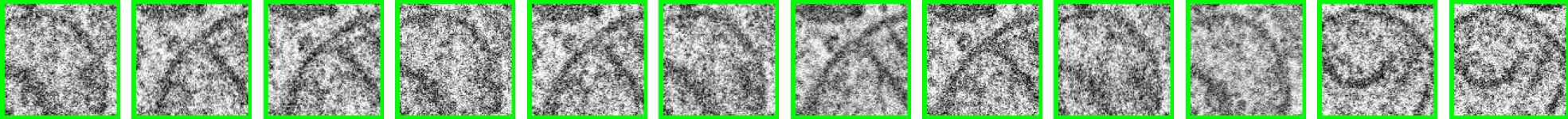}\\
    \vspace{0.5cm}
    \includegraphics[width=\linewidth]{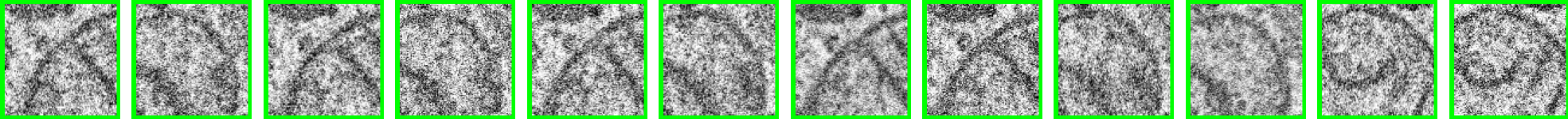}\\
	\hfill
	\caption{Examples of retrieved patches based on specified positive/negative queries (left) from the \EMBL{} dataset. Right: AE/VAE results; bottom rows show the retrieved patches sorted by decreasing similarity score. Green indicates true positives; red indicates false positives.}
	\label{fig:comparison_EMBL}
\end{figure}

\begin{figure}[t]
	\centering
        \hfill
	\begin{minipage}[c]{0.2\linewidth}
        \centering
        $\underbrace{\includegraphics[width=0.4\linewidth]{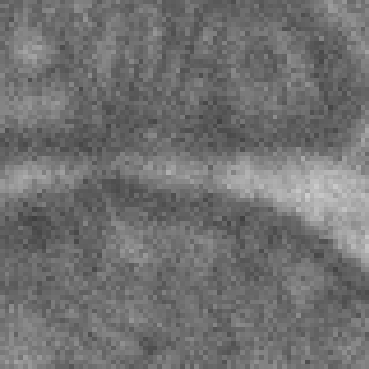}
        \hspace{1cm}
        \includegraphics[width=0.4\linewidth]{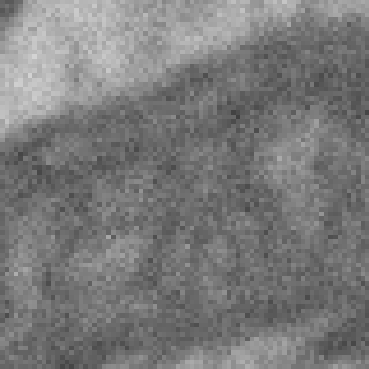}}_{\mathcal{Q}^+}$\\
        \vspace{1cm}
        $\underbrace{\includegraphics[width=0.4\linewidth]{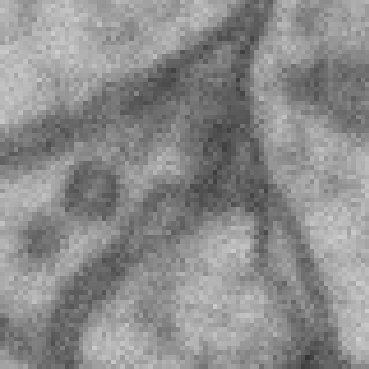}
        \hspace{1cm}
        \includegraphics[width=0.4\linewidth]{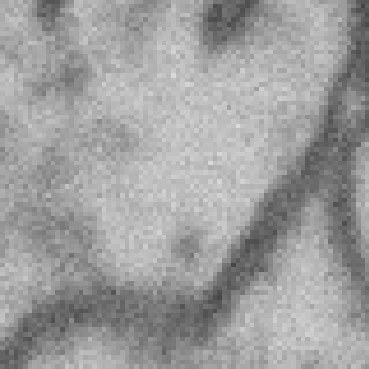}}_{\mathcal{Q}^-}$
    \end{minipage}%
    \hfill
    \begin{minipage}[c]{0.7\linewidth}
        \centering
        \includegraphics[width=0.7\linewidth]{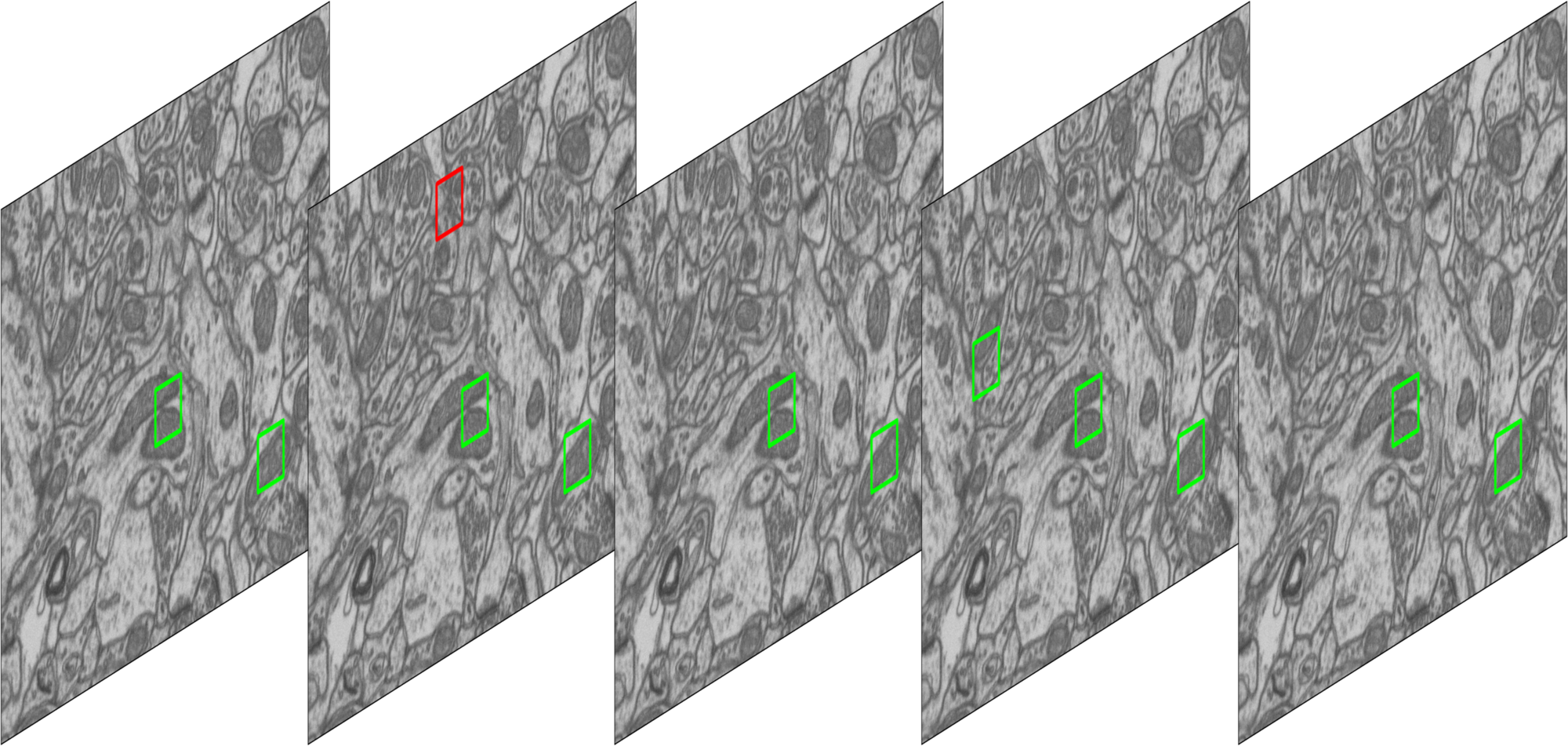}
        
        \vspace{3mm}
        
        \includegraphics[width=0.7\linewidth]{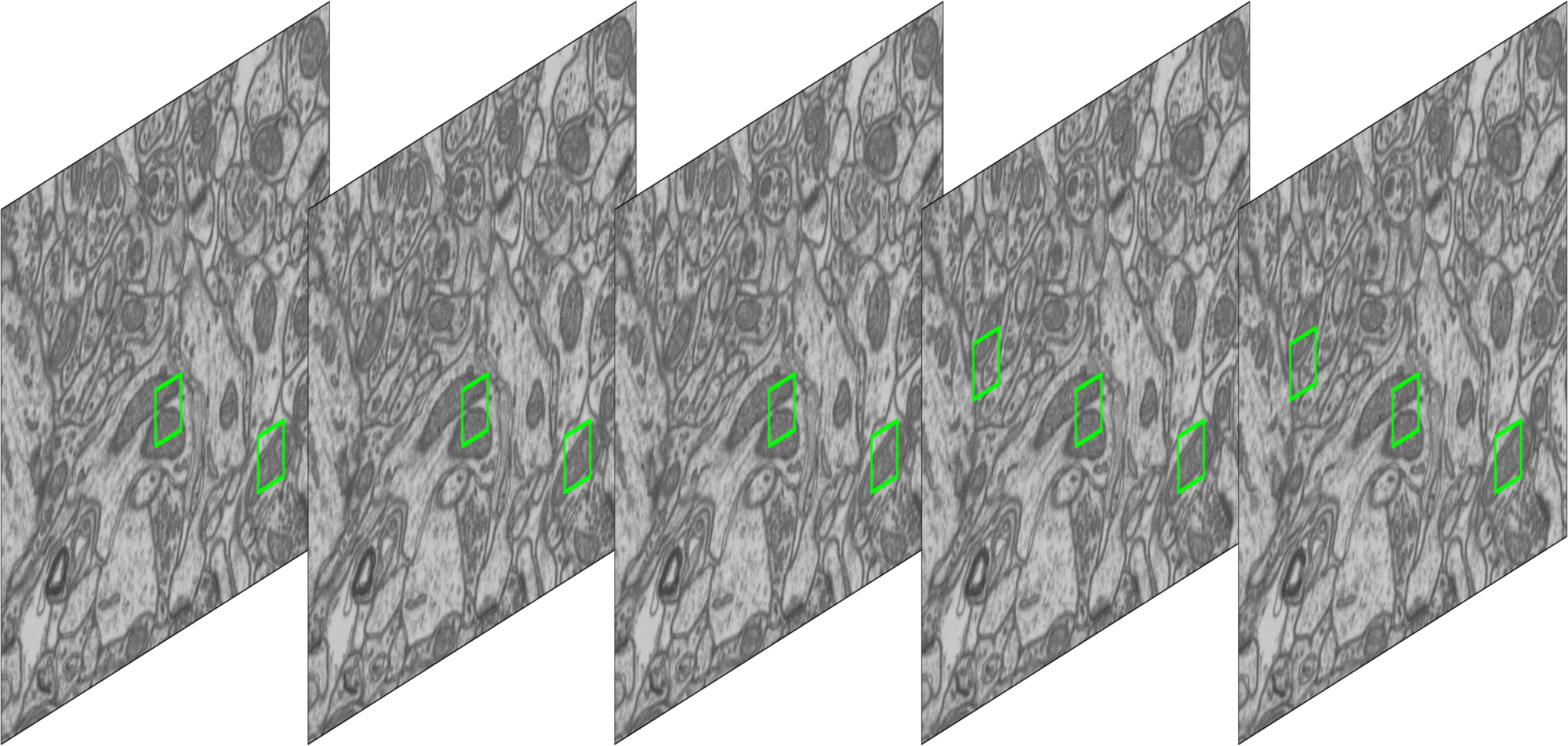}
    \end{minipage}

    \vspace{1cm}
    \includegraphics[width=\linewidth]{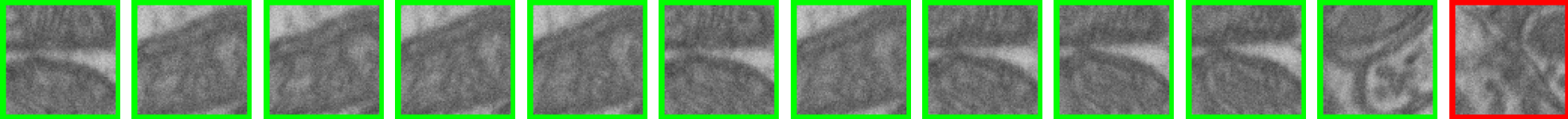}\\
    \vspace{0.5cm}
    \includegraphics[width=\linewidth]{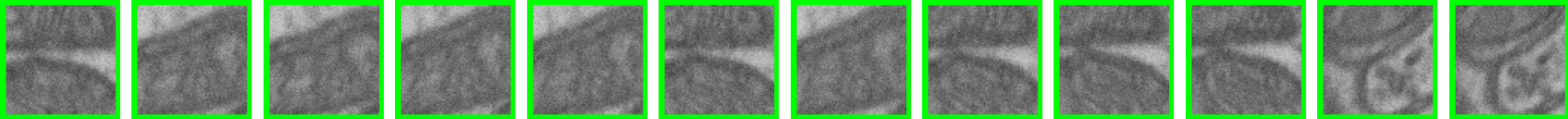}\\
	\hfill
	\caption{Examples of retrieved patches based on specified positive/negative queries (left) from the \EPFL{} dataset. Right: AE/VAE results; bottom rows show the retrieved patches sorted by decreasing similarity score. Green indicates true positives; red indicates false positives.}
	\label{fig:comparison_EPFL}
\end{figure}

\begin{figure*}
	\centering
	\subfloat{%
		\includegraphics[width=0.72\textwidth]{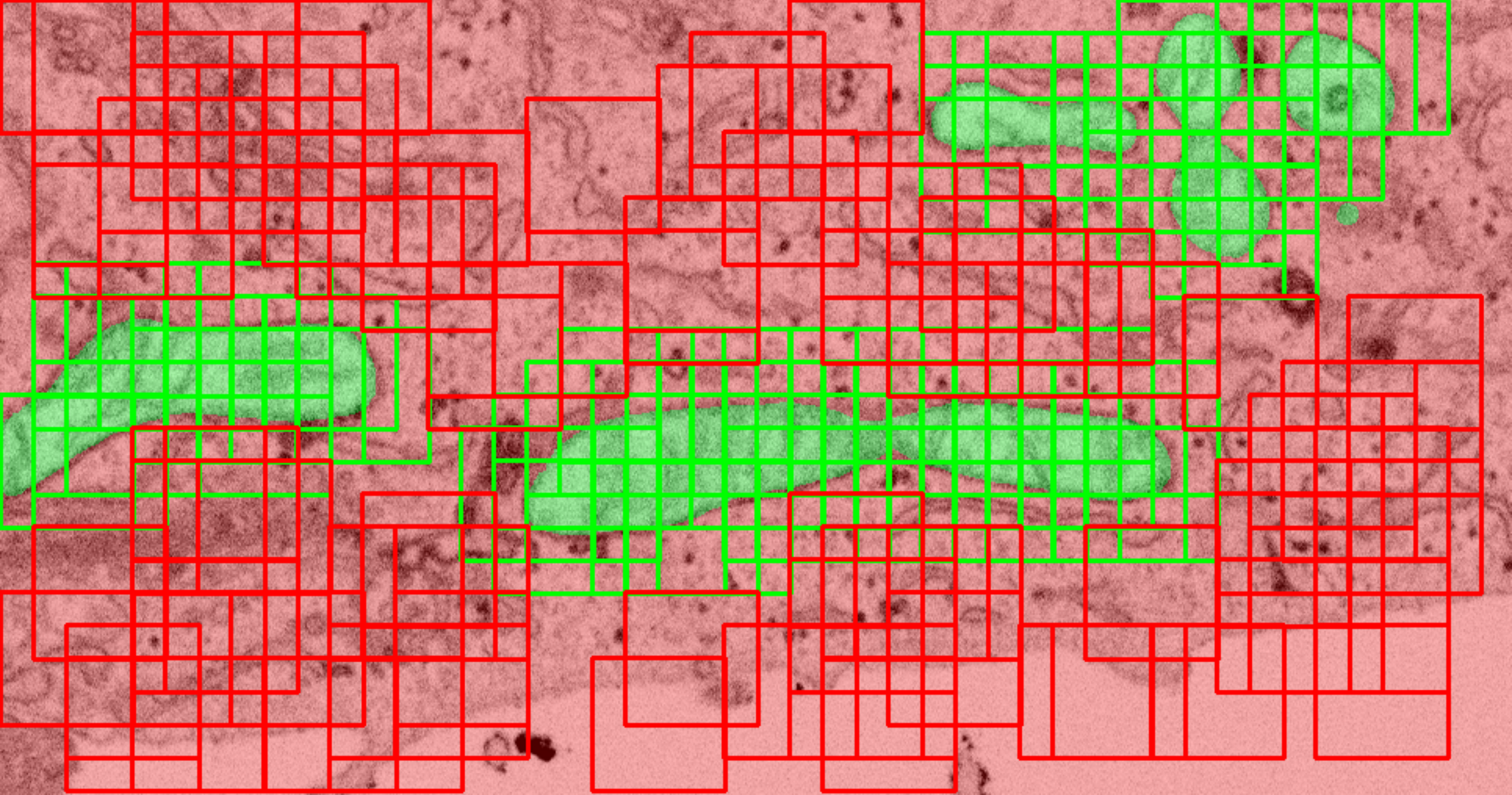}%
	}\\
	\subfloat{%
	\includegraphics[width=0.72\textwidth]{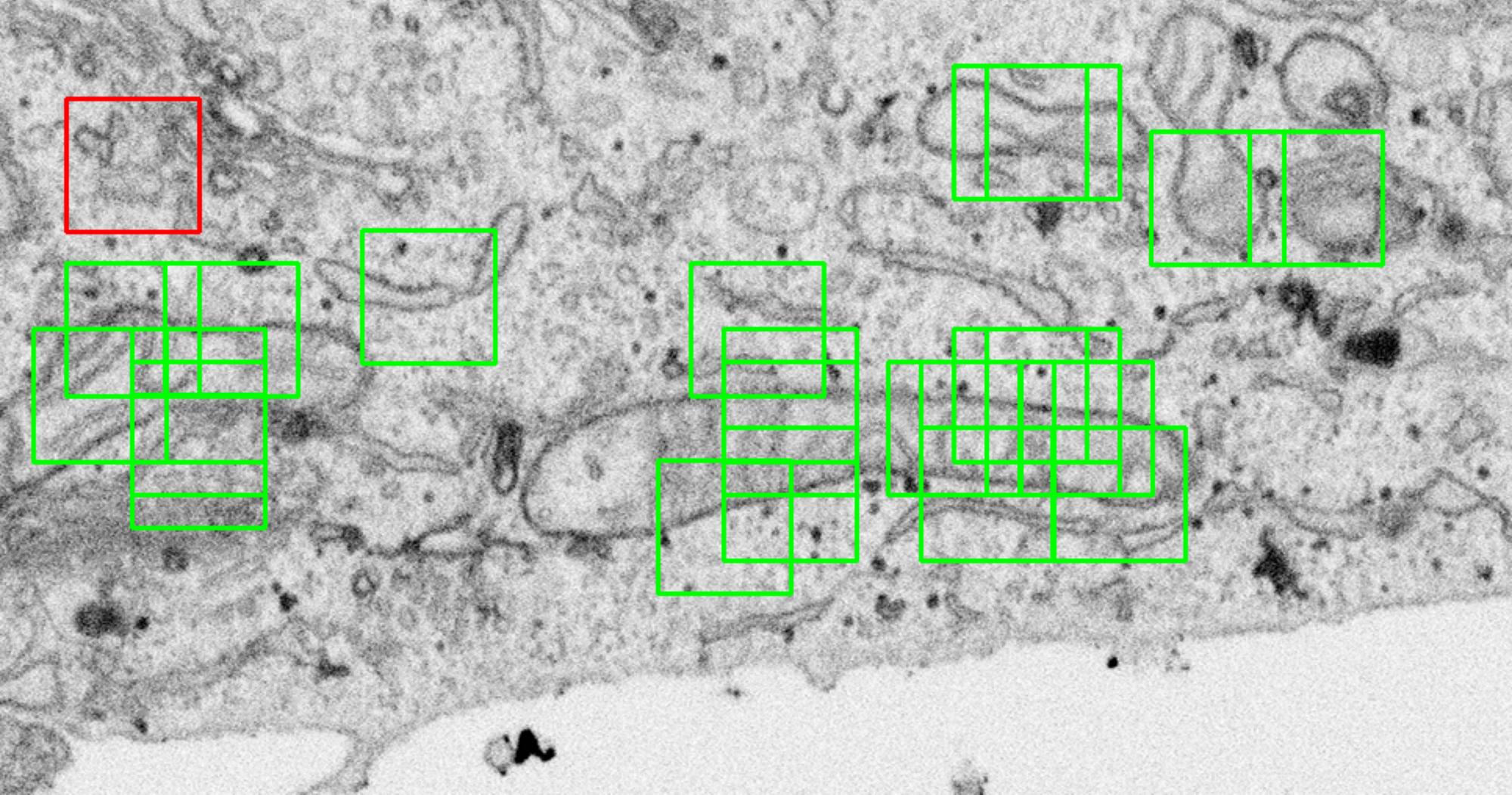}%
	}\\
	\subfloat{%
		\includegraphics[width=0.72\textwidth]{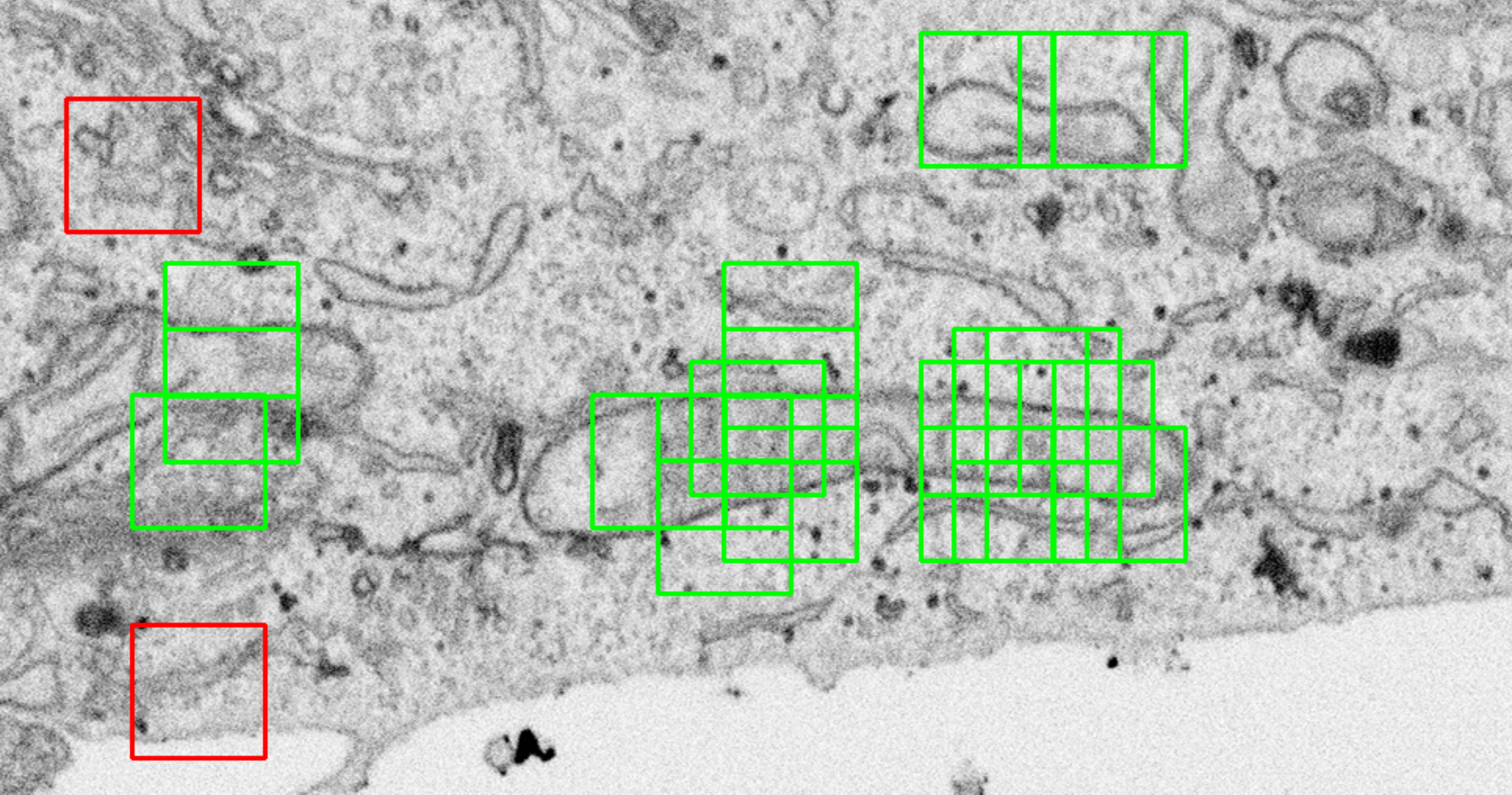}%
	}\\
	\caption{Examples of the retrieved patches corresponding to the specified positive and negative queries (100 from each class) taken from (pseudo-)labeling, from the \Po dataset. The query slice is shown on top, with positive (resp. negative) queries indicated in green (resp. red). Below are the 25 most similar patches for the classical autoencoder and the variational autoencoder, respectively. The patches are indicated in green (resp. red) if the patch is a true positive (resp. false positive).}
	\label{fig:comparison_2017_multiple_queries}
\end{figure*}

Similar results are shown for the \EMBL dataset (\Cref{fig:comparison_EMBL}) and \EPFL dataset (\Cref{fig:comparison_EPFL}), comparing both the AE and VAE. In \Cref{fig:comparison_2017_multiple_queries}, we illustrate the framework using 100 positive and 100 negative query patches (shown at the top), which are automatically extracted from a (pseudo-)labeled slice. The bottom two rows present the top 25 retrieved patches for the selected queries across two example slices. Despite some false positives, the results remain highly informative for human experts to enhance labeling efficiency.

\subsection{Performance evaluation of the proposed framework}
\label{sec:evaluation}

We evaluate our framework for all three datasets and different split sizes. We plot the precision-recall curves and compute the average precision, which is the integral over the PR-curve.

\subsubsection{Discussion on the PR-curves}

\begin{figure*}
	\centering
	\subfloat[\EMBL, split size 10.]{%
		\includegraphics[width=0.46\textwidth]{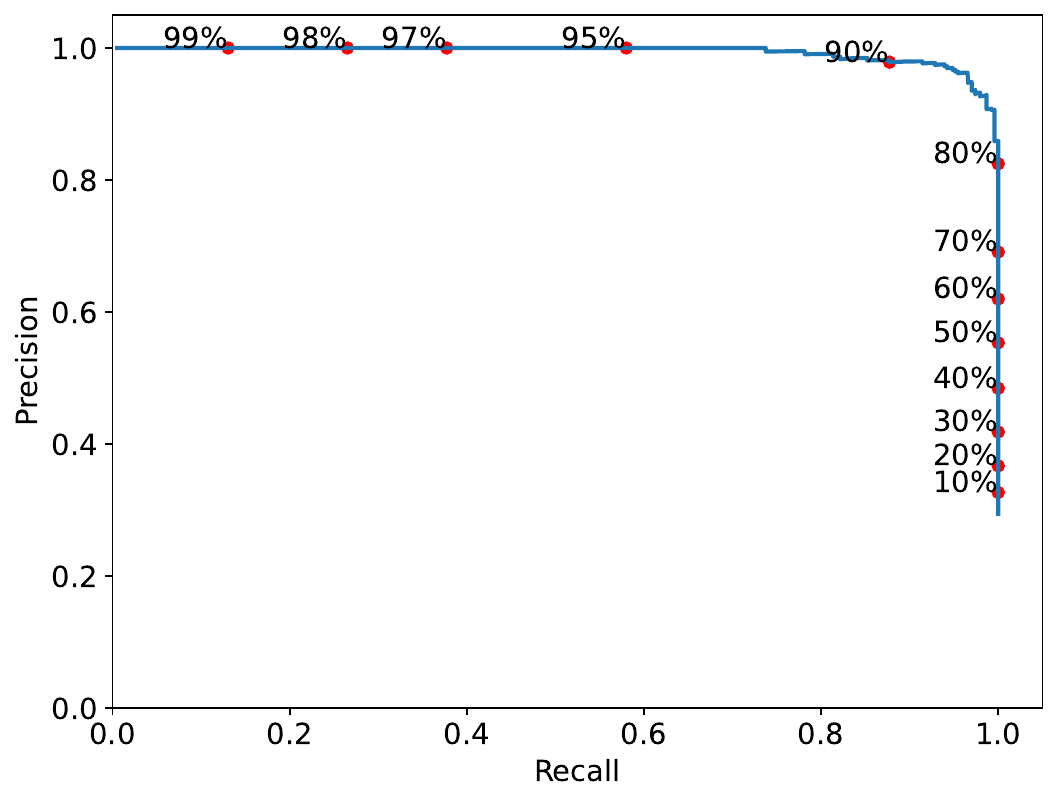}%
	}\hfill
	\subfloat[\EMBL, split size 30.]{%
		\includegraphics[width=0.46\textwidth]{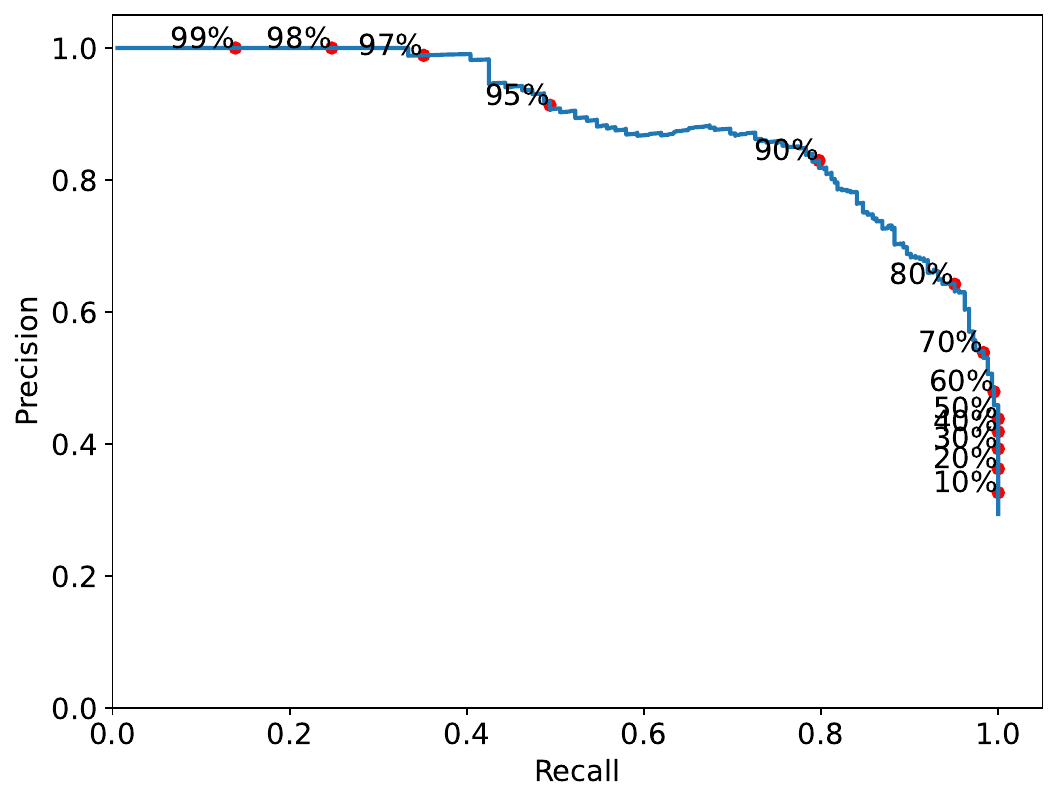}%
	}\\[1ex]
    \subfloat[\Po, split size 10.]{%
		\includegraphics[width=0.46\textwidth]{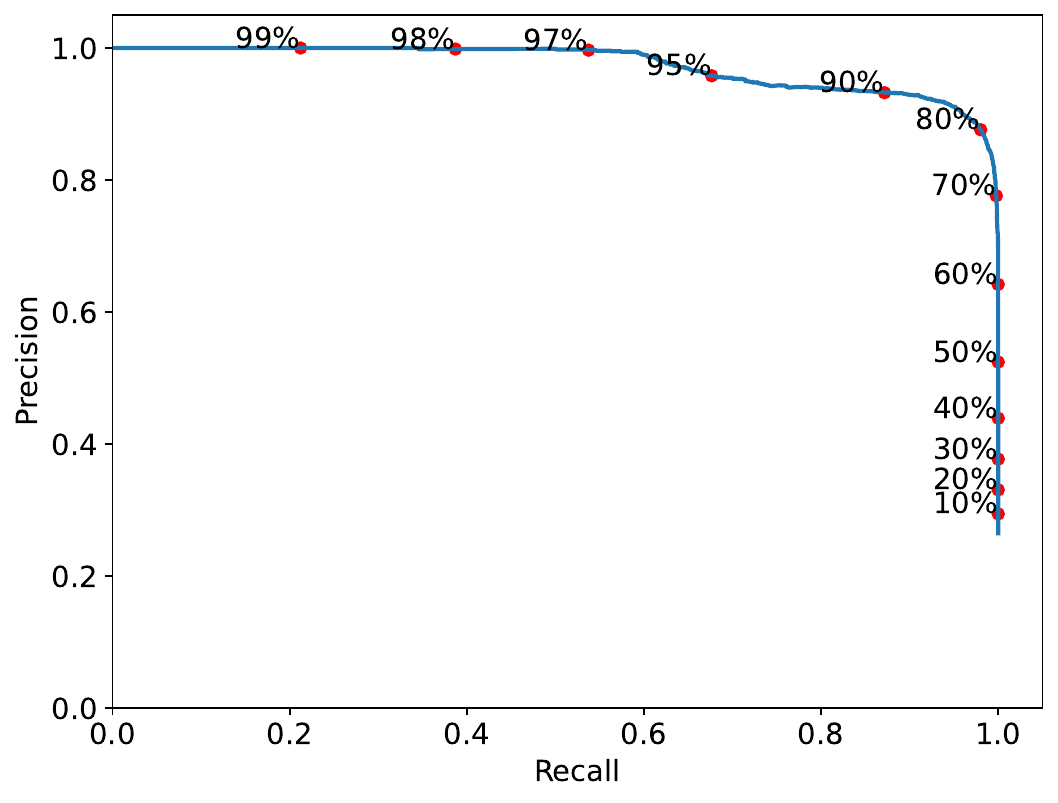}
	}\hfill
	\subfloat[\Po, split size 30.]{%
		\includegraphics[width=0.46\textwidth]{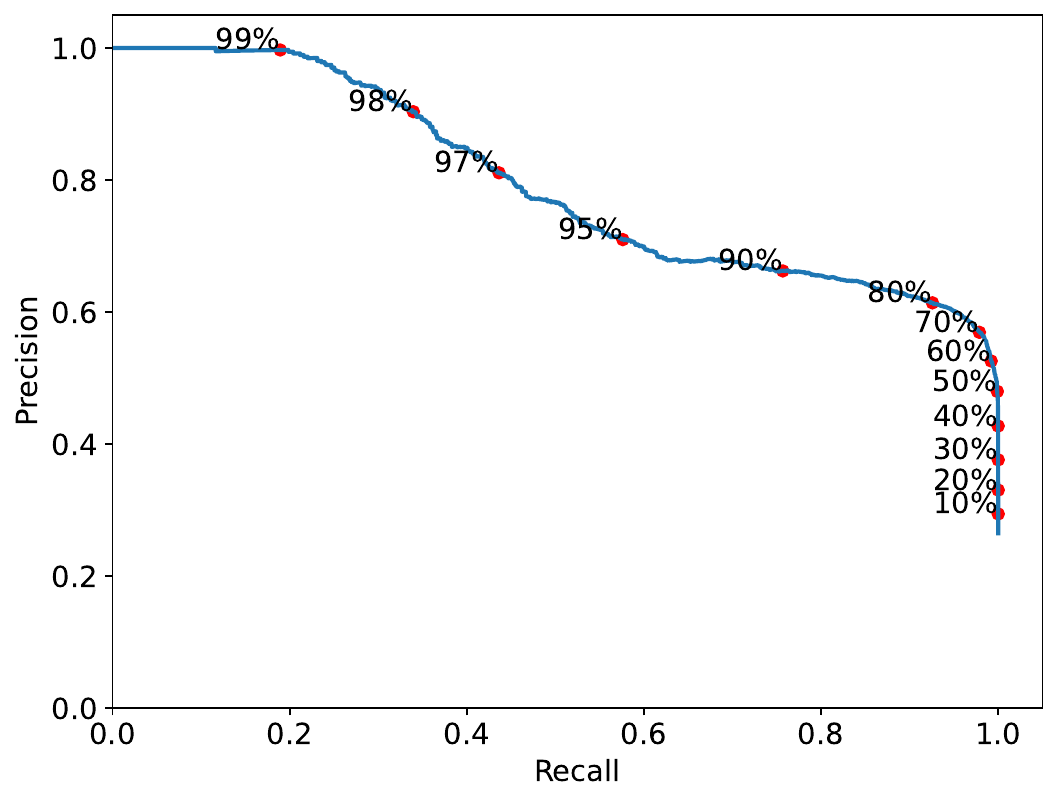}%
	}\\[1ex]
	\subfloat[\EPFL, split size 10.]{%
		\includegraphics[width=0.46\textwidth]{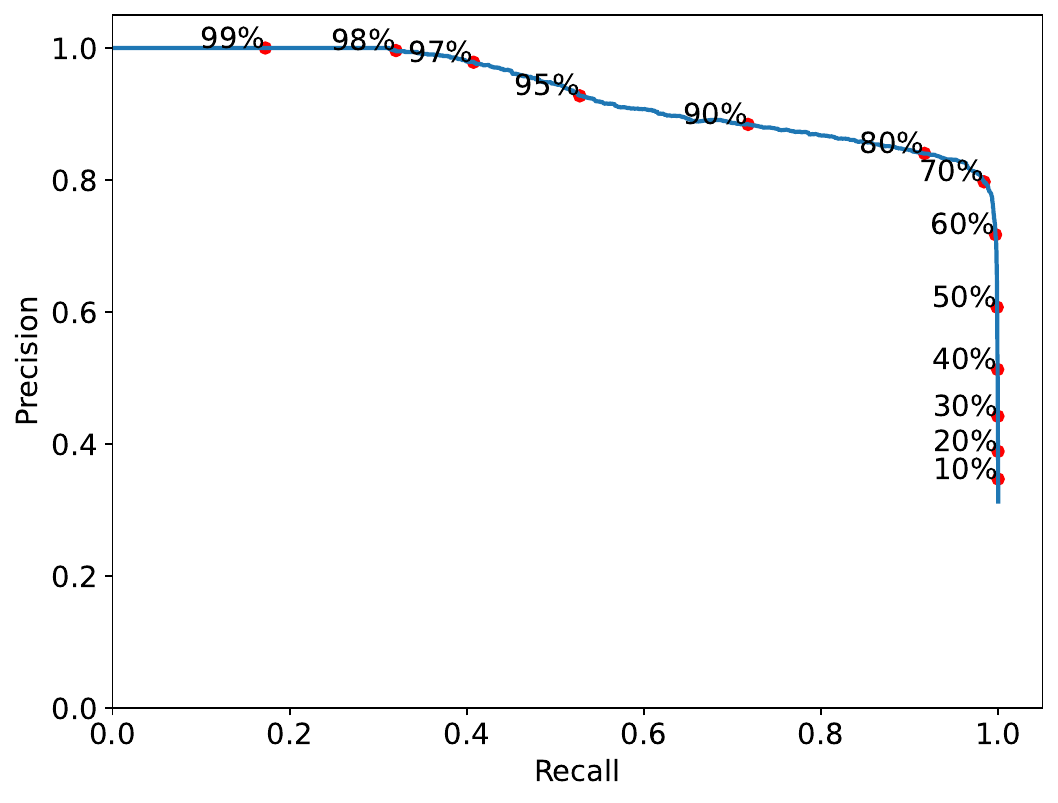}%
	}\hfill
	\subfloat[\EPFL, split size 30.]{%
		\includegraphics[width=0.46\textwidth]{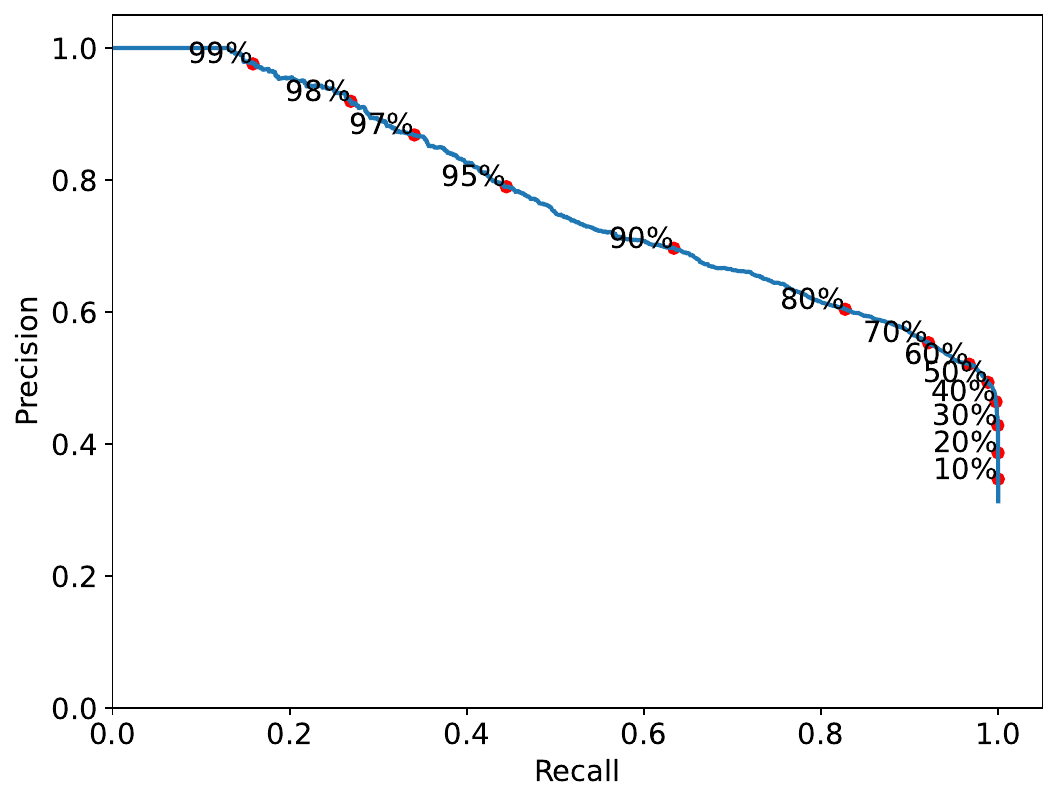}%
	}
	\caption{Precision-recall curves for the \EMBL dataset (top), \Po (middle), and \EPFL (bottom), and split sizes 10 (left) and 30 (right) for the fine-tuned classical autoencoder, which is the best-performing descriptor. \revisiontwo{Red points indicate reduction rates: the labeled value is the percentage of patches the user no longer needs to inspect; the corresponding retention rate (percentage of patches kept, sorted by similarity) is $100\%$ minus the labeled value.}}
	\label{fig:pr-curves}
\end{figure*}

In \Cref{fig:pr-curves}, the PR-curves are shown for the \EMBL dataset. We show the PR-curves for the fine-tuned classical autoencoder, which gave the best overall accuracy (see \Cref{sec:average-precision}). The PR-curves correspond to split sizes of 10 (left, queries from 1 of every 10 slices) and 30 (right, queries from 1 of every 30 slices). The performance improves as the number of query slices increases, likely due to spatial coherence in the dataset, whereby slices at farther distances exhibit less similarity.

\revisiontwo{The points represent \emph{reduction rates}: the labeled value is the fraction of patches the user no longer needs to inspect, i.e., $1$ minus the \emph{retention rate} (the fraction of highest-similarity patches that the user retains and inspects).} For reduction rates as high as 70\%, we observe a recall of nearly one in most cases, indicating that almost all structures are retrieved from the data while discarding the majority of irrelevant patches. Similar results are observed for the \Po and \EPFL datasets. This demonstrates the method's potential to reduce the search space when searching for a structure of interest, which is highly valuable for biologists.

\subsubsection{Discussion on the average precisions}
\label{sec:average-precision}

In \Cref{tab:results}, the average precisions (APs) are given for the pretrained and fine-tuned autoencoder models and different combinations of datasets and split sizes.

From these results, we can conclude that the fine-tuned models perform significantly better than the pretrained models. In all experiments, the pretrained models are outperformed by the fine-tuned models, sometimes with over 5~\% performance increase. \revision{To verify this observation statistically, we performed paired one-sided Wilcoxon signed-rank tests over all $n=14$ (dataset, split-size) configurations of \Cref{tab:results}. Fine-tuning yields a significant improvement both for the classical autoencoder (median $\Delta\text{AP}=+3.65$~pp, $W=105$, $p=4.9\cdot 10^{-4}$) and for the variational autoencoder (median $\Delta\text{AP}=+2.80$~pp, $W=97$, $p=1.5\cdot 10^{-3}$).}

Also, the classical autoencoder performs comparatively better than the variational autoencoder, especially after fine-tuning. \revision{The corresponding paired one-sided Wilcoxon test (fine-tuned AE vs.\ fine-tuned VAE, $n=14$) confirms that the gap is significant (median $\Delta\text{AP}=+0.90$~pp, $W=96$, $p=3.1\cdot 10^{-3}$).} In some cases, the fine-tuned variational autoencoder performs worse than the pretrained version, which could be explained by the fact that variational autoencoders are more challenging to train as a descriptor, and we cannot assume that we can simply fine-tune an existing VAE with the same $\beta$ parameter. This aligns with the conclusions of \cite{zizakic_efficient_2022}, which state that classical autoencoders are perhaps better suited for descriptor models due to the difficulty of training variational autoencoders.

Finally, we can also notice different performance gains for the various datasets. These gains are relatively high for the \EMBL and \EPFL datasets but much lower for the \Po dataset. This can be explained by the fact that it is difficult to fine-tune learned features on electron microscopy data in general, as we have only fine-tuned on a few datasets, and these datasets have different statistical properties. \revision{In particular, the \Po dataset differs from \EMBL and \EPFL in ways that make descriptor matching more challenging. The structures of interest are smaller relative to the image scale, increasing sensitivity to partial matches, local background clutter, and patch-level misalignment. Moreover, their appearance differs from the corresponding structures in the other datasets, so retrieval cannot rely as strongly on shared global appearance cues such as intensity or texture. Instead, the descriptor must capture more local shape- and boundary-related information. These factors make \Po a more demanding setting and help explain the lower performance observed on this dataset. The \Po results should therefore be interpreted as a conservative estimate of the framework's performance, rather than as a fundamental limitation.}

\begin{table*}
	\caption{Overview of the Average Precision (AP) for the classical and variational autoencoder, and the pretrained and fine-tuned versions. The best-performing descriptor is indicated in \textbf{bold}.}
	\label{tab:results}
	\centering
	\begin{tabular*}{\linewidth}{@{\extracolsep{\fill}} cccccc }
		\hline
		\multirow{2}{*}{\textbf{Dataset}} & \multirow{2}{*}{\textbf{Split size}} & \multicolumn{2}{c}{\textbf{Pretrained}} & \multicolumn{2}{c}{\textbf{Fine-tuned}}\\
		\cline{3-6}
		& & AE & VAE & AE & VAE\\
		\hline
		 \EMBL      &           10 &         0.971 &          0.974  &        \textbf{0.988} &         0.982 \\
 \EMBL      &           20 &         0.890 &          0.873  &        \textbf{0.928} &         0.922 \\
 \EMBL      &           30 &         0.808 &          0.811 &        \textbf{0.887} &         0.867 \\
 \EMBL      &           50 &         0.595 &          0.580  &        \textbf{0.779} &         0.755 \\
 \EPFL      &           10 &         0.916   &          0.923 &        \textbf{0.932} &         0.930 \\
 \EPFL      &           20 &         0.797 &          0.800 &        0.825  &         \textbf{0.832}  \\
 \EPFL      &           30 &         0.723 &          0.732 &        \textbf{0.773} &         0.770  \\
 \EPFL      &           50 &         0.700 &          0.699 &        0.726 &         \textbf{0.731}  \\
 \EPFL      &          100 &         0.632 &          0.584 &        \textbf{0.680} &         0.670 \\
 \Po       &           10 &         0.959 &          0.969 &        \textbf{0.973} &         0.965 \\
 \Po       &           20 &         0.849 &          0.871 &        \textbf{0.884} &         0.869 \\
 \Po       &           30 &         0.747 &          0.790 &        \textbf{0.795} &         0.779 \\
 \Po       &           50 &         0.676 &          0.700 &        \textbf{0.741}  &         0.714 \\
 \Po       &          100 &         0.613 &          0.587  &        \textbf{0.626} &         0.611 \\
		\hline
	\end{tabular*}
\end{table*}

\subsubsection{\revisiontwo{Operational use in downstream analysis}}
\label{sec:operational}
\begin{figure*}
	\centering
	\includegraphics[width=0.95\textwidth]{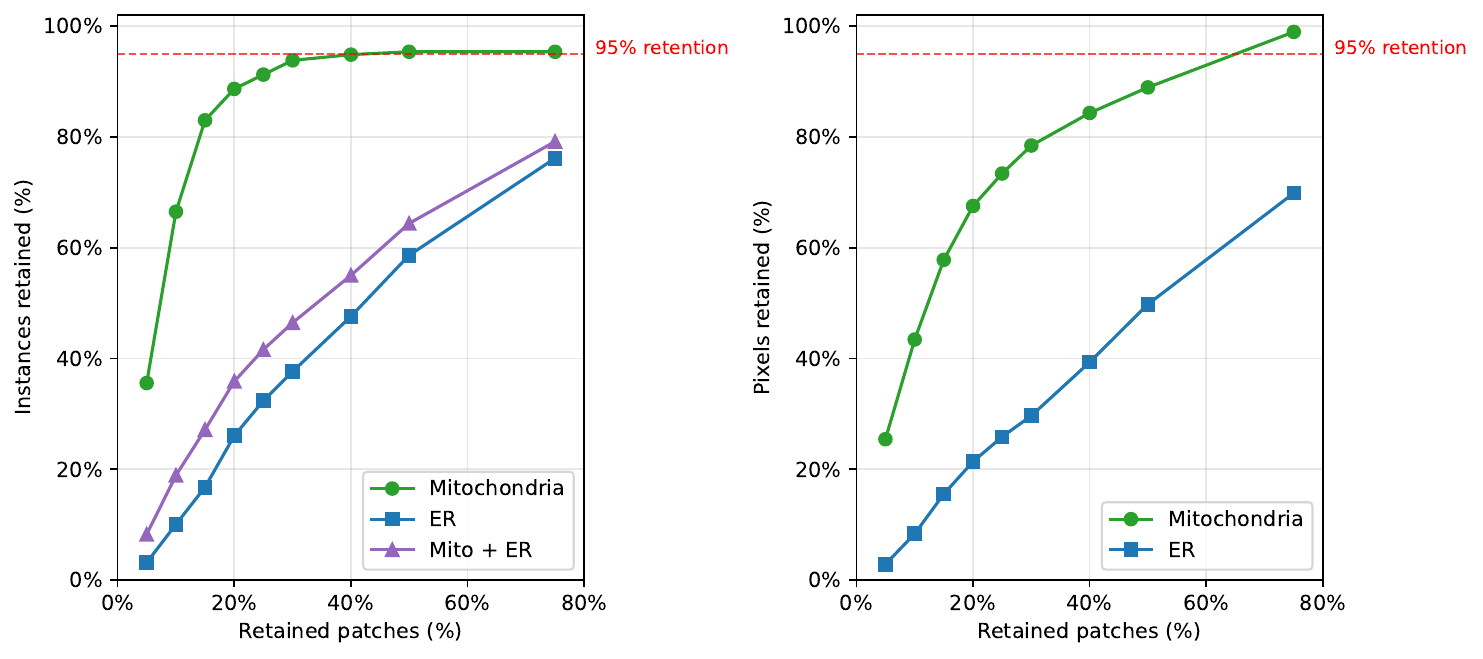}
	\caption{\revisiontwo{Operational retention analysis for the \EMBL dataset (fine-tuned classical autoencoder, split size $30$).  Patches are ranked by similarity and the top $q$ fraction is retained.  (Left) percentage of two-dimensional connected components that intersect at least one retained patch; (right) percentage of target pixels that lie in the union of retained patches.  The dashed line marks $95\%$ retention.  Retaining $30\%$ of patches already preserves $93.8\%$ of mitochondrial instances and $78.4\%$ of mitochondrial pixels; retaining $50\%$ reaches $95.4\%$ and $88.9\%$, respectively.}}
	\label{fig:operational-retention}
\end{figure*}

\revisiontwo{To quantify the screening efficiency of our algorithm, we measure how much of the target structures are retained when only the highest-ranked patches are passed to downstream analysis. On the \EMBL dataset, retaining the top $30\%$ of patches preserves $93.8\%$ of mitochondrial instances and $78.4\%$ of mitochondrial pixels; retaining $50\%$ increases these values to $95.4\%$ and $88.9\%$, respectively (\Cref{fig:operational-retention}). This demonstrates that downstream segmentation or quantitative analysis can be restricted to a substantially smaller search space while retaining most target occurrences.}

\subsubsection{\revisiontwo{Generalization to larger z-distances}}

\begin{figure}
    \centering
    \subfloat[\revision{Precision.}]{%
        \includegraphics[width=0.48\linewidth]{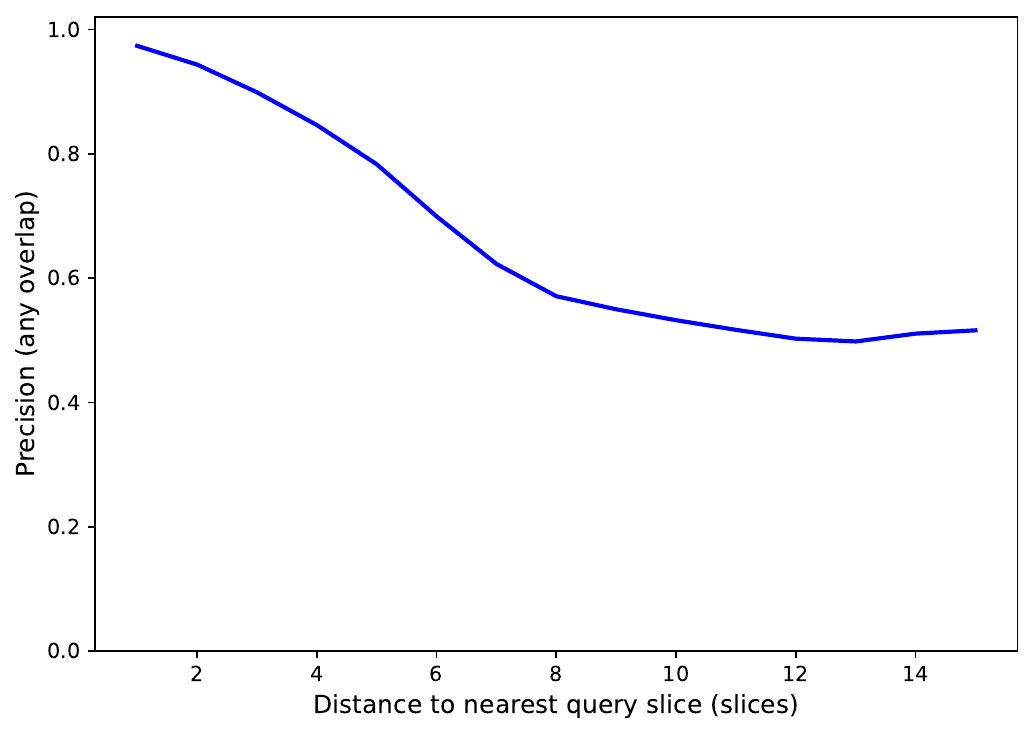}%
        \label{fig:prec-vs-distance-prec}%
    }\hfill
    \subfloat[\revision{Recall.}]{%
        \includegraphics[width=0.48\linewidth]{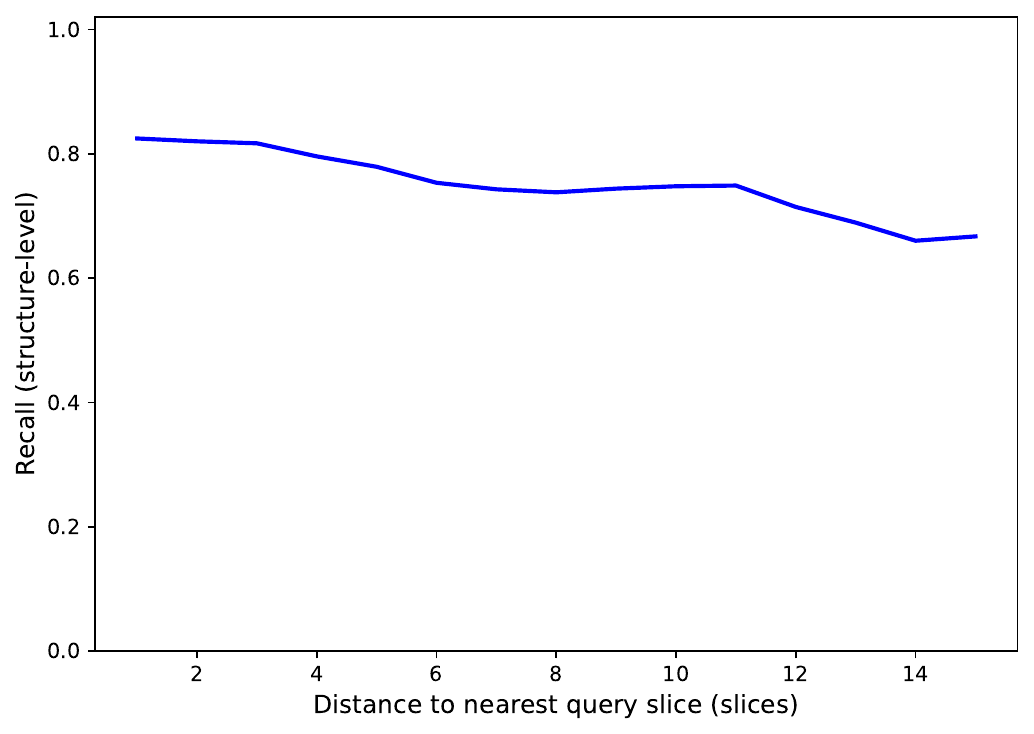}%
        \label{fig:prec-vs-distance-rec}%
    }
    \caption{\revision{Retrieval quality as a function of the $z$-distance from each retrieved patch to the nearest query slice (\Po dataset, fine-tuned classical autoencoder, split size $30$, top-$10$\% similarity threshold). Both metrics decay smoothly with distance but remain useful well beyond a single neighboring slice.}}
    \label{fig:prec-vs-distance}
\end{figure}

\revision{To test whether the proposed search algorithm mainly retrieves nearby patches or generalizes across the volume, we stratify retrieved patches by their $z$-distance to the nearest query slice and recompute precision and recall. \Cref{fig:prec-vs-distance} reports the results on the \Po dataset with split size $30$ and a top-$10$\% similarity threshold. Within $\pm 5$ slices, precision is approximately 90\% and recall is 83\%; both decline smoothly with distance but remain useful across the full $30$-slice group.}

\begin{figure}
    \centering
    \subfloat{%
        \includegraphics[width=0.72\textwidth]{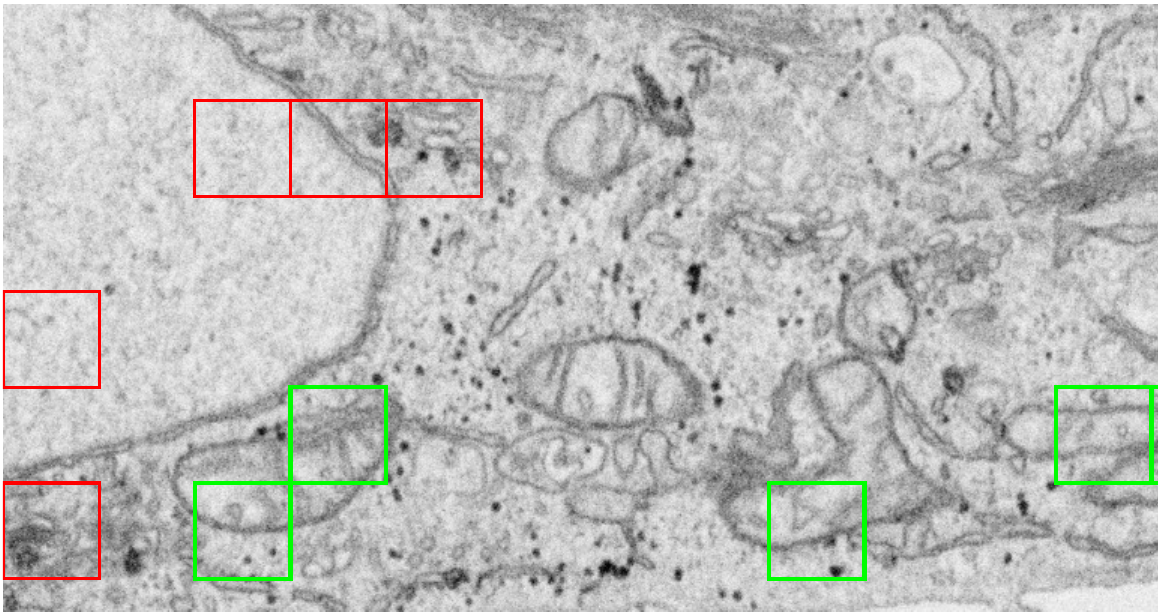}%
    }\\
    \subfloat{%
        \includegraphics[width=0.72\textwidth]{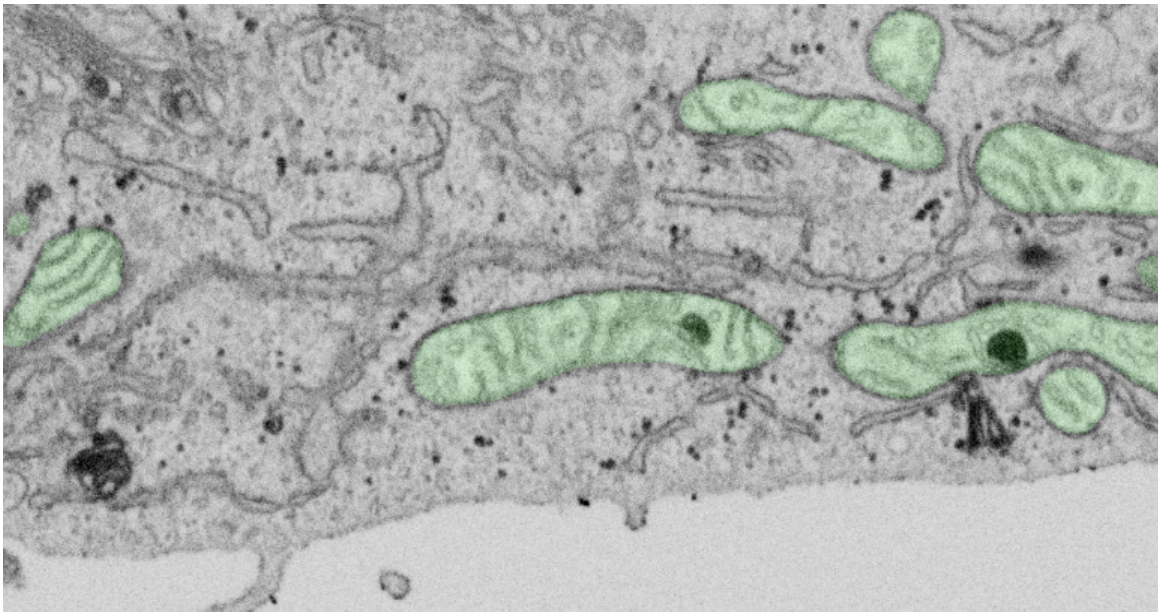}%
    }\\
    \subfloat{%
        \includegraphics[width=0.72\textwidth]{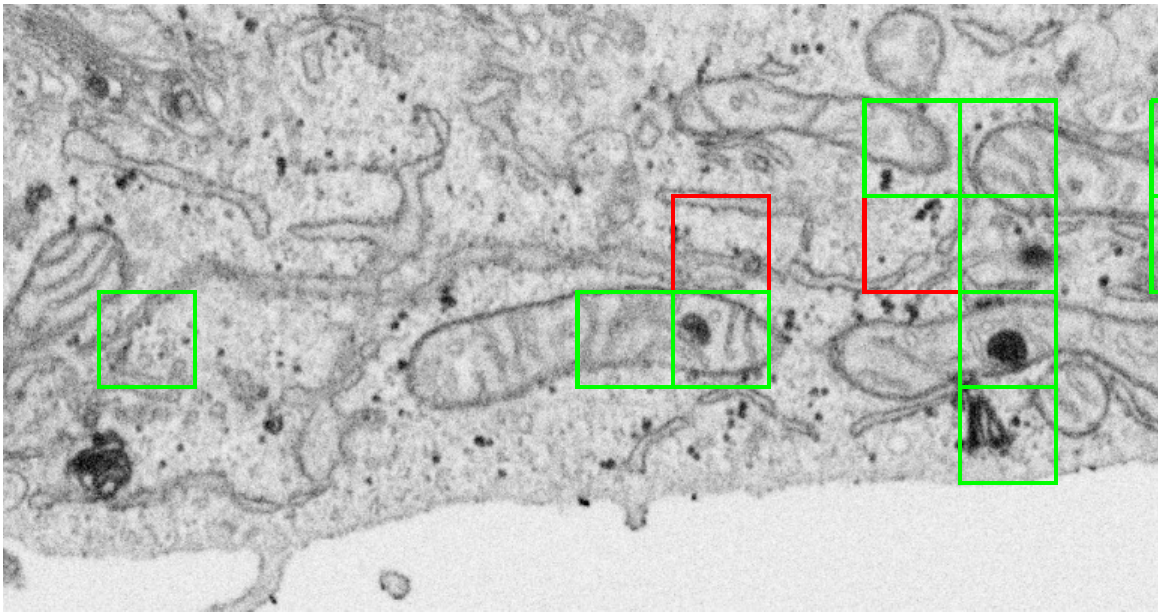}%
    }
    \caption{\revision{Long-range generalization. Queries are extracted from the first $20$ slices (a) and retrieval is performed on a slice $323$ slices away (b, c). Despite the very large $z$-gap, 8 out of 9 structures present in the target slice are recovered.} \revisiontwo{The pixel-level coloring in (b) is the manually annotated ground truth, shown only to indicate where mitochondria are located; it is not produced by the method. In (c), colored rectangles mark whole retrieved \emph{patches} (green: overlaps a mitochondrion; red: no structure), illustrating that the framework returns candidate regions rather than pixel-accurate segmentations.}}
    \label{fig:prec-vs-distance-long}
\end{figure}

\revision{Beyond the standard split-size regime, a complementary long-range experiment in which queries are taken only from the first $20$ slices of the volume and retrieval is performed on a target slice, approximately 320 slices away. \Cref{fig:prec-vs-distance-long} shows that, even at this extreme $z$-gap, the descriptor still discriminates structure-bearing patches: with a well-chosen similarity threshold, 8 out of 9 structures present in the target slice are recovered. Together, these two experiments indicate that (i) the practically used split-size strategy operates in the high-quality regime, (ii) retrieval quality degrades progressively rather than abruptly with $z$-distance, and (iii) increasing the density of query slices yields rapidly diminishing returns once the typical $z$-correlation length of the structures of interest is reached. Sparse queries, therefore, remain informative for screening tasks, in which high recall is more important than high precision.}

\subsubsection{\revisiontwo{Retrieval of a second, non-mitochondrial structure: endoplasmic reticulum}}
\label{sec:er-retrieval}

\revisiontwo{The quantitative evaluation above focuses on mitochondria, which are abundant across the three datasets and allow direct comparison. To provide quantitative evidence that the framework is not specific to mitochondria, we repeat the same protocol on the endoplasmic reticulum (ER) annotations available for the \EMBL dataset, for split sizes 10 and 30.}

\revisiontwo{ER differs qualitatively from mitochondria: it forms thin, sheet-like membranous structures rather than compact, roughly convex organelles. Since instances of ER form smaller structures than mitochondria, we need to adapt the positive-overlap threshold to $15\%$ for the ER experiment; patches with $0\%$ overlap are still used as negatives, unchanged.}

\begin{figure}
	\centering
	\subfloat[Split size 10.]{%
		\includegraphics[width=0.48\linewidth]{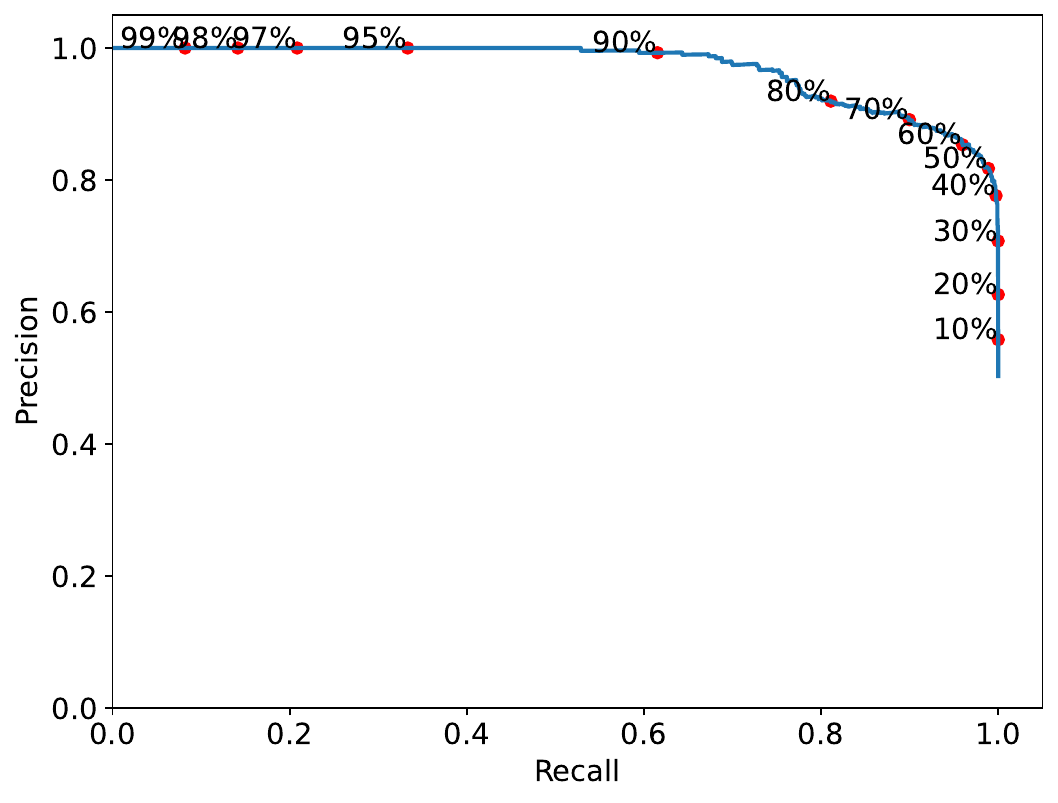}%
	}\hfill
	\subfloat[Split size 30.]{%
		\includegraphics[width=0.48\linewidth]{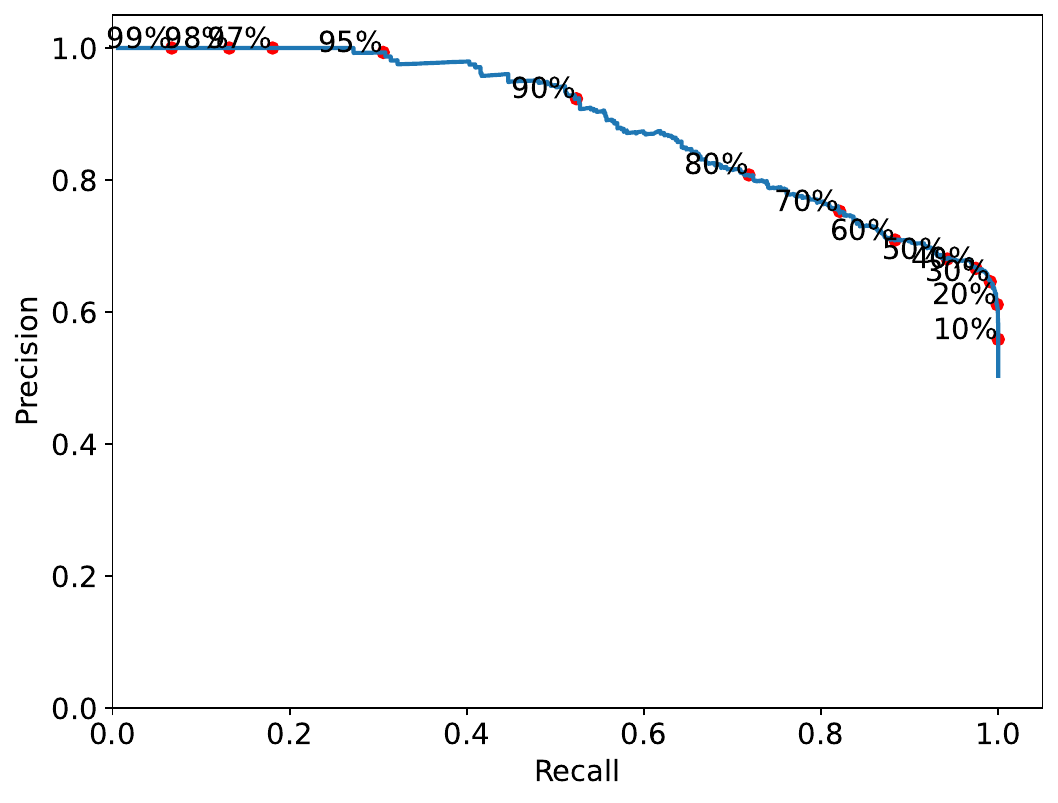}%
	}
	\caption{\revisiontwo{Precision-recall curves for ER retrieval on the \EMBL dataset (fine-tuned classical autoencoder), split sizes 10 (left) and 30 (right). Red points indicate reduction rates, as in \Cref{fig:pr-curves}.}}
	\label{fig:pr-curves-er}
\end{figure}

\revisiontwo{\Cref{tab:er-results} reports the AP for ER retrieval. The fine-tuned classical autoencoder again performs best, reaching an AP of $0.969$ (split size 10) and $0.882$ (split size 30) -- closely comparable to the mitochondria AP of $0.988$ and $0.887$ on the same dataset and split sizes (\Cref{tab:results}). \Cref{fig:pr-curves-er} shows the corresponding PR curves. This indicates that the descriptor and search framework generalize to a structurally distinct organelle without any changes to the model architecture or training procedure, once the overlap threshold is adapted to the structure's morphology.}

\begin{table}
	\caption{\revisiontwo{Average Precision (AP) for ER retrieval on the \EMBL dataset (structure label 2, positive-overlap threshold $15\%$), for the pretrained and fine-tuned (V)AE descriptors. The best-performing descriptor is indicated in \textbf{bold}.}}
	\label{tab:er-results}
	\centering
	\begin{tabular}{cccccc}
		\hline
		\multirow{2}{*}{\textbf{Split size}} & \multicolumn{2}{c}{\textbf{Pretrained}} & \multicolumn{2}{c}{\textbf{Fine-tuned}}\\
		\cline{2-5}
		 & AE & VAE & AE & VAE\\
		\hline
		10 & 0.961 & 0.964 & \textbf{0.969} & 0.963 \\
		30 & 0.851 & 0.859 & \textbf{0.882} & 0.872 \\
		\hline
	\end{tabular}
\end{table}

\subsubsection{\revisiontwo{Comparison with handcrafted descriptors}}
\label{sec:baselines}

\revisiontwo{In this section, we compare the fine-tuned autoencoder to three handcrafted descriptors: \emph{HOG}~\citep{dalal_histograms_2005}, \emph{LBP}~\citep{ojala_multiresolution_2002}, and dense \emph{SIFT}~\citep{lowe_object_1999}. %
To ensure a fair comparison with our compact 32-dimensional descriptor, the handcrafted descriptors, which contain up to thousands of features in their native form, are reduced to the same 32-dimensional representation before retrieval.}

\revisiontwo{We substitute each descriptor for the learned AE descriptor in the same search pipeline (split size $30$) and evaluate every labeled structure across the three datasets: mitochondria and ER in \EMBL, mitochondria in \EPFL, and mitochondria, nuclear envelope, vesicles, and ER in \Po. To account for differences in morphology, the positive-overlap threshold is set per structure to the $75^{\mathrm{th}}$ percentile of non-zero query-patch overlaps, capped at $50\%$; the resulting thresholds range from $16\%$ to $50\%$.}

\begin{figure}
	\centering
	\includegraphics[width=\linewidth]{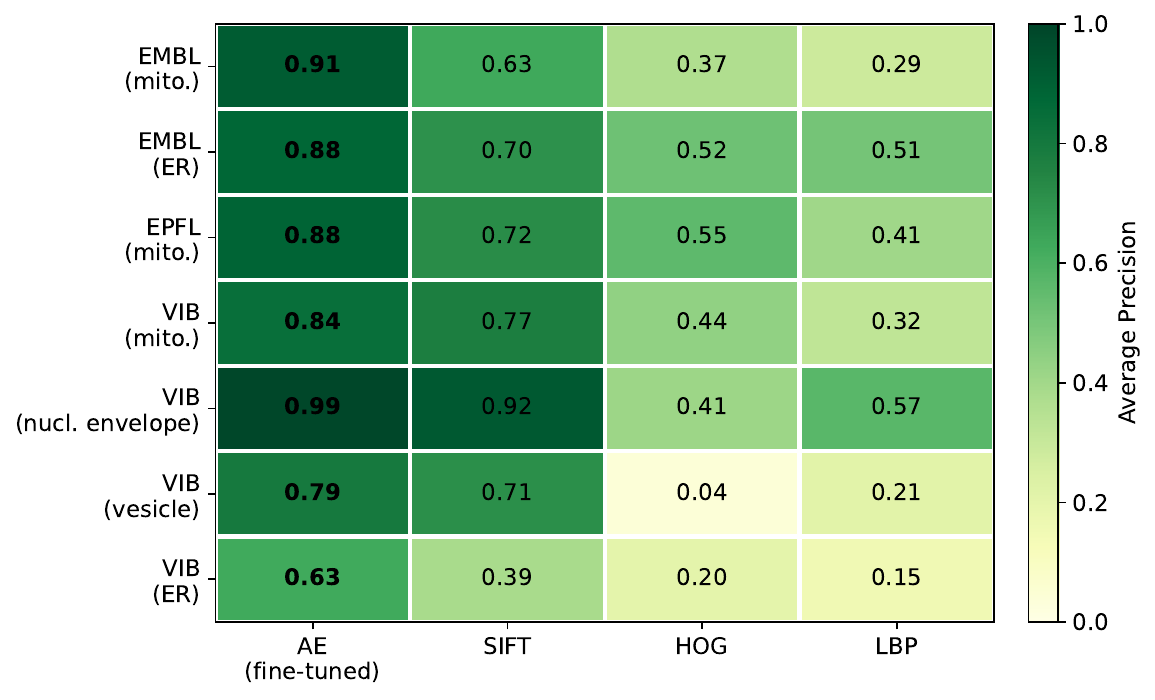}
	\caption{\revisiontwo{Average Precision (AP) of the fine-tuned autoencoder against three handcrafted descriptors (HOG, LBP, dense SIFT) compressed to the same $32$ dimensions by averaging contiguous blocks of their original feature vectors (split size $30$), for every labeled structure across the three datasets (mitochondria in \EMBL, \EPFL, and \Po; ER in \EMBL and \Po; nuclear envelope and vesicles in \Po). The AE attains the highest AP for every structure/dataset combination.}}
	\label{fig:equal-capacity-heatmap}
\end{figure}

\revisiontwo{\Cref{fig:equal-capacity-heatmap} shows that the fine-tuned AE achieves the highest AP for all seven structure/dataset combinations. It substantially outperforms HOG and LBP and is consistently ahead of SIFT, the strongest handcrafted baseline (e.g.\ $0.79$ vs.\ $0.71$ for \Po vesicles). These results indicate that the learned descriptor provides a more informative representation for the retrieval task under equal dimensionality.}

\subsubsection{\revisiontwo{Generalization to a different acquisition modality: serial block-face SEM}}
\label{sec:sbf-sem}

\revisiontwo{To test generalization beyond the (near-)isotropic FIB-SEM datasets, we additionally evaluate the framework on the publicly available \emph{Platelet-EM} SBF-SEM dataset~\citep{guay_platelet_2021}, which has a $5$--$10\times$ larger and strongly anisotropic slice spacing. We evaluate retrieval of three organelles---mitochondria, alpha granules, and the canalicular system---using the same protocol as for the other datasets (split size 10, patch size $80\times80$, and structure-specific overlap thresholds; \Cref{sec:baselines}).}

\begin{figure}
	\centering
	\subfloat[Mitochondria.]{%
		\includegraphics[width=0.31\linewidth]{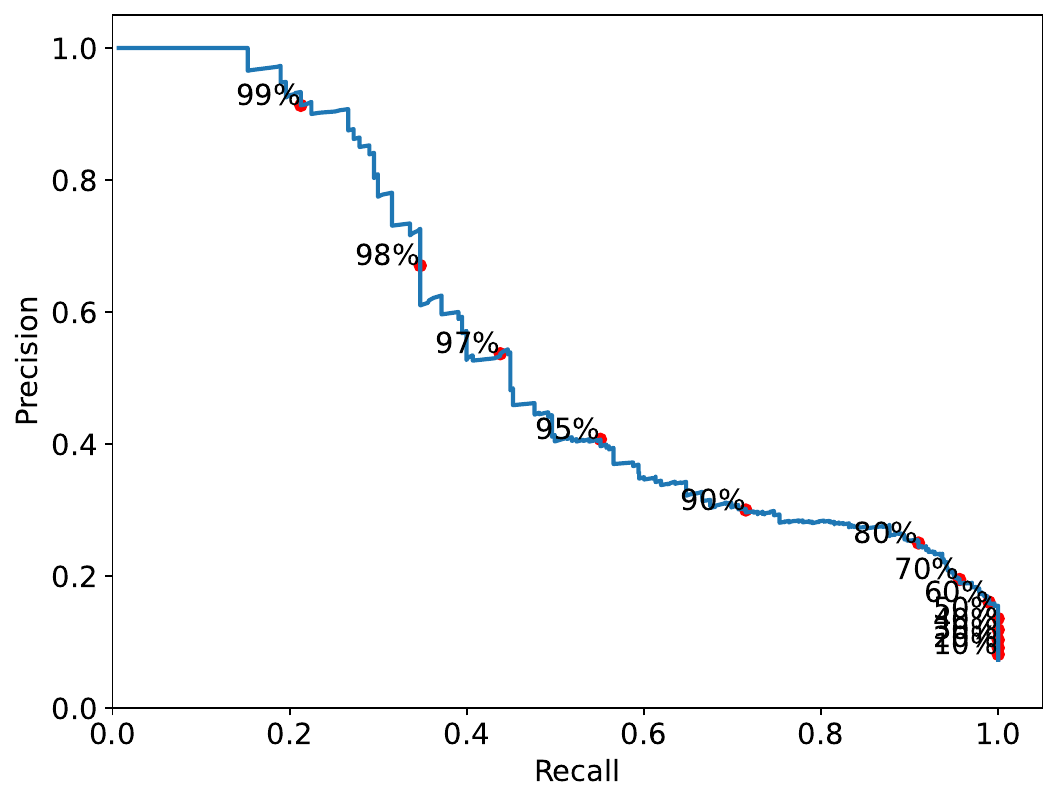}%
	}\hfill
	\subfloat[Alpha granule.]{%
		\includegraphics[width=0.31\linewidth]{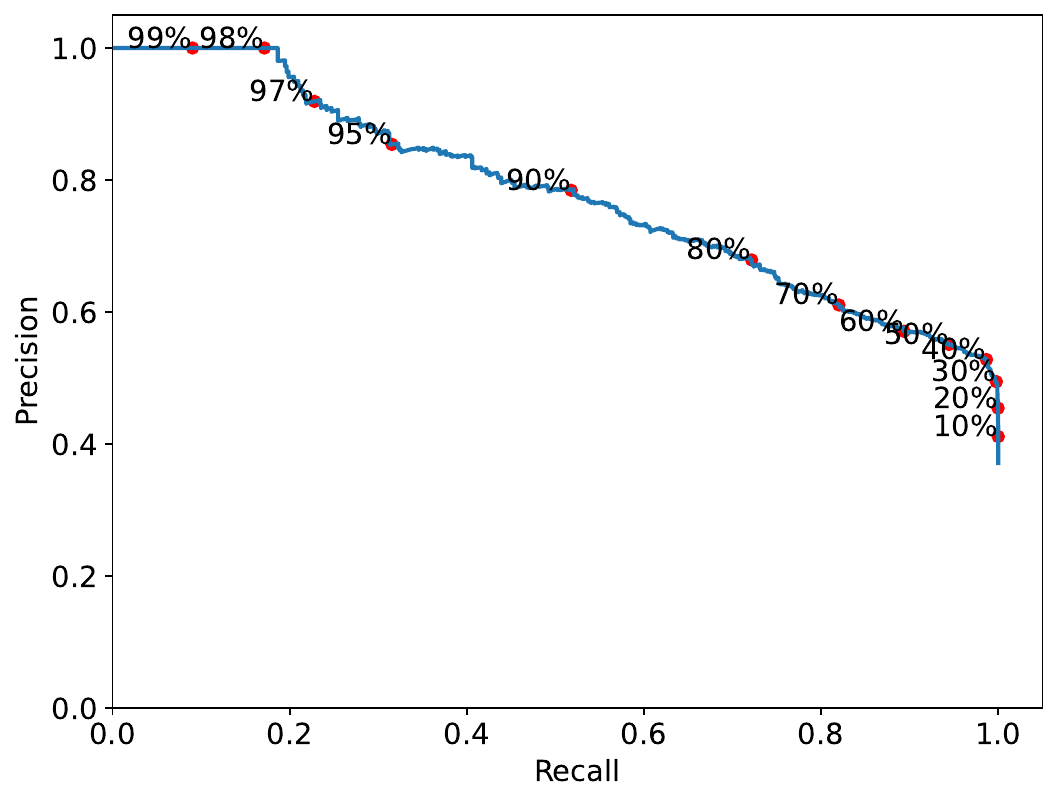}%
	}\hfill
	\subfloat[Canalicular vessel.]{%
		\includegraphics[width=0.31\linewidth]{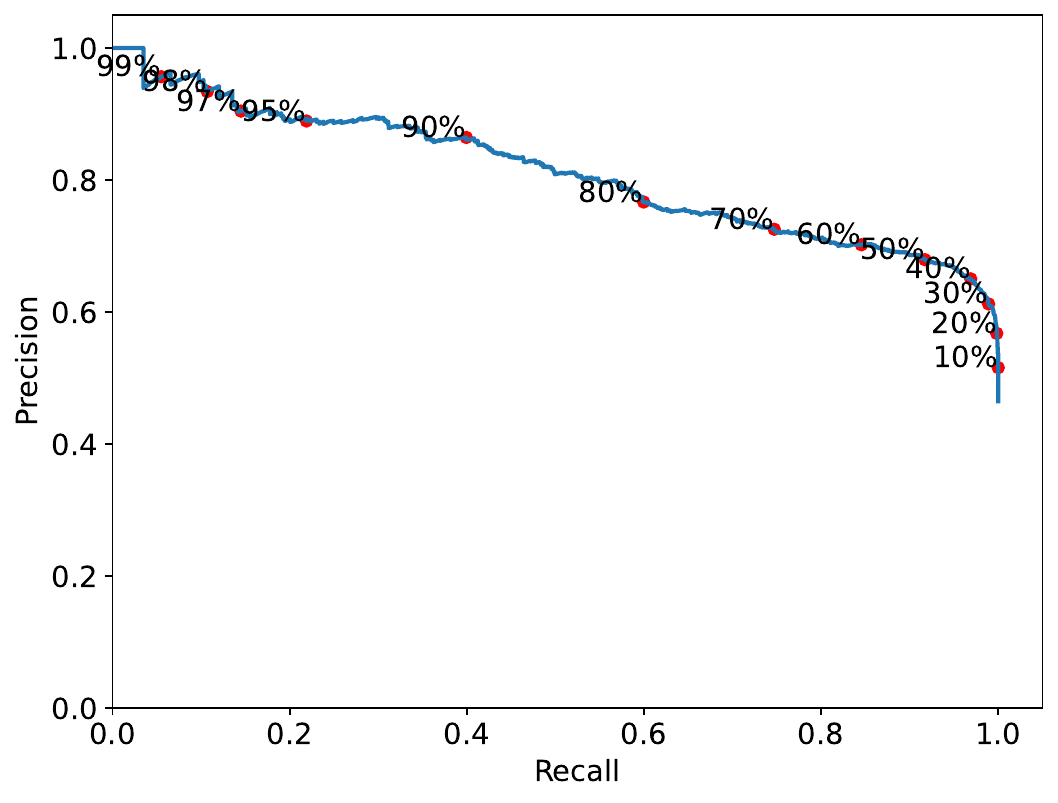}%
	}
	\caption{\revisiontwo{Precision-recall curves for retrieval on the SBF-SEM \emph{Platelet-EM} dataset (fine-tuned classical autoencoder, split size 10). Red points indicate reduction rates, as in \Cref{fig:pr-curves}.}}
	\label{fig:pr-curves-platelet}
\end{figure}

\revisiontwo{\Cref{tab:platelet} reports AP together with the base positive rate for each structure. Retrieval substantially exceeds the corresponding base rates: mitochondria reach an AP of $0.55$ versus a $7\%$ base rate, while alpha granules and canalicular vessels reach $0.78$ and $0.81$, respectively. The corresponding PR curves are shown in \Cref{fig:pr-curves-platelet}. Although performance is lower than on the FIB-SEM datasets (\Cref{tab:results}), likely due to the smaller volume and different acquisition characteristics, the results show that the framework transfers to strongly anisotropic SBF-SEM data without architectural changes. This provides evidence that the method is not intrinsically tied to (near-)isotropic FIB-SEM acquisition.}

\begin{table}
	\caption{\revisiontwo{Average Precision (AP) for retrieval on the SBF-SEM \emph{Platelet-EM} dataset~\citep{guay_platelet_2021} (split size 10), with the base positive rate (AP of a random ranking) shown for context.}}
	\label{tab:platelet}
	\centering
	\begin{tabular}{cccc}
		\hline
		\textbf{Structure} & \textbf{Threshold} & \textbf{Base rate} & \textbf{AP} \\
		\hline
		Mitochondria & 9\% & 7\% & 0.55 \\
		Alpha granule & 20\% & 37\% & 0.78 \\
		Canalicular vessel & 9\% & 46\% & 0.81 \\
		\hline
	\end{tabular}
\end{table}

\subsubsection{\revision{Analysis of time savings}}
\label{sec:timing}

\revisiontwo{We additionally estimate the corresponding workflow time savings under explicit assumptions and report measured retrieval times for reproducibility. Manual inspection is assumed to require $20$~s per slice, while the retrieval-based workflow adds $5$~s per slice for candidate validation. Full-volume retrieval (fine-tuned classical autoencoder, $80\times80$ patches) was benchmarked on all three datasets and split sizes 10 and 30 using an AMD Ryzen~9 9900X CPU and NVIDIA GeForce RTX~5060~Ti GPU (16~GB), with PyTorch 2.12 and CUDA 13.2. \Cref{tab:timing} reports retrieval time, throughput, and estimated workflow speed-up. Retrieval takes at most a few tens of seconds and yields an estimated $\approx4\times$ speed-up ($74$--$75\%$ reduction in analysis time). Since patch encoding is currently performed one at a time on the CPU without GPU batching, these timings are conservative and could be further improved.}

\begin{table}
	\caption{\revisiontwo{Measured full-volume retrieval time and estimated time savings versus manual inspection, per dataset and split size (fine-tuned classical autoencoder, patch size $80\times 80$; hardware and protocol described in the text). Manual time assumes $20$~s/slice; retrieval-based time is the measured retrieval time plus $5$~s/slice for candidate validation. Measured throughput ranges from $1{,}043$ to $1{,}451$ patches/s across all configurations.}}
	\label{tab:timing}
	\centering
	\resizebox{\textwidth}{!}{\begin{tabular}{cccccc}
		\hline
		\textbf{Dataset} & \textbf{Split} & \textbf{Patches} & \textbf{Retrieval} & \textbf{Manual} & \textbf{Speed-up} \\
		 & \textbf{size} & & \textbf{time (s)} & \textbf{vs.\ retrieval} & \\
		\hline
		\multirow{2}{*}{\EMBL} & 10 & 2{,}793 & 2.3 & 21.3~min vs.\ 5.4~min & $3.97\times$ \\
		 & 30 & 2{,}989 & 2.1 & 21.3~min vs.\ 5.4~min & $3.97\times$ \\
		\hline
		\multirow{2}{*}{\EPFL} & 10 & 38{,}610 & 38.1 & 110.0~min vs.\ 28.1~min & $3.91\times$ \\
		 & 30 & 41{,}470 & 34.3 & 110.0~min vs.\ 28.1~min & $3.92\times$ \\
		\hline
		\multirow{2}{*}{\Po} & 10 & 30{,}849 & 31.5 & 125.7~min vs.\ 31.9~min & $3.93\times$ \\
		 & 30 & 33{,}124 & 29.8 & 125.7~min vs.\ 31.9~min & $3.94\times$ \\
		\hline
	\end{tabular}}
\end{table}

\section{Conclusion}
\label{sec:conclusion}

In this work, we proposed and evaluated a novel search framework for locating structures of interest in large datasets.

Our framework effectively locates image areas of interest, thus reducing significantly the search space for subsequent visual analysis or further automatic process steps. This is achieved with a small and flexibly chosen number of queries and minimal computation time. Furthermore, it enables users to refine search queries interactively, resulting in improved performance.

\revisiontwo{Rather than replacing structure-specific segmentation, our framework provides an upstream, target-flexible screening stage that identifies regions worth further analysis from only a few example patches. This allows manual inspection, segmentation, or quantitative analysis to be restricted to a smaller search space while retaining most target occurrences. Retrieval and segmentation are therefore complementary: the former determines \emph{where} to look, while the latter delineates \emph{what} is there.}

We also presented a thorough evaluation of classical and variational autoencoders to discriminate between structures of interest. We have demonstrated the ability to distinguish between structures of interest within the latent space. We have shown that both the AE and the VAE learn efficient features for locating structures of interest.

Finally, we demonstrated that fine-tuning the pretrained models to better capture the specific structures of vEM data significantly enhances their performance. However, the AE benefits much more from fine-tuning than the VAE, which might be attributed to the additional hyperparameter $\beta$, which requires careful tuning~\citep{zizakic_efficient_2022} to obtain optimal results, which is impractical for this use case.

\revisiontwo{The framework also has some limitations: (1) retrieval performance depends on the morphology and prevalence of the target structure, requiring adaptation of parameters such as the positive-overlap threshold and potentially limiting performance for very rare structures; (2) performance also decreases with increasing $z$-distance from the query slices; (3) while we demonstrate retrieval across several organelles and provide initial validation on strongly anisotropic SBF-SEM data, broader evaluation across structures, acquisition modalities, and unseen datasets remains necessary; (4) finally, the current method retrieves 2D candidate patches rather than 3D object boundaries and is therefore intended as an upstream step for visual inspection or subsequent automatic processing.}

Further research directions could include implementing this search framework in a labeling tool~\citep{labelstudio,biosegment}. The similarity computation on the considered patches could be parallelized through the Dask~\citep{dask} library. Also, clustering techniques could be interesting, which we did not study in this work. Finally, it could be interesting to analyze the learning curve of our framework when using a human-in-the-loop approach.
\revisiontwo{Future work will also include broader multi-organelle and anisotropic-vEM evaluation, cross-dataset transfer, and extensions to 3D retrieval.}

\section*{Data availability}

The code underlying this article is available at \url{https://gaim.ugent.be/software/patch-based-querying}.
The \EPFL dataset is publicly available at \url{https://cvlab.epfl.ch/data/data-em/}.
The \Po dataset is not publicly available, but will be shared on reasonable request to the corresponding author.
The \EMBL dataset \revision{is publicly available at \url{https://www.ebi.ac.uk/empiar/EMPIAR-10311/}}.
\revisiontwo{The \emph{Platelet-EM} dataset used in \Cref{sec:sbf-sem} is publicly available at \url{https://bio3d-vision.github.io/platelet-description}.}

\section*{Declaration of competing interest}

None declared.

\section*{CRediT authorship contribution statement}

\textbf{N. Vyncke:}
Formal analysis, investigation, methodology, software, visualization, writing -- original draft.
\textbf{N. Nadisic:}
Supervision, writing -- review \& editing.
\textbf{Y. Saeys:}
Conceptualization, data curation, resources, validation, writing -- review \& editing.
\textbf{A. Pi\v{z}urica:}
Conceptualization, formal analysis, resources, supervision, writing -- review \& editing.

\section*{Ethics statement}

This study did not involve any new experiments on human or animal subjects. All analyses were based exclusively on existing \revisiontwo{human and animal datasets,}
all of which are either publicly available or were obtained through data-sharing agreements subject to policies at Ghent University and the VIB institute.
Ethical approval was therefore not required for the present study, as no new experimental procedures, animal handling, or interventions were performed by the authors. The original data collection for each dataset was conducted in accordance with relevant laws, institutional guidelines, and ethical approvals, as reported by the original data providers.

\section*{Funding}

This work was supported by the Flanders AI Research Program [174B09119].

\bibliographystyle{elsarticle-num} 
\bibliography{bib}
\end{document}